\documentstyle[fullname,pstricks,pst-node]{report}
\newcommand{\unif}{\sqcup}
\newcommand{\unific}{\sqcup}
\newcommand{\meet}{\sqcap}
\newcommand{\qbar}{\bar{q}}
\newcommand{\Qbar}{\bar{Q}}
\newcommand{\onto}{\rightarrow}

\newcommand{\len}[1]{\mbox{$len(#1)$}}
\newcommand{\amrs}[1]{\mbox{$\langle Ind_{#1},\Pi_{#1},\Theta_{#1},\approx_{#1} \rangle$}}
\newcommand{\afs}[1]{\mbox{$\langle \Pi_{#1},\Theta_{#1},\approx_{#1} \rangle$}}
\newcommand{\seq}[1]{\mbox{$\{ 1,\ldots,#1 \}$}}

\newcommand{\mf}{\tt} 
\newcommand{\type}{\em}

\newcommand{\nterm}[1]{\mbox{\em #1}}

\newcommand{\tag}[1]{\fbox{\footnotesize #1}}

\newcommand{\rules}{{\cal R}}

\newcommand{\feats}{\mbox{\sc Feats}}
\newcommand{\types}{\mbox{\sc Types}}
\newcommand{\nodes}{\mbox{\sc Nodes}}
\newcommand{\items}{\mbox{\sc Items}}

\newcommand{\fs}{\mbox{\sc FS}}

\newcommand{\paths}{\mbox{\sc Paths}}
\newcommand{\words}{\mbox{\sc Words}}

\newcommand{\subsumes}{\sqsubseteq}
\newcommand{\subsumed}{\sqsupseteq}

\newcommand{\cross}{\times}

\newcommand{\get}{\mbox{$\leftarrow\;$}}

\newcommand{\derives}{\rightarrow}

\newcommand{\aderives}{\leadsto}
\newcommand{\derivess}{\stackrel{*}{\rightarrow}}

\newcommand{\ns}{\mbox{$I\!\!N$}}
\newcommand{\isdef}{\!\!\downarrow}
\newcommand{\isndef}{\!\!\uparrow}

\newcommand{\emptymrs}{\lambda}

\newcommand{\act}{\mbox{\sc Act}}
\newcommand{\comp}{\mbox{\sc Comp}}

\newcommand{\amalia}{\mbox{$\cal{AMALIA}$}}

\newtheorem{definition}{Definition}[section]
\newtheorem{theorem}[definition]{Theorem}
\newtheorem{lemma}[definition]{Lemma}
\newtheorem{corollary}[definition]{Corollary}

\newcommand{\proof}[1]{{\bf Proof:}#1}
\newenvironment{patr-rule}[2]{\begin{tabbing} #1 $\rightarrow$ \= #2\\ }{\end{tabbing}}

\newenvironment{tfs}[1]
    {\left[\!\!\!\begin{array}{ll}
      \multicolumn{2}{l}{\mbox{\bf #1}}\\
    }
    {\end{array}\!\!\!\right]}

\newcommand{\Tabs}{xxxx\= xxxx\= xxxx\= xxxx\= xxxx\= xxxx\= xxxx\= xxxx\= xxxx \= xxxx\= xxxx\= xxxx\= \kill}
\newenvironment{program}[3]{\begin{figure*}[hbtp]
                         \begin{center}
                         \fbox{\mf
                         \begin{minipage}{\textwidth}
                         \begin{tabbing}
                         \Tabs 
                         #3
                         \end{tabbing}
                         \end{minipage}
                         }
                         \end{center}
                         \caption{#2}
                         \label{#1}
                         \end{figure*}}{}

\input{psfig}
\renewcommand{\topfraction}{0.9}

\renewcommand{\thepage}{} 

\newcommand{\mycaption}[2]{\caption{#1}}

\def\thebibliography#1{\chapter*{Bibliography}\list
 {}{\setlength{\labelwidth}{0pt}\setlength{\leftmargin}{\parindent}
 \setlength{\itemindent}{-\parindent}}
 \def\newblock{\hskip .11em plus .33em minus -.07em}
 \sloppy\clubpenalty4000\widowpenalty4000
 \sfcode`\.=1000\relax}

\begin{document}

\begin{titlepage}
\begin{center}

\vspace{6cm}

{\huge\bf An Abstract Machine for Unification Grammars}\\
\vspace{0.5cm}
{\Large\bf with Applications to an HPSG Grammar for Hebrew}\\
\vfill
{\Large\bf Research Thesis}\\
\vspace{1cm}
{\Large Submitted in partial fulfillment of the requirements for the
degree of Doctor of Science}\\
\vspace{1cm}
{\Large\bf Shalom Wintner}\\
\vfill
{\Large Submitted to the Technion -- Israel Institute of Technology}\\
\vspace{1cm}
{Tevet 5757 \hfill Haifa \hfill January 1997}
\end{center}
\end{titlepage}

\newpage

{\flushleft
The work described herein was supervised by Prof.\ Nissim Francez
Under the Auspices of the Faculty of Computer Science
}

\vfill

\section*{Acknowledgments}
This work could never have been what it is without the support I
received from my advisor, Nissim Francez. I am grateful to Nissim for
introducing me to this subject, leading and advising me throughout the
project, bearing with me when I messed things up and encouraging me
all along the way.

Many thanks are due to the members of my Thesis Committee: Bob
Carpenter, Michael Elhadad, Alon Itai, Uzzi Ornan and Mori Rimon. I
am especially indebted to Bob for his major help, mostly in the
preliminary stages of this project. I also want to thank Uzzi for
replacing Nissim when he was on sabbatical.

While working on this thesis I spent a fruitful summer in the
University of T\"ubingen, Seminar f\"ur Sprachwissenschaft, during
which I learned a lot from many discussions. I wish to thank Paul King
for his generosity and his wisdom. Thanks are also due to Erhard
Hinrichs, Dale Gerdemann, Thilo G\"otz, Detmar Meurers, John Griffith
and Frank Morawietz. I am grateful to Evgeniy Gabrilovich for many
fruitful discussions and for going over my code.

Special thanks to Holger Maier and Katrine Kirk for their hospitality
and their company. Finally, I want to thank Yifat and Galia for being
there when I needed it.

The generous financial help of the Minerva Stipendien Komitee and
Intel Israel Ltd.\ is gratefully acknowledged.  The research was supported
by a grant from the Israeli Ministry of Science: ``Programming
Languages Induced Computational Linguistics'' and by the Fund for the
Promotion of Research in the Technion. The project was carried out
under the auspices of the Laboratory for Computational Linguistics of
the Technion.

\tableofcontents
\listoffigures
\clearpage
\pagestyle{plain}
\renewcommand{\thepage}{\arabic{page}}
\setcounter{page}{1}
\pagenumbering{arabic}

\newpage
\addtolength{\baselineskip}{0.1cm}
\begin{center}
{\Large\bf Abstract}
\end{center}
\addcontentsline{toc}{chapter}{\protect\numberline{}{Abstract}}

Contemporary linguistic formalisms have become so
rigorous that it is now possible to view them as very high level
declarative programming languages.  
Consequently, {\em grammars\/} for
natural languages can be viewed as {\em programs}; this view enables
the application of various methods and techniques that were proved
useful for programming languages to the study of natural languages.

One of the most successful implementation techniques for logic
programming languages involves the use of an abstract machine. In this
approach one defines an abstract machine with the following
properties: it is close enough to the high-level language, thus
allowing efficient compilation to the abstract machine language; and
it is sufficiently low-level to allow efficient interpretation of the
machine instructions on a variety of host architectures.  Abstract
machines were used for processing procedural and functional languages,
but they gained much popularity for logic programming languages since
the introduction of the Warren Abstract Machine (WAM). Most current
implementations of Prolog, as well as other logic languages, are based
on abstract machines.  The incorporation of such techniques usually
leads to very efficient compilers in terms of both space and time
requirements.

In this work we have designed and implemented an abstract machine,
\amalia, for the linguistic formalism ALE, which is based on typed
feature structures. This formalism is one of the most widely accepted
in computational linguistics and has been used for designing grammars
in various linguistic theories, most notably HPSG. \amalia\ is
composed of data structures and a set of instructions, augmented by a
compiler from the grammatical formalism to the abstract instructions,
and a (portable) interpreter of the abstract instructions. The effect
of each instruction is defined using a low-level language that can be
executed on ordinary hardware.

The advantages of the abstract machine approach are twofold. From a
theoretical point of view, the abstract machine gives a well-defined
operational semantics to the grammatical formalism. This ensures that
grammars specified using our system are endowed with well defined
meaning. It enables, for example, to formally verify the correctness
of a compiler for HPSG, given an independent definition. From a
practical point of view, \amalia\ is the first system that employs a
direct compilation scheme for unification grammars that are based on
typed feature structures. The use of \amalia\ results in a much
improved performance over existing systems.

In order to test the machine on a realistic application, we have
developed a small-scale, HPSG-based grammar for a fragment of the
Hebrew language, using \amalia\ as the development platform. This is
the first application of HPSG to a Semitic language.

\newpage
\addtolength{\baselineskip}{-0.1cm}
\section*{List of Abbreviations}
\begin{tabular}{ll}
AFS     & Abstract Feature Structure\\
ALE     & Attribute Logic Engine\\
\amalia & Abstract MAchine for LInguistic Applications\\
AMRS    & Abstract Multi-rooted Feature Structure\\
AVM     & Attribute Value Matrix\\
DLR     & Definite Lexical Rule\\
FOT     & First Order Term\\
GUI     & Graphical User Interface\\
HPSG    & Head-Driven Phrase Structure Grammar\\
LFG     & Lexical-Functional Grammar\\
LUB     & Least Upper Bound\\
MRS     & Multi-rooted Feature Structure\\
NL      & Natural Language\\
NLR     & `Nismak' Lexical Rule\\
NP      & Noun Phrase\\
PT      & Pre-terminal\\
REF     & REFerence\\
STR     & STRucture\\
TFS     & Typed Feature Structure\\
WAM     & Warren Abstract Machine\\
\end{tabular}

\chapter{Introduction}
\label{sec:intro}

\section{Motivation}
Research in linguistics has traditionally been aimed at describing the
structure of Natural Languages (NLs). Since the 1950s, however, the
focus has shifted from attempts to provide such descriptions to the
definition of the right way in which to stipulate them. During the
past few decades many such formalisms were devised. A `good' model,
according to \namecite{shieber86}, is linguistically felicitous,
expressive and computationally effective. It must be powerful enough
to capture the wealth and diversity of NLs, yet it must be
computationally tractable to allow for computational processing.

Contemporary linguistic formalisms such as LFG \cite{lfg} or HPSG
\cite{hpsg2} have become so rigorous that it is now possible to view
them as very high level declarative programming languages.  In this
metaphor a {\em grammar} for a natural language, formally specified
using one of the modern frameworks described above, can be viewed as a
{\em program}. The execution of a grammar on an input sentence yields
an output which represents the sentence's structure. This view enables
the application of various methods and techniques that were proved
useful for programming languages to the study of natural languages.

Historically, many computational fields of research originated from
the study of natural languages: important aspects the theory of formal
languages are due to Chomsky, for example; and more recently, Prolog
originated out of an attempt to provide a language for description
of natural languages. Today, however, much progress was achieved in
the area of programming languages. Tools and techniques were developed
that enable efficient processing of such languages and, more
importantly, formal propositions to be made and proved over languages
in general and specific programs in particular. These advances are now
being incorporated into the realm of natural languages. Grammars for
natural languages are specified more precisely; their properties can
be mathematically stated; and their processing becomes more efficient.
For a survey of some such approaches, see \cite{shieber86}; for
examples of the advantages of regarding natural language formalisms as
programming languages, see \cite{barberris,manaster}.

This work introduces such an application: an implementation technique
that is common for logic programming languages, namely the use of an
{\em abstract machine}, is applied to (a subset of) the ALE formalism
\cite{ale}, originally designed for specifying feature-structure based
phrase-structure grammars. Abstract machines were used for processing
procedural and functional languages, but they gained much popularity
for logic programming languages since the introduction of the Warren
Abstract Machine (WAM -- see~\cite{waren83} and a tutorial
in~\cite{wam}). Most current implementations of Prolog, as well as
other logic languages, are based on abstract machines.  The
incorporation of such techniques usually leads to very efficient
compilers in terms of both space and time requirements.

\amalia\ is an abstract machine, specifically tailored for processing
ALE grammars. It is composed of data structures and a set of
instructions, augmented by a compiler from the grammatical formalism
to the abstract instructions, and a (portable) interpreter of the
abstract instructions. The effect of each instruction is defined using
a low-level language that can be executed on ordinary hardware.  The
advantages of the abstract machine approach are twofold. From a
theoretical point of view, the abstract machine gives a well-defined
operational semantics to the grammatical formalism. This ensures that
grammars specified using our system are endowed with well defined
meaning. It enables, for example, to formally verify the correctness
of a compiler for HPSG, given an independent definition. From a
practical point of view, \amalia\ is the first system that employs a
direct compilation scheme for unification grammars that are based on
typed feature structures. The use of \amalia\ results in a much
improved performance over existing systems (in particular, ALE itself).

\section{Literature Survey}

\subsection{Grammatical Formalisms}
Much of the recent research in computational linguistics has been
directed towards defining a good model in which natural languages
would be naturally describable.  Having its roots in the study of formal
languages, this endeavor started with considering the computational
power needed for describing natural languages in general; context free
grammars were thus ruled out quite early. But even if one did believe
that natural language were, indeed, within the scope of context free
languages, one had to admit that context free grammars were not the
ideal framework in which to develop grammars for the natural
languages. It was understood that the weak generation properties of a
grammar in a given formalism (i.e., its ability to recognize all and
only the sentences of a language) are not sufficient -- there is a
need in providing syntactic descriptions that cohere with the way
linguists capture the language.

The resulting trend in computational linguistics was to use
unification-based formalisms to obtain these two goals. While many
such frameworks were developed (see \cite{shieber86} for a good
review), some notions are common to most of them. They are all based
on a context free skeleton, where non-terminal symbols are replaced
with structured, more complex entities; and the basic operation on
these structures is unification. Among these frameworks are Functional
Unification Grammar~\cite{fug}, Lexical Functional Grammar~\cite{lfg},
Generalized Phrase-Structure Grammar~\cite{gpsg} and many others.

A unification-based grammar formalism is a meta-language for
describing grammars for (natural) languages. The basic entity of such
formalisms is the {\em feature structure} -- a data structure
consisting of a set of feature-value pairs. While different frameworks
define feature structures differently, they can in general be captured
as directed graphs, where the arcs are labeled with feature names and
an $f$-labeled arc connects nodes $v$ and $u$ if and only if the value
of the feature $f$ in the feature structure associated with $v$ is the
feature structure associated with $u$. For a good, informal survey of
feature structures and their properties refer to~\namecite{shieber86}.

Feature structures can be thought of as an extension of first order
terms (see \cite{carp91,life-meaning}), where the sub-terms are coded
by feature names rather than by positions. They extend first order
terms in that they are in general graphs, whereas first order terms
are trees, with possibly shared leaves. Hence, a term might be a
common part of more than one sub-term. Some grammatical formalisms
decorate feature structures with {\em types}, or {\em sorts}, that can
be captured by labels on the nodes of the graph.  Feature structures
are used by grammatical formalisms to represent linguistic concepts
such as words, phrases and sometimes even grammatical rules.

The basic operation on feature structures is {\em unification}. Being
very much like first-order term unification, this operation combines
the information that is encoded by two feature structures and produces
a result that contains the unified information, provided that the two
arguments don't contain contradicting information. If the arguments
are inconsistent, unification is said to {\em fail}.

In this work we are mainly concerned with {\em typed\/} feature
structures (TFSs), as described in~\cite{carp92}. As their name
suggests, each such structure has a type, drawn from a pre-defined,
partially ordered set of types. The {\em type hierarchy\/} helps the
grammar writer to organize linguistic knowledge in a similar way to
common knowledge representation languages. The hierarchy is accompanied by
an {\em appropriateness specification\/} that associates features with
types; for example, the `case' feature might be defined to be
appropriate for feature structures of type `noun' but not for
structures of type `verb'. Moreover, appropriateness is inherited: if
a feature is appropriate for a type $t$, then it is appropriate for
all the sub-types of $t$ as well. This property is reminiscent of
object-oriented systems; in particular, multiple inheritance is
supported in this framework.

It is important to note that while some of the above-mentioned
formalisms were designed as computational frameworks for developing
grammars, others were linguistically oriented in the sense that a
grammatical theory was encoded within them in one way or another.
Obviously, any such formalism defines at least the expressive power of
grammars that can be stipulated within it.  But many other linguistic
considerations and generalizations can be, and actually are,
hard-wired into some formalisms.

\subsection{The Current Role of HPSG}
Recently HPSG has become prominent among the various unification-based
formalisms. HPSG \cite{hpsg1,hpsg2} was developed by Pollard and Sag
as a variant of GPSG and Categorial Grammar, but immediately gained a
position of a well-founded, promising formalism for the description of
natural languages. It incorporates the notion of {\em typed} feature
structures, where types are partially ordered according to a defined
hierarchy, thus enabling very concise, general rules to be
stipulated. Much of the information carried by linguistic entities is
stored in the lexicon; as a result, grammar rules become few and very
general. HPSG defines a set of linguistically plausible schemas, or
universal principles, that are said to hold for all natural languages
and are part of every grammar. In addition, language specific rules
can be specified in any given grammar.

Due to its generality and elegance, HPSG has gained a lot of
popularity.  It enables the designing of grammars for various,
linguistically different, languages: work has been done on HPSG
grammars for English, German~\cite{german}, French,
Japanese~\cite{japanese}, Korean and many other languages (see a
bibliography in \cite{hpsg-bib} and an electronic bibliography in
\cite{hpsgbib}).  HPSG principles were used to describe not only the
syntax and semantics of languages, but also their morphology (e.g.,
\cite{nerbonne92}) and phonology (e.g.,
\cite{bird90,bird92}). It seems that linguists find this kind of
typed-feature-structures based formalism, with lexical rules and a
small set of very general grammatical rules, very convenient.

In spite of the interest that HPSG invokes, no formal definition of
the formalism exists. Both \cite{hpsg1} and \cite{hpsg2} are rather
linguistically oriented, in the sense that no mathematical definitions
are given for the language of HPSG
itself. King~\shortcite{king89,king92} gives a logical
formalization of \namecite{hpsg1}; it is an attempt to provide a
logical framework within which both the elementary entities of HPSG,
such as feature structures and types, and the principles and the
rules, can be described. While \namecite{king89} encompasses
\namecite{hpsg1} in its entirety, it does not provide a
characterization of all possible HPSG grammars, nor does it describe
the current formulation of the theory as expressed in
\namecite{hpsg2}.  A similar drawback can be found in
\cite{polmosh90}: while it gives a denotational semantics for a typed
feature structures system, it is not directed specifically towards
HPSG, and no formulation of the properties of HPSG grammars is given.

A different work is described in \cite{carp92}; a wide, concise
theory of the logic of typed feature structures is presented, with
many variations and applications. \namecite{carp92} serves as the
main reference point for any attempt to define such formalisms;
however, as it is not concentrated on HPSG per se, no formal
definition for it can be found there either.

Not only a denotational semantics for HPSG is required; operational
semantics of the formalism is missing, too.  While some compilers for
HPSG were developed (see section~\ref{comp-hpsg}), they all rely on
\cite{hpsg1} and \cite{hpsg2} as their source for interpreting the
formalism, and as we mentioned above, both references are not formal
enough. Since HPSG is not defined formally enough, we opted in this
work to implement ALE (see below), which is the most common platform
for designing HPSG grammars.

\subsection{Abstract Machine Techniques}
High-level programming languages, especially ones with dynamic
structures, have always been hard to develop compilers for. A common
technique for overcoming the problems involves the notion of an {\bf
abstract machine}. It is a machine that, on one hand, captures the
essentials of the high-level language in its architecture and its
instruction set, such that a compiler from the source language to the
(abstract) machine language becomes relatively simple to design. On
the other hand, the architecture must be simple enough for the machine
language to be easily interpretable by common, Von-Neumann machine
languages.  This attitude also enables the design of portable front
ends for the compilers: as the machine language is abstract, it can be
easily interpreted by different (concrete) machine languages.

The design of such an abstract architecture must be careful enough to
compromise the two, usually conflicting, requirements: the closer the
machine architecture is to common architectures, the harder it is to
develop compilers for it; and on the other hand, if such a machine is
too complex, then while a compiler for it is easier to produce, it
becomes more complicated to execute its language on normal
architectures.

Abstract machines were used for various kinds of languages: they date
back to the P-Code for Pascal. Starting from Landin's SECD
\cite{landin}, many compilers for functional languages were designed
this way. When logic programming languages appeared, such techniques
were applied to them as well.
While Prolog has gained a recognition as a practical implementation of
the idea of programming in logic, a method for interpreting the
declarative logical statements was needed for such an implementation
to be well-founded.  In 1983 David Warren designed an abstract machine
for the execution of Prolog, consisting of a memory architecture and a
set of instructions \cite{waren83,wam}.  Even though there were prior
attempts to construct both interpreters and compilers for Prolog, it
was the Warren Abstract Machine (WAM) that gave Prolog not only a
good, efficient compiler, but, perhaps more importantly, an elegant
operational semantics.

The WAM consists of an architecture of the machine, augmented by a
compiler from Prolog to the instruction set of the abstract machine.
The operational semantics of each instruction is defined using a
low-level language that can be trivially mapped to any ordinary
hardware.  In fact, there is even a formal verification of the
correctness of the WAM compiler \cite{russ92}.  The WAM captures in
an elegant way the substantial elements of Prolog.  First-order term
unification is supported by special data structures and instructions
of the machine architecture. Several instructions that deal with
control issues implement the backtracking mechanism.
  
The WAM immediately became the starting point for many compiler
designs for Prolog. The techniques it delineates serve not only for
Prolog proper, but also for constructing compilers for related
languages.  To list just a few examples, abstract machine techniques
were used for a parallel Prolog compiler \cite{herm86}, for variants
of Prolog that use different resolution methods \cite{oldt}, extend
Prolog with types \cite{beierle-meyer} or with record structures
\cite{smolka-treinen}, and for a general theorem prover
\cite{schumann}.  There have even been attempts to construct a
methodology for the design of abstract machines for logic programming
languages \cite{kursawe,nilson93}.

\subsection{Processing HPSG}
\label{comp-hpsg}
Linguistic formalisms provide means for describing the structure of
natural languages; they do not specify methods for determining whether
a given string is indeed a member of the language defined by a
grammar; nor do they prescribe ways for computing the structure that
the grammar assigns to the permissible strings. These tasks are
performed by {\em parsing\/} algorithms. Different parsing algorithms
exist for various classes of languages, both formal
(see~\cite{parsing} for a survey) and natural (see,
e.g.,~\cite{gazmel,sikkel,parsing-as-deduction}). In this work we
implement a simple {\em chart\/} parsing algorithm; such parsers were
first introduced by~\cite{kaplan73,kay73} and are widely used nowadays.

Various parsers for HPSG have been designed in the past, some of which
compile their input grammars into an executable program. The first
work is described in \cite{prupol85}; it is an implementation of a
very early version of HPSG. For instance, most of the features are
limited to accommodate only a small set of atomic values. Rules are
specified in a way reminiscent of GPSG rules. This work cannot be
considered as reflecting HPSG today.

Franz has implemented an HPSG parser in LISP \cite{franz}. This
parser was designed in accord with \namecite{hpsg1}, and doesn't cover the
modifications introduced by \namecite{hpsg2}. It is rather limited, for
example by allowing only tree-shaped type hierarchies to be defined --
no multiple inheritance is permitted.  While a specific HPSG grammar
for English is a part of this implementation, the system can be used
as a framework for developing different grammars. According to Franz's
reports, the parser is very slow, even when used on a limited grammar
and short inputs: example sentences were parsed in 12-65 seconds.

A different implementation is HPSG-PL \cite{popvog91,hpsg-pl}. This
system allows more complex type hierarchies to be defined; it enables
the definition of grammar rules, principles and lexical rules, and an
HPSG grammar for English is supplied, based on \namecite{hpsg1}. It
incorporates a chart parser where the parsing algorithm makes specific
use of some grammar features (e.g., HEAD-DTR), and thus the
stipulation of rules does not involve explicit phrase structure. The
grammar is compiled into a Prolog program where each feature structure
is transformed to a fixed-arity list. Yet the performance of the
parser is rather low: according to \cite{popvog91}, simple sentences
take 1-25 seconds of CPU time to parse.

Another system that was adapted for HPSG is Unicorn \cite{unicorn}.
Originated as a generalization of a Context-Free parser, this system
uses Shieber's extension to Earley's algorithm, thus enabling the
definition of various augmented context-free grammars. An HPSG grammar
was defined in terms of the Unicorn framework, with some divergence
from \namecite{hpsg2}, by \namecite{russell}.  The most important
aspect in which Unicorn differs from the above mentioned parsers is
that it doesn't incorporate a typing system for feature structures at
all. This system was not specifically designed with HPSG
implementation in mind, and the grammar was not intended to be
complete in any sense, as it was used only as part of a more complex
project. We have no reports on the results of this parser for HPSG.

It is important to note that HPSG falls naturally into the class of
general constraint systems, and thus the problem of providing the
correct structure for an input sentence can be naturally reduced to
the problem of solving a constraint system. Many general constraint
solvers have been developed recently that were used for linguistic
applications, including some for which HPSG grammars were designed.  A
typical representative is ALE \cite{ale}. Not being specifically
designed for HPSG, this system is a general Attribute Logic Engine
incorporating a chart parser with a formalism for specifying relations
among typed feature structures in a way that enables encoding of HPSG
grammars in a very natural manner. In fact, an HPSG grammar for
English has been constructed in this framework by
\namecite{penn} that covers most of \namecite{hpsg2}.
Compilation of ALE programs generates a rather efficient Prolog code.

A very similar project is Troll \cite{troll}. It is a framework for
processing typed feature structures, much in the same way as ALE does,
albeit with a slightly different underlying theory. As Troll is still
in preliminary phases, not much is reported regarding its use.
Another work, aiming at covering as many as possible of the extensions
to simple unification formalisms, is CUF \cite{cuf}. This system is
still under development. Two more general systems that were used for
developing HPSG grammars are TFS \cite{tfs}, which is a general
constraint solver, and PROFIT \cite{profit}, which simply compiles TFS
based specifications to Prolog.

\subsection{Computational Grammars for Hebrew}
\label{heb-hpsg}
The Hebrew language poses some interesting problems for the grammar
designer.  The Hebrew script\footnote{We refer here to the
non-vocalized script which is in everyday use, and not to the
vocalized script that is used for special purposes (such as poems or
children books) only.} is highly ambiguous, a fact
that results in many part-of-speech tags for almost every
word~\cite{ornan94}.  Another problem of the script is that short
prepositions, articles and conjunctions are usually attached to the
words that immediately succeed them, which makes it harder to parse
the input sentences.  In addition to these two features, the Hebrew
morphology is very rich. A noun base form might have over fifteen
different derivations, and a verb base form -- more than thirty. All
these call for some pre-processing of the input to the
parser. Disambiguation of the script, as well as morphological
analysis, were covered by different
works~\cite{ibm92,choueka95,ornan95}; some major decisions have to be
taken, including the representation of the Hebrew script and the treatment
of morphological analysis. As the current trend is to use
constraint-based formalisms for tasks other than syntax and semantics,
this is the approach we choose.

At the syntactic level, Hebrew exhibits a rather free constituent order,
although many constraints are placed on the order of words within
constituents. 
The use of agreement features in Hebrew is more extensive than, say,
English. For example, nouns and adjectives must agree on number,
gender and definiteness. Agreement checking becomes more complicated
in coordinated constructs (see \cite{shuly:master}) and much thought
must be given to the correct treatment of agreement.

There have been some attempts to provide a computational grammar for
(the syntax of) the Hebrew language. The first work was done by
\namecite{cohen84}, who had written a special software system for
performing both the morphological and the syntactic analysis of Hebrew
sentences. This work was very preliminary and its coverage was
limited. Another preliminary work is described in \cite{nirenburg}: it
is a small-scale ATN for Hebrew, capable of recognizing very limited
structures. A transformation-based grammar is suggested in~\cite{chendror}.

Unification-based formalisms were used for developing Hebrew grammars
only recently. A very limited experiment was done using PATR-II
\cite{shuly:patr} but was later extended
\cite{shuly:syntactic-analysis,shuly:jnle} to a reasonable subset of
the language, on a more convenient platform: Tomita's LR
Parser/Compiler, which is based on LFG. The grammar is capable of
recognizing sentences of rather wide variety and complexity, but
produces only the syntactic structures of the input sentences. See
\cite{shuly:master} for a detailed discussion. A different work along
the same lines is \cite{dana}: it uses the same framework but
concentrates on the syntax of NPs in Hebrew, employing ideas from
different linguistic theories.  All in all, no broad-coverage,
efficient, concise computational grammar for Hebrew exists.

\section{Achievements of the Thesis}
The main objective of this work was to formally define an operational
semantics for a unification-based grammar formalism, suitable for
specifying HPSG grammars, through the use of an abstract machine. To
this end we have first conducted a theoretical investigation into the
properties of such formalisms. 
The main contributions of this endeavor are:
\begin{itemize}
\item
Formalization and explication of the notion of multi-rooted feature
structures (MRSs) that are used implicitly in the computational linguistics
literature;
\item
Concise definitions of a TFS-based linguistic formalism, based on
abstract MRSs;
\item 
Algebraic specification of a {\em parsing step} operator, $T_{G,w}$,
that induces algebraic semantics for this formalism;
\item
Treatment of parsing as a model for computation, assigning operational
semantics to the linguistic formalism;
\item
Specification and correctness proofs for parsing in this framework;
\item
A new definition for {\em off-line parsability}, less
strict than the existing one, and termination proof for off-line
parsable grammars.
\end{itemize}
This more theoretical work was presented as \cite{shuly:iwpt95}. The
off-line parsability result is presented in \cite{shuly:jolli}.

Once the theoretical background was set, we have designed \amalia, an
abstract machine for unification-based grammars. This is the first
application of abstract machine techniques to a linguistic formalism.
The core engine of the machine was presented in \cite{shuly:nlulp-95},
and a more detailed presentation is in preparation.  The machine is
accompanied by a compiler from the ALE specification language to the
machine instructions, an interpreter for the machine instructions and
a debugger for machine language programs.  The abstract machine endows
natural language grammars with an operational meaning; furthermore,
its use results in highly efficient processing: the compiled grammars
are executed much faster than with the existing ALE processor. Some
tests we have conducted showed a speed-up of a factor of 20 in
compilation time, and a factor of 5-15 in execution time.

In order to test the machine on a realistic application, we have
developed a small-scale, HPSG-based grammar for a fragment of the
Hebrew language, using \amalia\ as the development platform. This is
the first application of HPSG to a Semitic language.

Another track of research we are exploring (with Evgeniy Gabrilovich)
is the adaptation of \amalia\ to perform natural language {\em
generation}, as opposed to parsing. Based upon the algorithm
of~\cite{samuelsson95}, a characterization of generation
with unification based grammars can be found in~\cite{gabr:thesis}. A separate
compiler is constructed, based upon \amalia's compiler, that
transforms a grammar to an inverted, normalized form, more
suited for generation. To execute the inverted grammar on \amalia,
very few modifications in the machine are needed. Once this project is
completed, \amalia\ will become a unified framework for processing
grammars, supporting both parsing and generation.

\section{Structure of this Document}
In chapter~\ref{sec:parsing} a theory of parsing with typed feature
structures is presented. We start in a survey of the theory of TFSs,
along the lines of \namecite{carp92}, but we extend it to multi-rooted
structures in section~\ref{mrs}. In particular, we discuss the
computational properties of TFS-based grammars and show in
section~\ref{olp} a condition on grammars that guarantees termination
of parsing. 

Chapter~\ref{sec:amalia} describes the abstract machine itself,
starting with its core, aimed at unifying two feature structures. In
section~\ref{machine-parsing} this engine is enveloped with control
structures to accommodate for {\em parsing}. We conclude this chapter with a
discussion of some implementation details
(section~\ref{sec:implementation}).

Chapter~\ref{sec:hebrew} describes the HPSG-based grammar for Hebrew.
Conclusions and suggestions for further research are given in
chapter~\ref{sec:summary}. 
Appendix~\ref{inst-list} lists all the machine instructions, and the
Hebrew grammar is listed in appendix~\ref{app:grammar}.

\chapter{Parsing with Typed Feature Structures}
\label{sec:parsing}
This chapter provides the theoretical background for the design of the
machine: we discuss below the theory of typed feature structures and
the details of parsing with grammars that are based upon them.
Section~\ref{basics} outlines the theory of TFSs of
\cite{carp91,carp92}. We repeat it here in order to make this document
as self contained as possible. However, the well-foundedness result
(section~\ref{sec:well-founded}) is an original contribution.  We
deviate from the presentation of \namecite{carp92} in
section~\ref{sec:afs}, where we emphasize {\em abstract typed feature
structures} (AFSs). Encoding the essential information of TFSs, AFSs
were introduced by \namecite{moshier} but we use a different
presentation that is suited for {\em typed\/} feature structures.
Unification is defined over AFSs rather than TFSs. Section~\ref{mrs}
introduces an explicit construct of {\em multi-rooted feature
structures} (MRSs) that naturally extend TFSs, used to represent
phrasal signs as well as grammar rules. Abstraction is extended to
MRSs and the mathematical foundations needed for manipulating them is
given. The concepts of grammars and the languages they generate are
formally defined in section~\ref{rules-and-grammars}, and the
TFS-based formalism is thus acquired a denotational semantics. In
section~\ref{parsing} a model for computation, corresponding to
bottom-up chart parsing for the formalism, is presented. The TFS-based
formalism is thus endowed with an {\em operational\/} semantics.  Next,
we prove that both semantics coincide. Finally, we discuss the class
of grammars for which computations terminate. We give a more relaxed
definition for off-line parsability and prove that termination is
guaranteed for off-line parsable grammars. The presentation is
accompanied by a running example of a grammar and the parsing process
it induces.

\section{Theory of Feature Structures}
\label{basics}
\label{types-and-fs}
The first part of this section summarizes some preliminary notions
along the lines of~\cite{carp92}.
For the following discussion we fix
non-empty, finite, disjoint sets
\types\ and \feats\ of types and feature names, respectively.
We assume that the set \feats\ is totally ordered.
We also fix an infinite set \nodes\ of nodes, disjoint of \types\ and
\feats, each member of which is decorated
by a type from \types\ through a fixed typing function $\theta: \nodes
\onto \types$. The set \nodes\ is `rich' in the sense that
for every $t \in \types$, the set $\{q \in \nodes \mid \theta(q) =
t\}$ is infinite.

Below, the metavariable $T$ ranges over subsets of types, $t$ -- over
types, $f$ -- over features and $q$ -- over nodes.
When dealing with partial functions the
notation `$F(x)\,\,\isdef$' means that $F$ is defined for the value $x$ 
and the symbol `$\,\,\isndef$' means undefinedness.
Whenever the result of an application of a partial function is used as
an operand, it is meant that the function is defined for its arguments.

\begin{definition}[Type hierarchy]
A partial order
$\subsumes$ over $\types \cross \types$ is a {\bf type hierarchy} (or
{\bf inheritance hierarchy}) if it is bounded complete, i.e., if every
up-bounded subset 
$T$ of \types\ has a (unique) least upper bound, $\unif T$, 
referred to as the {\bf unification} of the types in $T$.

If $t_1 \subsumes t_2$ we say that $t_1$ {\bf subsumes}, or is {\bf
more general than}, $t_2$; $t_2$ is a {\bf subtype} of (more {\bf
specific} than) $t_1$.

Let $\bot = \unif \emptyset$ be the most general type.  Let the most
specific type be $\top = \unif \types$. If $\unif T = \top$ we say
that $T$ is {\bf inconsistent}. 
Let $\meet T = \unif
\{t' \mid t' \subsumes t$ for every $t \in T\}$ be the greatest lower
bound of the set $T$.
\end{definition} 

\begin{definition}[Feature structures]
A {\bf (typed) feature structure} (TFS) is
a directed, connected, labeled graph consisting of a finite, nonempty
set of nodes $Q \subseteq \nodes$, a root
$\qbar \in Q$, and a partial function $\delta : Q \cross \feats\ \onto
Q$ specifying the arcs such that every node $q \in Q$ is accessible
from $\qbar$.
\end{definition} 
The nodes of a feature structure are thus labeled by types while the
arcs are labeled by feature names.  The root $\qbar$ is a
distinguished node from which all other nodes are reachable.  A
feature structure is of type $t$ when $\theta(\qbar) = t$.  When we
say that a feature structure $A$ {\em exists} we mean that no node of
$A$ is typed $\top$.

\begin{figure}[hbt]
\center
\begin{minipage}[t]{8cm}
\center
Graph representation:\\
\fbox{
\psfig{figure=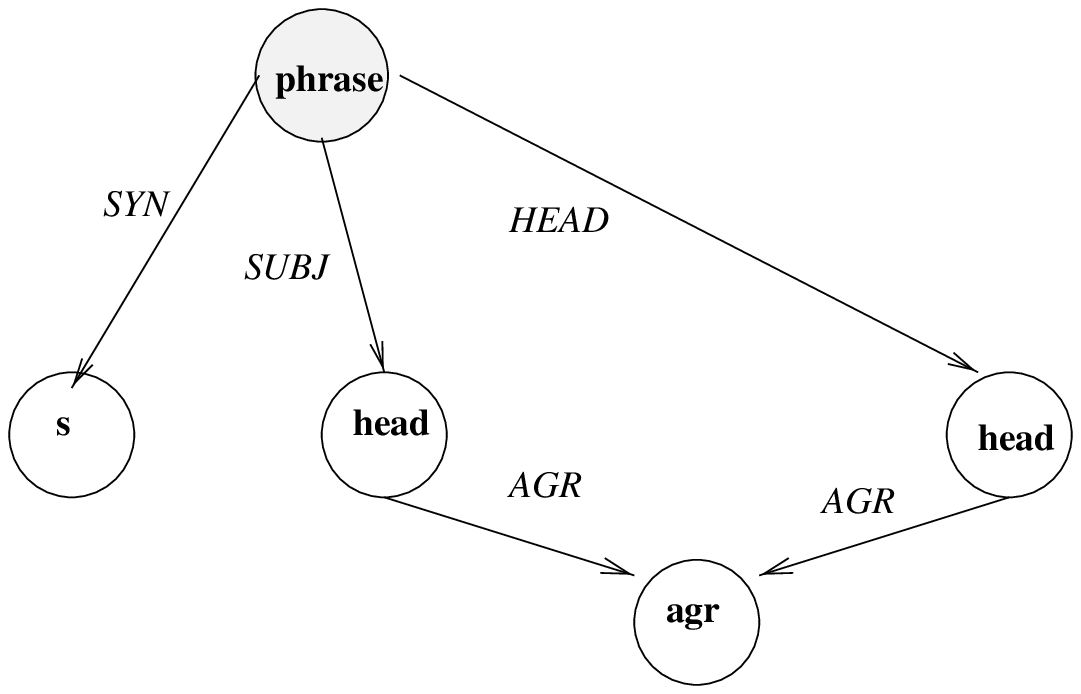,height=4cm}
}
\end{minipage} \ 
\begin{minipage}[t]{6cm}
\center
AVM representation:
\[
\begin{tfs}{phrase}
	SYN:	& \begin{tfs}{s} \end{tfs} \\
	SUBJ:	& \begin{tfs}{head}
			AGR: 	& \tag{3}\begin{tfs}{agr} \end{tfs}
			\end{tfs}  \\
	HEAD:	& \begin{tfs}{head}
			AGR:	& \tag{3}
			\end{tfs} 
\end{tfs}
\]
\end{minipage}
\mycaption{A feature structure}{תוינוכת הנבמ}
\label{fig:fs}
\end{figure}

Let \fs\ be the collection of all feature structures over the given 
\feats\ and \types.
We use upper-case letters (with or without tags, subscripts etc.) to
refer to feature structures. We use $Q, \qbar, \delta$ (with the same
tags or subscripts) to refer to constituents of feature structures.
Figure~\ref{fig:fs} depicts an example feature structure, represented
both as an Attribute-Value Matrix (AVM) and as a graph.

Note that all feature structures are, by definition, graphs. Some
grammatical formalisms used to have a special kind of feature
structures, namely {\em atoms}; atoms are represented in our framework
as nodes with no outgoing edges. For a discussion regarding the
implications of such an approach, refer to \cite[Chapter 8]{carp92}.

\begin{definition}[Appropriateness]
\label{approp-spec}
An {\bf appropriateness specification} over the type 
hierarchy and the set \feats\ is a partial function
$Approp : \feats \cross \types \onto \types$,
such that: 
\begin{itemize}
\item
let $T_f = \{t \in \types \mid Approp(f,t) \isdef\}$; then for every $f
\in \feats$, $T_f \neq \emptyset$ and $\meet T_f \in T_f$.
\item
if $Approp(f,t_1) \isdef$ and $t_1 \subsumes t_2$ then
$Approp(f,t_2) \isdef$ and $Approp(f,t_1) \subsumes Approp(f,t_2)$.
\end{itemize}
\end{definition} 
i.e., every feature is introduced by some most general type, and is
appropriate for all its subtypes; 
and if the appropriate type for a feature in $t_1$ is some type $t$,
then the appropriate type of the same feature in $t_2$, which is a
subtype of $t_1$, must be at least as specific as $t$.

If $Approp(f,t) \isdef$ we say that $f$ is appropriate for $t$ and 
that $Approp(f,t)$ is the appropriate type for the feature f
in the type t.
The set of features appropriate for some type is
ordered (since \feats\ is ordered).

\begin{definition}[Well-typed feature structures]
A feature structure $(Q,\qbar,\delta)$ is {\bf
well typed} iff 
for every $q \in Q, \theta(q) \neq \top$ and
for all $f \in \feats$ and $q \in Q$, if $\delta (q,f)\isdef$
then $Approp (f,\theta (q))\isdef$ and
$Approp(f,\theta(q))$ $\subsumes \theta(\delta(q,f))$.
\end{definition} 
i.e., if an arc
labeled $f$ connects two nodes, then $f$ is appropriate for the type
of the source node; and the 
appropriate type for $f$ in the type of the source node
subsumes the type of target node.

\begin{definition}[Total well-typedness]
A feature structure is {\bf totally well-typed} iff it is well typed
and for all $f \in \feats$ and $q \in Q$, if
$Approp(f, \theta(q)) \isdef$ then $\delta(q,f) \isdef$.
\end{definition} 
i.e., every feature which is appropriate for the type labeling 
some node labels an outgoing arc from that node.

\begin{sloppypar}
\begin{definition}[Appropriateness loops]
\label{approp-loops}
The appropriateness specification contains a {\bf loop} if there 
exist $t_1, t_2, \ldots, t_n \in \types$ such that for every i, $1 \le
i \le n$, there is a feature $f_i \in \feats$ such that
$Approp(f_i, t_i) = t_{i+1}$, where $t_{n+1} = t_1$.
\end{definition} 
\end{sloppypar}

\begin{definition}[Paths]
A {\bf path} is a finite sequence of feature names, and the set
$\paths\ = \feats ^{*}$ is the collection of paths. 
We use $\pi, \alpha$ (with or without subscripts) to refer to paths.
$\epsilon$ is the empty path.
The definition of $\delta$ is
extended to paths in the natural way:
\[
\begin{array}{l}
\delta (q,\epsilon) = q \;  \\
\delta (q, f \pi) = \delta (\delta (q,f), \pi) 
\end{array}
\] 
The paths of a feature
structure $A$ are $\Pi(A)=\{\pi\mid\pi\in\paths$ and
$\delta(\qbar_A,\pi)\isdef\}$.
\end{definition} 

\begin{definition}[Cycles]
A feature structure $A=(Q,\qbar,\delta)$ is {\bf cyclic} if there
exist a non-empty path $\alpha\in\paths$ and a node $q \in Q$ such that
$\delta(q,\alpha)=q$. It is {\bf acyclic} otherwise.
\end{definition}

\begin{sloppypar}
\begin{definition}[Path values]
The {\bf value} of a path $\pi$ in a feature structure $A = (Q, \qbar,
\delta)$, denoted by $val(A, \pi)$, is {\bf non-trivial} if and only 
if $\delta (\qbar, \pi) \isdef$, in which
case it is a feature structure $A' = (Q', \qbar ', \delta ')$, where:
\begin{itemize}
\item
$\qbar' = \delta(\qbar,\pi)$
\item
$Q' = \{q' \mid \mbox{there exists a path $\pi'$ such that
$\delta(\qbar',\pi') = q'$} \}$ ($Q'$ is the set of nodes reachable
from $\qbar'$)
\item
for every feature $f$ and for every $q' \in Q'$, $\delta'(q',f) =
\delta(q',f)$ ($\delta'$ is the restriction of $\delta$ to $Q'$)
\end{itemize}
If $\delta(\qbar,\pi) \isndef$, $val(A,\pi)$ is defined to be a
single node whose type is $\top$.
\end{definition} 
\end{sloppypar}

\begin{definition}[Reentrancy]
A feature structure $A$ is {\bf reentrant} iff there exist two
different paths
$\pi_1, \pi_2$ such that $\delta(\qbar,\pi_1) = \delta(\qbar, \pi_2)$.
In this case the two paths are said to share the same value.
\end{definition} 

\section{Subsumption}
\begin{definition}[Subsumption]
$A_1 = (Q_1, \qbar_1, \delta_1)$ {\bf subsumes} $A_2 = (Q_2, \qbar_2,
\delta_2)$ (denoted by $A_1 \subsumes A_2$)
iff there exists a total function $h : Q_1 \onto Q_2$, called a {\bf
subsumption morphism}, such that
\begin{itemize}
\item
$h(\qbar_1) = \qbar_2$
\item
for every $q \in Q_1$, $\theta(q) \subsumes \theta(h(q))$
\item
for every $q \in Q_1$ and for every $f$ such that $\delta_1(q,f)
\isdef$, $h(\delta_1(q,f)) = \delta_2(h(q),f)$.
\end{itemize}
$A_1 \sqsubset A_2$ iff $A_1 \subsumes A_2$ and $A_1 \neq A_2$.
\end{definition} 
$h$ associates with every node in $Q_1$ a node in $Q_2$ with at least
as specific a type; moreover, if an arc labeled $f$ connects $q$ with $q'$,
then such an arc connects $h(q)$ with $h(q')$. If $A\sqsubseteq B$
then every path defined in $A$ is defined in $B$, and if two paths are
reentrant in $A$ they are reentrant in $B$.
\begin{lemma}
\label{lemma:subset}
If $A\sqsubseteq B$ then $\Pi(A)\subseteq\Pi(B)$.
\end{lemma}
\begin{lemma}
\label{lemma:reent}
If $A\sqsubseteq B$ then for every $\pi_1,\pi_2\in\Pi(A)$, 
$\delta_A(\qbar_A,\pi_1)=\delta_A(\qbar_A,\pi_2)$ implies that
$\delta_B(\qbar_B,\pi_1)=\delta_B(\qbar_B,\pi_2)$.
\end{lemma}


\section{Well-Foundedness of Subsumption}
\label{sec:well-founded}
\begin{definition}
A partial order $\succ$ on $D$ is {\bf well-founded} iff
there exists no infinite decreasing sequence $d_0 \succ d_1 \succ d_2
\succ \ldots$ of elements of $D$.
\end{definition}
We prove below that subsumption of TFSs is well-founded iff they are
acyclic.

\begin{lemma}
\label{lemma:finite}
A TFS $A$ is cyclic iff $\Pi(A)$ is infinite.
\end{lemma}
\proof{}
If $A$ is cyclic, there exist a node $q\in Q$ and a non-empty path
$\alpha$ that $\delta(q,\alpha)=q$. Let $\pi$ be such that
$q=\delta(\qbar,\pi)$, then the infinite set of paths
$\{\pi\alpha^{i}\mid i \ge 0\}$ is contained in $\Pi(A)$. If $\Pi(A)$
is infinite then since $Q$ is finite, there exists a node $q\in Q$
that $\delta(q,\pi_i)\isdef$ for an infinite number of different paths
$\pi_i$. Since \feats\ is finite, the out-degree of every node in $Q$
is finite; hence $q$ must be part of a cycle.

\begin{definition}[Rank]
Let $r:\types\rightarrow\ns$ be a total function such that
$r(t)<r(t')$ if $t \sqsubset t'$. For an acyclic TFS $A$, let
$\Delta(A)=|\Pi(A)|-|Q_A|$ and let $\Theta(A)=\sum_{\pi\in\Pi(A)}
r(\theta(\delta(\qbar,\pi)))$. Define a rank for acyclic TFSs:
$rank(A)=\Delta(A)+\Theta(A)$.
\end{definition} 
By lemma~\ref{lemma:finite}, $rank$ is well defined for acyclic TFSs.
$\Delta(A)$ can be thought of as the number of reentrancies in $A$, or
the number of different paths that lead to the same node in $A$. For
every acyclic TFS $A$, $\Delta(A)\ge 0$ and hence $rank(A)\ge 0$.

\begin{lemma}
\label{lemma:rank}
If $A \sqsubset B$ and both are acyclic then $rank(A)<rank(B)$.
\end{lemma}
\proof{}
Assume $A \sqsubset B$ and both are acyclic; hence by
lemma~\ref{lemma:subset}, $\Pi(A)\subseteq\Pi(B)$ and by
lemma~\ref{lemma:finite} both are finite. Let $h:Q_A\rightarrow Q_B$
be a subsumption morphism.
\begin{enumerate}
\item
If $\Pi(A)=\Pi(B)$ then $|\Pi(A)|=|\Pi(B)|$. $A\sqsubset B$, hence
either there exists a node $q\in Q_A$ that
$\theta(q)\sqsubset\theta(h(q))$, and hence $\Theta(A)<\Theta(B)$
(while $\Delta(A)\le\Delta(B)$); or (by lemma~\ref{lemma:reent}) there
exist two paths $\pi_1,\pi_2$ that
$\delta_A(\qbar_A,\pi_1)\neq\delta_A(\qbar_A,\pi_2)$, but
$\delta_B(\qbar_B,\pi_1)=\delta_B(\qbar_B,\pi_2)$, in which case
$\Delta(A)<\Delta(B)$ (while $\Theta(A)\le\Theta(B)$). In any case,
$rank(A)<rank(B)$.
\item
If $\Pi(A)\subset\Pi(B)$ then $|\Pi(A)|<|\Pi(B)|$; as above,
$\Theta(A)<\Theta(B)$. However, it might be the case that
$|Q_A|<|Q_B|$. But for every node $q\in Q_B$ that is
not the image of any node in $Q_A$, there exists a path $\pi$ such
that $\delta(\qbar_B,\pi)=q$ and $\pi\not\in\Pi(A)$. Hence
$|\Pi(A)|-|Q_A|\le |\Pi(B)|-|Q_B|$, and $rank(A)<rank(B)$.
\end{enumerate}

\begin{figure}
\center
\psfig{figure=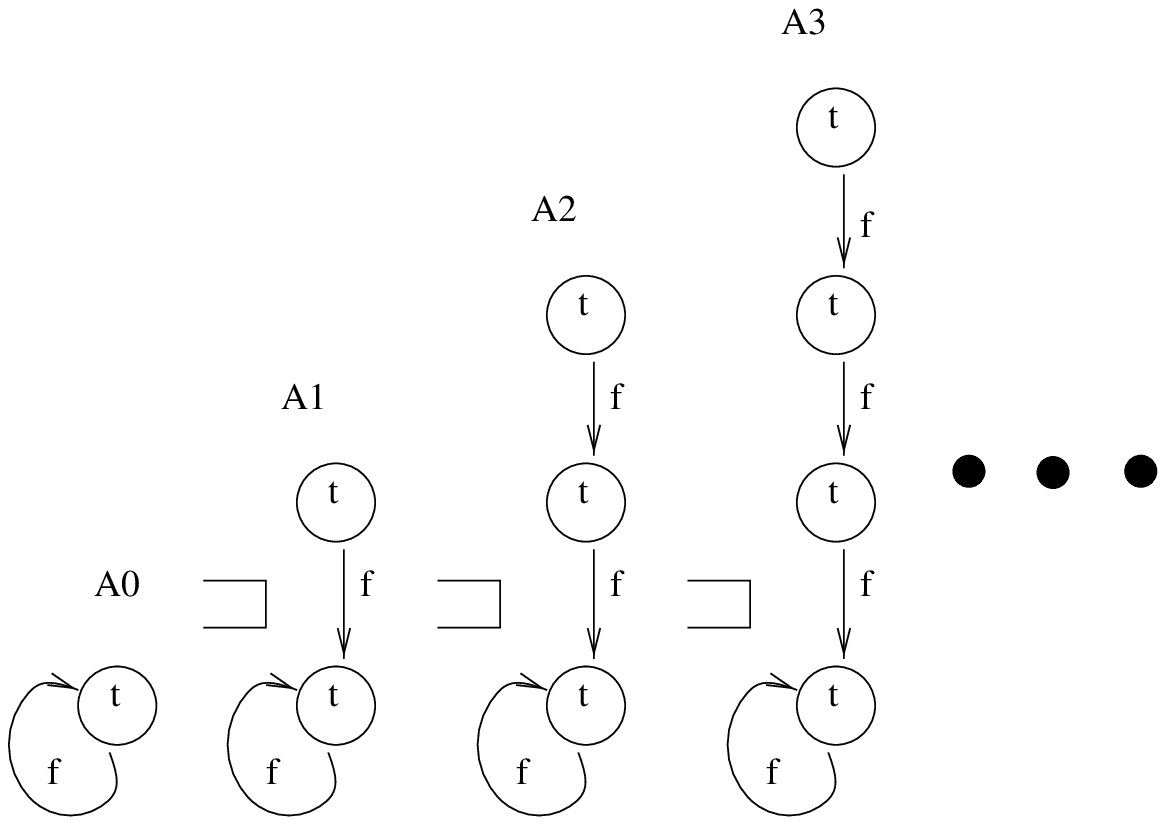}
\mycaption{An infinite decreasing sequence of TFSs}{תוינוכת ינבמ לש תדרוי תיפוסניא הרדס}
\label{fig:sequence}
\end{figure}

\begin{theorem}
Subsumption of TFSs is not well-founded.
\end{theorem}
\proof{}
Consider the infinite sequence of TFSs $A_0,A_1,\ldots$ depicted
graphically in figure~\ref{fig:sequence}.  For every $i \ge 0$, $A_i
\sqsupset A_{i+1}$: to see that consider the morphism $h$ that maps
$\qbar_{i+1}$ to $\qbar_{i}$ and $\delta_{i+1}(q,f)$ to
$\delta_i(h(q),f)$ (i.e., the first $i+1$ nodes of $A_{i+1}$ are mapped
to the first $i+1$ nodes of $A_i$, and the additional node of $A_{i+1}$
is mapped to the last node of $A_i$). Thus there exists a decreasing
infinite sequence of cyclic TFSs and subsumption is not well-founded.

\begin{theorem}
Subsumption of acyclic TFSs is well-founded.
\end{theorem}
\proof{}
For every acyclic TFS $A$, $rank(A)$ is finite and $rank(A) \ge 0$. By
lemma~\ref{lemma:rank}, if $A \sqsubset B$ then $rank(A)<rank(B)$. If
an infinite decreasing sequence of acyclic TFSs existed, $rank$ would
have mapped them to an infinite decreasing subsequence of $\ns$, which
is a contradiction. Hence subsumption is well-founded.

\section{Abstract Feature Structures}
\label{sec:afs}
The essential properties of a feature structure, excluding the
identities of its nodes, can be captured by three components: the set
of paths, the type that is assigned to every path, and the sets of
paths that lead to the same node.  In this section we elaborate on
ideas presented in \cite{moshier-rounds,moshier}; in contrast to the approach
pursued in \cite{carp92}, we first define abstract feature structures
and then show their relation to concrete ones. The representation of
graphs as sets of paths is inspired by works on the semantics of
concurrent programming languages, and the notion of fusion-closure is
due to \cite{emerson}.

\begin{definition}[Alphabetic variants]
Two feature structures $A$ and $B$ are {\bf alphabetic
variants} ($A \sim B$) iff $A \subsumes
B$ and $B \subsumes A$.
\end{definition} 
Alphabetic variants have exactly the same structure, and corresponding
nodes have the same types. Only the identities of the nodes
distinguish them.

\begin{definition}[Abstract feature structures]
A {\bf pre- abstract feature structure} (pre-AFS) is a triple
$\langle\Pi,\Theta,\approx\rangle$, where
\begin{itemize}
\item
$\Pi \subseteq \paths$ is a non-empty set of paths
\item
$\Theta: \Pi \onto \types$ is a total function, assigning a type to
every path
\item
$\approx \subseteq \Pi \times \Pi$ is a relation specifying reentrancy.
\end{itemize}
An {\bf abstract feature structure} (AFS) is a pre-AFS $A$
for which the following requirements hold:
\begin{itemize}
\item
$\Pi$ is prefix-closed: if $\pi\alpha \in \Pi$ then $\pi \in \Pi$
(where $\pi,\alpha \in \paths$)
\item
$A$ is fusion-closed: if $\pi\alpha \in \Pi$ and 
$\pi'\alpha' \in 
\Pi$ and $\pi \approx \pi'$ then $\pi\alpha' \in \Pi$ (as well as
$\pi'\alpha \in \Pi$) and
$\pi\alpha' \approx \pi'\alpha'$ (as well as $\pi'\alpha \approx \pi\alpha$)
\item
$\approx$ is an equivalence relation with a finite index
(with $[\approx]$ the set of its equivalence classes)
\item
$\Theta$ respects the equivalence: if $\pi_1 \approx \pi_2$ then
$\Theta(\pi_1) = \Theta(\pi_2)$.
\end{itemize}
\end{definition} 
An AFS $\langle \Pi, \Theta, \approx \rangle$ is {\em well-typed} if
$\Theta(\pi) \neq \top$ for every $\pi \in \Pi$ and if $\pi f \in \Pi$
then $Approp(f,\Theta(\pi)) \isdef$ and $Approp(f,\Theta(\pi))
\subsumes \Theta(\pi f)$.
It is {\em totally well typed} if, in addition, for every $\pi \in \Pi$, if
$Approp(f,\Theta(\pi))\isdef$ then $\pi f \in \Pi$.

Abstract features structures can be related to concrete ones in a
natural way:
If $A = (Q,\qbar,\delta)$ is a TFS then $Abs(A) = \langle \Pi_A, \Theta_A,
\approx_A \rangle$ is defined by:
\begin{itemize}
\item
$\Pi_A = \{\pi\mid\delta(\qbar,\pi)\isdef\}$
\item
$\Theta_A(\pi) = \theta(\delta(\qbar,\pi))$
\item 
$\pi_1
\approx_A \pi_2$ iff $\delta(\qbar,\pi_1) = \delta(\qbar,\pi_2)$.
\end{itemize}

\begin{lemma}
If $A$ is a feature structure then $Abs(A)$ is an abstract feature
structure.
\end{lemma}
\proof{}
\begin{enumerate}
\item
$\Pi$ is prefix-closed: $\Pi = \{ \pi \mid \delta(\qbar,\pi) \isdef
\}$. If $\pi \alpha \in \Pi$ then $\delta(\qbar,\pi \alpha) \isdef$
and by the definition of $\delta$, $\delta(\qbar,\pi) \isdef$, too.
\item
$Abs(A)$ is fusion-closed: Suppose that $\pi \alpha \in \Pi, \pi'
\alpha' \in \Pi$ and $\pi \approx \pi'$. Then $\delta(\qbar,\pi) =
\delta(\qbar,\pi')$. Hence $\delta(\qbar,\pi \alpha') \isdef$
(therefore $\pi \alpha' \in \Pi$), and $\delta(\qbar, \pi \alpha') =
\delta(\pi' \alpha')$, therefore $\pi \alpha' \approx \pi' \alpha'$.
In the same way, $\pi' \alpha \in \Pi$ and $\pi' \alpha \approx \pi'
\alpha'$.
\item
$\approx$ is an equivalence relation with a finite index: $\pi_1
\approx \pi_2$ iff $\delta(\qbar,\pi_1) = \delta(\qbar,\pi_2)$, namely
iff $\pi_1$ and $\pi_2$ lead to the same node (from $\qbar$) in $A$.
Hence $\approx$ is an equivalence relation and since $Q$ is finite,
$\approx$ has a finite index.
\item
$\Theta$ respects the equivalence: $\Theta(\pi) =
\theta(\delta(\qbar,\pi))$ and if $\pi_1 \approx \pi_2$ then
$\delta(\qbar,\pi_1) = \delta(\qbar,\pi_2)$, hence $\Theta(\pi_1) =
\Theta(\pi_2)$.
\end{enumerate}

For the reverse direction, consider an AFS $A = \langle \Pi, \Theta,
\approx \rangle$. 
First construct a `pseudo-TFS', $Conc(A) =
(Q,\qbar,\delta)$, that differs from a TFS
only in that its nodes are not drawn from the set \nodes.
Let $Q = \{q_{[\pi]} \mid [\pi] \in [\approx]\}$, making use of the
fact that `$\approx$' is of finite index.
Let $\theta(q_{[\pi]}) = \Theta(\pi)$ for every
node -- since $A$ is an AFS, $\Theta$ respects the equivalence
and therefore $\theta$ is representative-independent.
Let $\qbar = q_{[\epsilon]}$ and $\delta(q_{[\pi]},f) = q_{[\pi f]}$ for
every node $q_{[\pi]}$ and feature $f$. 
Since $A$ is fusion-closed, $\delta$ is representative-independent.
By injecting $Q$ into \nodes,
making use of the richness of \nodes, a concrete TFS $Conc(A)$ is
obtained, representing the equivalence class of alphabetic variants
that can be obtained that way. 
We abuse the notation $Conc(A)$ in the
sequel to refer to this set of alphabetic variants.
Figure~\ref{fig:afs} depicts the abstraction of the example feature
structure of figure~\ref{fig:fs}.
\begin{figure}[hbt]
\begin{center}
$\Pi = \{ \epsilon$, SYN, SUBJ, SUBJ AGR, HEAD, HEAD AGR $\}$ \\
\vspace{0.2cm}
\begin{tabular}{||l|c|c|c|c|c|c||c||}\hline
$\approx$  & $\epsilon$ & SYN & SUBJ & SUBJ AGR & HEAD & HEAD AGR & $\Theta$ \\ \hline
$\epsilon$ & + &    &     &     &      &      & {\bf phrase} \\ \hline
SYN        &   & +  &     &     &      &      & {\bf s} \\ \hline
SUBJ       &   &    & +   &     &      &      & {\bf head} \\ \hline
SUBJ AGR   &   &    &     & +   &      &   +  & {\bf agr} \\ \hline
HEAD       &   &    &     &     &  +   &      & {\bf head} \\ \hline
HEAD AGR   &   &    &     & +   &      &   +  & {\bf agr} \\ \hline
\end{tabular}
\end{center}
\mycaption{An abstract feature structure}{יטקרטסבא תוינוכת הנבמ}
\label{fig:afs}
\end{figure}

\begin{theorem}
If $A' \in Conc(A)$ then $Abs(A') = A$.
\end{theorem}
\proof{}
Let $A = \langle \Pi_A, \Theta_A, \approx_A \rangle, A' = (Q, \qbar,
\delta), Abs(A') = \langle \Pi, \Theta, \approx \rangle$. If $A' \in
Conc(A)$ then $Q$ can be mapped by a one-to-one function to the set of
equivalence classes of
$\approx_A$ and $\delta$ determines the paths in $\Pi_A$. By the
definition of $Abs, \Pi = \Pi_A$. Given a path $\pi \in \Pi_A,
\Theta(\pi) = \theta(\delta(\qbar,\pi)) = \theta(q_{[\pi]}) =
\Theta_A(\pi)$. If $\pi_1 \approx_A \pi_2$ then $\delta(\qbar,\pi_1) =
\delta(\qbar,\pi_2)$ (since $A$ is fusion-closed) and hence $\pi_1
\approx \pi_2$.

AFSs can be partially ordered:
$\langle \Pi_A, \Theta_A, \approx_A \rangle \preceq \langle \Pi_B,
\Theta_B, \approx_B \rangle$ iff $\Pi_A \subseteq \Pi_B, \approx_A
\subseteq \approx_B$ and for every $\pi \in \Pi_A, \Theta_A(\pi)
\subsumes \Theta_B(\pi)$.
This order corresponds to the subsumption ordering on TFSs, as the
following theorems show.
\begin{theorem}
\label{abstraction}
$A \subsumes B$ iff $Abs(A) \preceq Abs(B)$.
\end{theorem}
\proof{}
Let $Abs(A) = \langle \Pi_A,\Theta_A,\approx_A \rangle, Abs(B) =
\langle \Pi_B,\Theta_B,\approx_B \rangle$. Assume that $A \subsumes
B$, that is, a subsumption morphism $h: Q_A \onto Q_B$ exists.
If $\pi \in \Pi_A$ then (from the definition of $Abs(A)$)
$\delta_A(\qbar_A,\pi) \isdef$, that is, there exists a sequence $q_0,
q_1, \ldots, q_n$ of nodes and a sequence $f_1,\ldots,f_n$ of features
such that for every $i$, $0 \le i < n, \delta_A(q_i,f_{i+1}) = q_{i+1}$,
$q_0 = \qbar_A$ and $\pi = f_1 \cdots f_n$.
Due to the subsumption morphism, there exists a sequence of nodes
$h(q_0), \ldots, h(q_n)$ such that $\delta_B(h(q_i),f_{i+1}) =
h(q_{i+1})$ for every $i$, $0 \le i < n$, and $h(q_0) = \qbar_B$. Hence
$\pi \in \Pi_B$.
Moreover, since $A \subsumes B$, for every node $q$, $\theta(q)
\subsumes \theta(h(q))$. In particular, $\theta(q_n) \subsumes
\theta(h(q_n))$ and thus $\Theta_A(\pi) \subsumes \Theta_B(\pi)$. Now
suppose that two paths $\pi_1,\pi_2$ are reentrant in $A$. By the
definition of subsumption, $\pi_1$ and $\pi_2$ are reentrant in $B$,
too. Therefore $\approx_A \subseteq \approx_B$.\\
If $Abs(A) \preceq Abs(B)$, construct a function $h: Q_A \onto Q_B$ such
that $h(\qbar_A) = \qbar_B$ and for every $q \in Q_A, h(\delta_A(q,f)) =
\delta_B(h(q),f)$. Trivially, $h$ is total and $h(\qbar_A) = \qbar_B$.
Also, if $\delta_A(q,f) \isdef$ then $h(\delta_A(q,f)) =
\delta_B(h(q),f)$. To show that $\theta(q) \subsumes \theta(h(q))$
for every $q$, consider a path $\pi$ leading from $\qbar_A$ to $q$.
Since $Abs(A) \preceq Abs(B), \Theta_A(\pi) \subsumes \Theta_B(\pi)$ and
hence $\theta(q) \subsumes \theta(h(q))$. Hence $h$ is a subsumption
morphism.

\begin{theorem}
\label{concretion}
For every $A \in Conc(A'), B \in Conc(B'), A \subsumes B$ iff $A'
\preceq B'$.
\end{theorem}
\proof{}
Select some $A \in Conc(A'), B \in Conc(B')$. 
If $A \subsumes B$ then, by
theorem~\ref{abstraction}, $Abs(A) \preceq Abs(B)$. By the definition
of $Conc$, $Abs(A) = A'$ and $Abs(B) = B'$, so that $A' \preceq B'$.\\
If $A' \preceq B'$, construct a function $h:Q_A \onto Q_B$ as
follows:
First, let $h(\qbar_A) = \qbar_B$. Then, perform a depth-first search
on the graph $A$ and for every node $q' = \delta_A(q,f)$ encountered,
if $h(q') \isndef$ set $h(q') = \delta_B(h(q),f)$.
The order of the search is irrelevant: since $A' \preceq B',
\approx_{A'} \subseteq \approx_{B'}$ and therefore if $\pi_1
\approx_{A'} \pi_2$ then $\pi_1 \approx_{B'} \pi_2$.
Since $A' \preceq B', \Pi_{A'} \subseteq \Pi_{B'}$
and hence $\delta_B(h(q),f)$ is defined whenever $\delta_A(q,f)$ is
defined. Hence $h$ is total and $h(\qbar_A) = \qbar_B$. For every node
$q \in Q_A$, some path $\pi$ exists that leads from $\qbar_A$ to $q$
and from $\qbar_B$ to $h(q)$. $\Theta_A(\pi) \subsumes \Theta_B(\pi)$,
and therefore $\theta(q) \subsumes \theta(h(q))$. Hence $h$ is a
subsumption morphism.

\begin{corollary}
$A \sim B$ iff $Abs(A) = Abs(B)$.
\end{corollary}
{\bf Proof:} Immediate from theorem~\ref{abstraction}.

\begin{corollary}
$Conc(A) \sim Conc(B)$ iff $A = B$.
\end{corollary}
{\bf Proof:} Immediate from theorem~\ref{concretion}.

\section{Unification}
\label{unification}
As there exists a one to one correspondence between abstract feature
structures and (alphabetic variants of) concrete ones, we define
unification directly over AFSs. 
This leads to a simpler definition that captures the essence of the
operation better than the traditional definition.
We use the term `unification' to refer to both the
operation and its result.

\begin{lemma}
If $A=\afs{A}$ is a pre-AFS then there exists a pre-AFS $B = \afs{B}$
such that $B$ is the least extension of $A$ to a fusion-closed
structure and $\Theta_B(\pi) = \Theta_A(\pi)$ for every $\pi \in \Pi_A$.
\end{lemma}

\begin{lemma}
If $A=\afs{A}$ is a pre-AFS then there exists a pre-AFS $B=\afs{B}$
such that $\Pi_A = \Pi_B, \Theta_A = \Theta_B$ and $\approx_B$ is
the least extension of $\approx_A$ to an equivalence relation.
\end{lemma}

\begin{sloppypar}
\begin{definition}[Closure operations]
Let $Cl$ be a fusion-closure operation on pre-AFSs:
$Cl(A) = A'$, where $A'$ is the least extension of $A$ to a
fusion-closed structure. Let $Eq(\langle \Pi,\Theta,\approx
\rangle) = \langle \Pi,\Theta,\approx'\rangle)$ where $\approx'$ is
the least extension of $\approx$ to an equivalence relation.
Let $Ty(\langle \Pi,\Theta,\approx \rangle) = \langle
\Pi,\Theta',\approx \rangle$ where $\Theta'(\pi) = \bigsqcup_{\pi'
\approx \pi} \Theta(\pi)$.
\end{definition} 
\end{sloppypar}

\begin{definition}[Unification]
\label{def:unification}
The unification $A \unif B$ of two AFSs $A = \langle \Pi_A, \Theta_A, \approx_A \rangle$
and $B = \langle \Pi_B, \Theta_B, \approx_B \rangle$
is the AFS $C' = Ty(Eq(Cl(C)))$, where:
\begin{itemize}
\item
$C = \langle \Pi_C, \Theta_C, \approx_C \rangle$
\item
$\Pi_C = \Pi_A \cup \Pi_B$
\item
$\Theta_C(\pi) = \left\{ \begin{array}{ll}
		\Theta_A(\pi) \unif \Theta_B(\pi)	& 
		\mbox{if $\pi \in \Pi_A$ and $\pi \in \Pi_B$} \\
		\Theta_A(\pi)	& 
		\mbox{if $\pi \in \Pi_A$ only} \\
		\Theta_B(\pi)	& 
		\mbox{if $\pi \in \Pi_B$ only}
			\end{array}
		\right.$
\item
$\approx_C = \approx_A \cup \approx_B$
\end{itemize}
The unification {\bf fails} if there exists a path $\pi \in \Pi_{C'}$
such that $\Theta_{C'}(\pi) = \top$.
\end{definition} 

\begin{lemma}
\label{lemma-cl}
$Cl$ preserves prefixes:
If  $A$ is a prefix-closed pre-AFS and $A' = Cl(A)$
then $A'$ is prefix-closed.
\end{lemma}
\proof{}
Let $\pi$ be a path in $\Pi'$. If $\pi \in \Pi$ then every prefix of
$\pi$ is in $\Pi'$, since $\Pi$ is prefix-closed and $Cl$ only adds
paths. Suppose that $\pi \in \Pi' \setminus \Pi$. Then there exist
$\pi_1,\pi_2,\alpha_1,\alpha_2 \in \paths$ such that $\pi_1 \alpha_1
\in \Pi$ and $\pi_2 \alpha_2 \in \Pi$ and $\pi_1 \approx \pi_2$ and $\pi =
\pi_1 \alpha_2$ (otherwise, $\pi$ can be removed from $\Pi'$, in
contradiction to the minimality of $Cl$). 
If $\pi'$ is a prefix of $\pi$ than 
either $\pi'$ is a prefix of $\pi_1$, in which case $\pi' \in \Pi$
since $\Pi$ is prefix-closed, or $\pi' = \pi_1
\alpha'$ for some $\alpha'$ that is a prefix of $\alpha$. Since $\Pi$
is prefix-closed, $\pi_1 \alpha' \in \Pi$ and $\pi_2 \alpha' \in \Pi$.
Therefore, as $\pi_1 \approx \pi_2, \pi_1 \alpha'$ is added to $\Pi'$
by the closure operation.

\begin{lemma}
\label{lemma-eq}
$Eq$ preserves prefixes and fusions:
If $A$ is a prefix- and fusion-closed pre-AFS and $A' = Eq(A)$ then
$A'$ is prefix- and fusion-closed.
\end{lemma}
\proof{}
$Eq$ extends $\approx$ to an equivalence relation. 
Since only $\approx$ is modified, prefix-closure is trivially maintained.
Select a pair
$(\pi_1,\pi_2) \in \; \approx' \setminus \approx$. 
Then either (1) $\pi_2 = \pi_1$;
(2) $\pi_2 \approx \pi_1$; or (3) there exists a path $\pi_3$ such
that $\pi_1 \approx \pi_3$ and $\pi_3 \approx \pi_2$. Trivially, (1)
and (2) preserve the closure properties.
In the case of (3), to show
that fusion-closure is maintained we have to show that if $\pi_1
\alpha_1 \in \Pi'$ and $\pi_2 \alpha_2 \in \Pi'$ then $\pi_1 \alpha_2
\in \Pi'$ and $\pi_1 \alpha_2 \approx' \pi_2 \alpha_2$. 
Since $\Pi = \Pi', \pi_1 \alpha_1 \in \Pi$ and $\pi_2 \alpha_2 \in \Pi$.
Since $\Pi$ is fusion-closed and $\pi_2 \approx \pi_3, \pi_3
\alpha_2 \in \Pi$ and $\pi_3 \alpha_2 \approx \pi_2 \alpha_2$. Since
$\pi_1 \approx \pi_3, \pi_1 \alpha_2 \in \Pi$ and $\pi_1 \alpha_2
\approx \pi_3 \alpha_2$, too. $\approx'$ is an extension of $\approx$
to an equivalence relation, and thus $\pi_1 \alpha_2 \approx' \pi_2
\alpha_2$.

\begin{corollary}
If $A$ and $B$ are AFSs, then so is $A \unif B$.
\end{corollary}
\proof{}
If $A$ and $B$ are AFSs then the pre-AFS $C$, defined as
in~\ref{def:unification}, is prefix-closed (since $A$ and $B$ are).
$Cl(C)$ is prefix- and fusion-closed, as is $Eq(Cl(C))$ in which,
additionally, $\approx$ is an equivalence relation. $Ty(Eq(Cl(C)))$ is an
AFS, since $Ty$ only modifies $\Theta$ such that it respects the equivalences.

$C'$ is the smallest AFS that contains $\Pi_C$ and $\approx_C$. Since
$\Pi_A$ and $\Pi_B$ are prefix-closed, so is $\Pi_C$.  
However,
$\Pi_C$ and $\approx_C$ might not be fusion-closed. This is why $Cl$
is applied to them. As a result of its application, new paths and
equivalence classes might be added. By lemma~\ref{lemma-cl}, 
if a path is added
all its prefixes are added, too, so the prefix-closure is preserved.
Then, $Eq$ extends $\approx$ to an equivalence
relation, without harming the prefix- and fusion-closure properties
(by lemma~\ref{lemma-eq}).
Finally, $Ty$ sees to it that $\Theta$ respects the equivalences.

\begin{lemma}
Unification is commutative: $A \unif B = B \unif A$.
\end{lemma}
\proof{}
Observe that unification is defined using set union ($\cup$) and type
unification ($\unif$) which are commutative. Therefore, the
unification is commutative, too.

\begin{lemma}
\label{associative}
Unification is associative: $(A \unif B) \unif C = A \unif (B \unif
C)$.
\end{lemma}
{\bf Proof:} as above.

The result of a unification can differ from any of its arguments 
in three ways:
paths that were not present can be added;
the types of nodes can become more specific; and
reentrancies can be added, that is, the number of equivalence classes
of paths can decrease.
Consequently, the result of a unification is always more specific than
any of its arguments.
\begin{theorem}
If $C' = A \unif B$ then $A \preceq C'$.
\end{theorem}
\proof{}
$\Pi_C = \Pi_A \cup \Pi_B$ and hence $\Pi_A \subseteq \Pi_C$.
$\approx_C = \; \approx_A \cup \approx_B$ and hence $\approx_A
\subseteq \approx_C$. If $\pi \in \Pi_A$ then $\Theta_C(\pi) =
\Theta_A(\pi)$ or $\Theta_C(\pi) = \Theta_A(\pi) \unif \Theta_B(\pi)$,
and in any case $\Theta_A(\pi) \subsumes \Theta_C(\pi)$.
$Cl$ and $Eq$ cannot remove paths or equivalences and $Ty$ only makes
types more specific, and therefore $A \preceq C'$.

\begin{theorem}
$A \unif B = A$ iff $B \preceq A$.
\end{theorem}
\proof{}
Suppose $B \preceq A$. Then $\Pi_B \subseteq \Pi_A, \approx_B
\subseteq \approx_A$ and for every $\pi \in \Pi_B, \Theta_B(\pi)
\subsumes \Theta_A(\pi)$. $A \unif B = Ty(Eq(Cl(C)))$ where $C =
\langle \Pi_C, \Theta_C, \approx_C \rangle$ and
\begin{itemize}
\item
$\Pi_C = \Pi_A \cup \Pi_B = \Pi_A$
\item
$\Theta_C(\pi) = \left\{ \begin{array}{ll}
		\Theta_A(\pi) \unif \Theta_B(\pi)	& 
		\mbox{if $\pi \in \Pi_A$ and $\pi \in \Pi_B$} \\
		\Theta_A(\pi)	& 
		\mbox{if $\pi \in \Pi_A$ only} \\
		\Theta_B(\pi)	& 
		\mbox{if $\pi \in \Pi_B$ only}
			\end{array}
		\right.
		= \Theta_A(\pi)$
\item
$\approx_C = \approx_A \cup \approx_B = \approx_A$
\end{itemize}
Hence $A = C$ and therefore $A \unif B = A$.\\
Suppose $A \unif B = A$ and assume toward a contradiction that $B
\not\preceq A$. Then at least one of the following cases holds:

\begin{itemize}
\item
$\Pi_B \not\subseteq \Pi_A$. Then there exists $\pi \in \Pi_B \cup
\Pi_A$ that $\pi \not\in \Pi_A$ and hence $A \unif B \neq A$.
\item
There exists some $\pi$ such that $\Theta_B(\pi) \not\subsumes
\Theta_A(\pi)$. Then $\Theta_A(\pi) \unif \Theta_B(\pi) \neq
\Theta_A(\pi)$ and hence $A \unif B \neq A$.
\item
$\approx_B \not\subseteq \approx_A$. Then there exist $\pi_1,\pi_2$
such that $(\pi_1 \approx_B \pi_2)$ but not $(\pi_1 \approx_A \pi_2)$.
Hence $(\approx_A \cup \approx_B) \neq \approx_A$ and $A \unif B \neq A$.
\end{itemize}

TFSs (and therefore AFSs) can be seen as a generalization of
first-order terms (FOTs) (see~\cite{carp91}). Accordingly, AFS unification
resembles FOT unification; however, the notion of {\em substitution}
that is central to the definition of FOT unification is missing here,
and as far as we know, no analog to substitutions in the
domain of feature structures was ever presented.

\section{A Linear Representation of Feature Structures}
\label{linear-rep}
Representing feature structures as either graphs or attribute-value
matrices is cumbersome; we now define a linear representation for
feature structures, based upon A\"{\i}t-Kaci's $\psi$-terms (though
the order relation we use is reversed). 

\begin{definition}[Arity]
The {\bf arity} of a type $t$ is the number of features appropriate for
it, i.e.\ $|\{f \mid Approp(f,t) \isdef\}|$. 
\end{definition} 
Note that in
every totally well-typed feature structure of type $t$ the number of
edges leaving the root is exactly the arity of $t$. Consequently, we
use the term `arity' for (totally well-typed) feature structures: the
arity of a feature structure of type $t$ is defined to be the arity of $t$.

In order to define the set of well-formed linear terms over \feats\ and
\types, we assume that the feature names are ordered in a fixed order.

Let $\{\tag{i} \mid i \; \mbox{is a natural number}\}$ be the set of
{\bf tags}. 

\begin{definition}[Terms]
A {\bf term} $\tau$ of type $t$ is an expression of the form $\tag{i}
t(\tau_1,\ldots,\tau_n)$ where \tag{i} is a tag, $n \ge 0$ and every $\tau_i$
is a term of some type.
\end{definition} 

\begin{definition}[Totally well-typed terms]
A term $\tau = \tag{i} t(\tau_1,\ldots,\tau_n)$ of type $t$ 
is {\bf totally well-typed} iff:
\begin{itemize}
\item
$t$ is a type of arity $n$;
\item
the appropriate features for the type $t$ are $f_1, \ldots,f_n$, in
this order;
\item
for every $i$, $1 \le i \le n$, $Approp(f_i,t) = t_i$;
\item
for every $i$, $1 \le i \le n$, if $\tau_i$ is a term of type $t'_i$ then
either $t_i \subsumes t'_i$ or $t'_i = \bot$
\end{itemize}
\end{definition}

We distinguish tags that appear in terms according to the type they
are attached to: if a sub-term consists of a tag and the type $\bot$,
we say that the tag is {\bf independent}. Otherwise, the tag is {\bf
dependent}. 
We will henceforth consider only terms for which the
following proposition holds:

\begin{definition}[Normal terms]
A totally well-typed term $\psi = \tag{i} t(\tau_1,\ldots,\tau_n)$ is {\bf normal}
iff:
\begin{itemize}
\item
$t \neq \top$;
\item
if a tag \tag{j} appears in $\psi$ then its first (leftmost) occurrence
might be
dependent. If it appears more than once, its other occurrences are
independent.
\item
$\tau_1,\ldots,\tau_n$ are normal terms.
\end{itemize}
\end{definition} 

We use terms to represent feature structures. We define below an
algebra over which terms are to be interpreted. The denotation of
a normal term is a totally well-typed feature structure.

\begin{definition}[Feature structure algebra]
A {\bf feature structure algebra} is a structure $\cal{A} = \langle
D_{\cal{A}}, \{t_{\cal{A}} \mid t \in \types \}, \{f_{\cal{A}} \mid f
\in \feats \} \rangle$, such that:
\begin{itemize}
\item
$D_{\cal{A}}$ is a non-empty set, the {\bf domain} of $\cal{A}$;
\item
for each $t \in \types$, $t_{\cal{A}} \subseteq D_{\cal{A}}$ and, in
particular:
\begin{itemize}
\item
$\top_{\cal{A}} = \emptyset$;
\item
$\bot_{\cal{A}} = D_{\cal{A}}$;
\item
if $t_1 \unif t_2 = t$ then ${t_1}_{\cal{A}} \cap {t_2}_{\cal{A}} = t_{\cal{A}}$
\end{itemize}
\item
for each $f \in \feats$, $f_{\cal{A}}$ is a total function
$f_{\cal{A}} : D_{\cal{A}} \onto D_{\cal{A}}$
\end{itemize}
\end{definition} 

Let $D_G$ be the domain of all typed feature structures over \types\
and \feats. The
interpretation of $t_G$ over this domain is the set of feature
structures whose roots have the type $t$; the interpretation of $f_G :
D_G \onto D_G$ is the function that, given a feature structure $A$,
returns 
$val(A,f)$.

We associate a normal term $\psi$ with a totally well-typed feature
structure $A$ in the following way: 
\begin{itemize}
\item
if $\psi = \tag{i}t()$ then $A =
(\{\tag{i}\},\tag{i},\delta_{\isndef},\theta_t)$ where $\delta_{\isndef}$
is undefined for every input and $\theta_t(q) = t$ if $q = \tag{i}$ and
undefined otherwise;
\item
if $\psi = \tag{i}t(\tau_1,\ldots,\tau_n)$ then $A =
(Q,\tag{i},\delta,\theta)$ where $\theta(\tag{i}) = t$ and for every
$j$, if $f_j$ is the $j$-th appropriate feature of the type $t$, then
$\delta(\tag{i},f_j) = q_j$ and $q_j$ is the root of the feature
structure associated with $\tau_j$.
\end{itemize}

Conversely, associate a feature structure $A =
(Q,\qbar,\delta,\theta)$ with a normal term $\psi =
\tag{i}t(\tau_1,\ldots,\tau_n)$, where:
\begin{itemize}
\item
$\tag{i} = \qbar$;
\item
$\theta(\qbar) = t$;
\item
$n$ is the number of outgoing edges from $\qbar$;
\item
for every $j$, $1 \le j \le n$, $\tau_j$ is the term associated with
$\delta(\qbar,f_j)$ where $f_j$ is the $j$-th appropriate feature of
$t$;
\item
if the tag \tag{i} occurs elsewhere in $\tau_1,\ldots,\tau_n$, we
replace the term that \tag{i} depends on with the term $\bot()$,
making this occurrence of \tag{i} independent.
\end{itemize}

To summarize, there is a one-to-one correspondence between totally
well-typed feature structures and normal terms.

Note that the tags are only a means of encoding reentrancy in feature
structures. Therefore, when displaying a term in which 
a tag \tag{i} appears just once in a
term,
we will sometimes omit the tag for the sake of compactness.
Then, we sometimes omit the type of independent tags, which are
implicitly typed by $\bot$, and display them as tags only.


\section{Multi-rooted Structures}
\label{mrs}
To be able to represent complex linguistic information, such as phrase
structure, the notion of feature structures is usually extended.
There are two different approaches for representing phrase structure
in feature structures: by adding special, designated features to the
FSs themselves; or by defining an extended notion of FSs. The first
approach is employed by HPSG: special features, such as DTRS
(daughters), encode trees in TFSs as lists. This makes it impossible
to directly access a particular daughter. \namecite{shieber92} uses a
variant of this approach, where a denumerable set of special features,
namely $0,1,\ldots,$ are added to encode the order of daughters in a
tree. In a typed system such as ours, this method would necessitate
the addition of special types as well; in general, no bound can be
placed on the number of features and types necessary to state rules
(see~\cite[p.\ 194]{carp92}).

As a more coherent, mathematically elegant solution, we adopt below
the other approach: a new notion of {\em multi-rooted feature
structures}, suggested by~\cite{sikkel}, is being defined to naturally
extend TFSs. These structures provide a means to represent phrasal
signs and grammar rules. They are used implicitly in the computational
linguistics literature, but to the best of our knowledge no explicit,
formal theory of these structures and their properties was formulated
before.

\begin{definition}[Multi-rooted structures]
A {\bf multi-rooted feature structure} (MRS) is a pair $\langle
\Qbar,G\rangle$ where $G =
\langle Q,\delta\rangle$ is a finite, directed, 
labeled graph consisting of a set $Q \subseteq \nodes$ of nodes and a
partial function $\delta: Q \cross \feats \onto Q$ specifying the arcs, and
where $\Qbar$ is an ordered, (repetition-free) list of
distinguished nodes in $Q$ called {\bf roots}.
$G$ is not necessarily connected, but the union of all the
nodes reachable from all the roots in $\Qbar$ is required to yield
exactly $Q$.
The {\bf length} of a MRS is the number of its roots,
$|\Qbar|$.
$\emptymrs$ denotes the empty MRS, where $Q = \emptyset$.
\end{definition} 

Meta-variables $\sigma,\rho$ range over
MRSs, and $\delta, Q$ and $\Qbar$ over their constituents.
If $\langle \Qbar,G\rangle$ is a MRS and $\qbar_i$ is a root in $\Qbar$ then
$\qbar_i$ naturally induces a feature structure 
$Pr(\Qbar,i) = (Q_i,\qbar_i,\delta_i)$, where $Q_i$ is
the set of nodes reachable from $\qbar_i$ and $\delta_i =  \delta |_{Q_i}$.

\begin{figure}[hbt]
\center
\fbox{
\psfig{figure=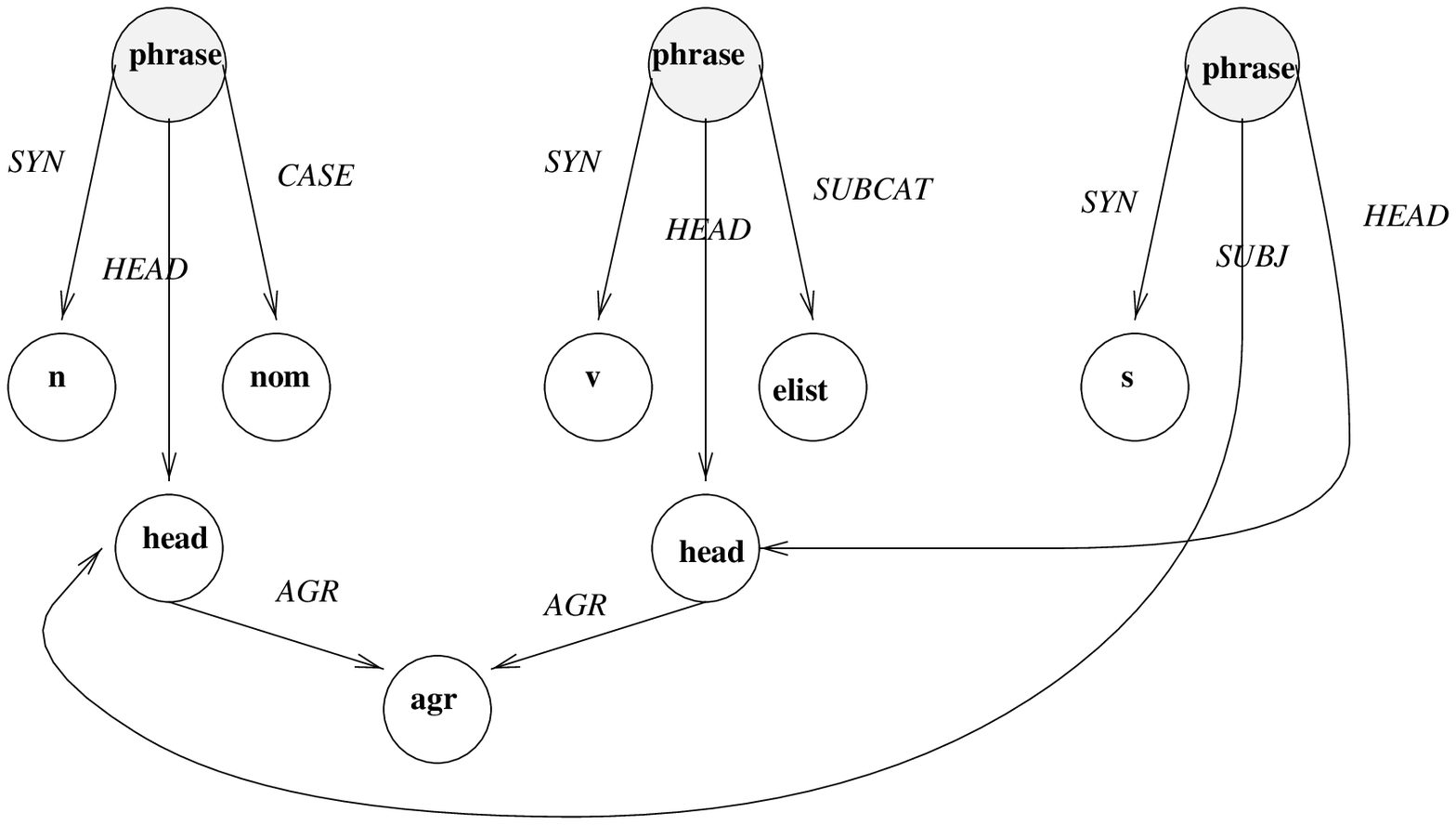,width=10cm}
}
\[
\begin{tfs}{phrase}
	SYN:	& \begin{tfs}{n} \end{tfs} \\
	HEAD:	& \tag{1}\begin{tfs}{head}
			AGR: 	& \tag{3}\begin{tfs}{agr} \end{tfs}
			\end{tfs} \\
	CASE:	& \begin{tfs}{nom} \end{tfs}
\end{tfs}
\begin{tfs}{phrase}
	SYN: 	& \begin{tfs}{v} \end{tfs} \\
	HEAD:	& \tag{2}\begin{tfs}{head}
			AGR:	& \tag{3}
			\end{tfs} \\
	SUBCAT:	& \begin{tfs}{elist} \end{tfs}
\end{tfs}
\begin{tfs}{phrase}
	SYN:	& \begin{tfs}{s} \end{tfs} \\
	SUBJ:	& \tag{1} \\
	HEAD:	& \tag{2}
\end{tfs}
\]
\mycaption{A graph- and AVM- representation of a MRS}{םישרוש-בר הנבמל םיגוציי ינש}
\label{fig:mrs}
\end{figure}

One can view a MRS $\langle \Qbar,G\rangle$ as an
ordered sequence $\langle A_1, \ldots, A_n\rangle$ of (not necessarily
disjoint) feature structures,
where $A_i = Pr(\Qbar,i)$ for $1 \le i \le n$.
Note that such an ordered list of feature structures is not a sequence
in the mathematical sense:
removing an element from the list may effect
the other elements (due to reentrancy among elements).
Nevertheless, we can think of a MRS as a sequence where a
subsequence is obtained by taking a subsequence of the roots and
considering only the feature structures they induce.
We use the two views interchangeably.
Figure~\ref{fig:mrs} depicts a MRS and its view as a sequence of feature
structures. The shaded nodes (ordered from left to right) constitute
$\Qbar$.

A MRS is well-typed if all its constituent feature structures are
well-typed, and is totally well-typed if all its constituents are.
Subsumption is extended to MRSs 
as follows:
\begin{definition}[Subsumption of multi-rooted structures]
A MRS $\sigma = \langle \Qbar,G\rangle$ {\bf subsumes} a MRS
$\sigma' = \langle \Qbar',G'\rangle$ (denoted by $\sigma \subsumes \sigma'$) 
if $|\Qbar| = |\Qbar'|$ and there
exists a total function $h:Q \onto Q'$ such that:
\begin{itemize}
\item
for every root $\qbar_i \in \Qbar, h(\qbar_i) = \qbar'_i$
\item
for every $q \in Q$, $\theta(q) \subsumes \theta'(h(q))$
\item
for every $q \in Q$ and $f \in \feats$, if $\delta(q,f) \isdef$ then
$h(\delta(q,f)) = \delta'(h(q),f)$
\end{itemize}
\end{definition} 

We define abstract multi-rooted structures in an analog way to abstract
feature structures.
\begin{definition}[Abstract multi-rooted structures]
A {\bf pre-abstract multi rooted structure} (pre-AMRS) is a quadruple $A
= \langle Ind,\Pi,\Theta,\approx \rangle$, where:
\begin{itemize}
\item
$Ind$, the {\bf indices} of $A$, 
is the set $\seq{n}$ for some $n$
\item
$\Pi \subseteq Ind \times \paths$ is a set of indexed paths, such
that for each $i \in Ind$ there exists some $\pi \in \paths$ that
$(i,\pi) \in \Pi$. 
\item
$\Theta: \Pi \onto \types$ is a total type-assignment function
\item
$\approx \subseteq \Pi \times \Pi$ is a relation
\end{itemize}
An {\bf abstract multi-rooted structure} (AMRS) is a pre-AMRS $A$ for
which the following requirements, naturally extending those of
AFSs, hold:
\begin{itemize}
\item
$\Pi$ is prefix-closed: if $(i,\pi\alpha) \in \Pi$ then $(i,\pi) \in \Pi$
\item
$A$ is fusion-closed: if $(i,\pi\alpha) \in \Pi$ and
$(i',\pi'\alpha') \in 
\Pi$ and $(i,\pi) \approx (i',\pi')$ then $(i,\pi\alpha') \in \Pi$ (as well as
$(i',\pi'\alpha) \in \Pi$), 
and $(i,\pi\alpha') \approx
(i',\pi'\alpha')$ (as well as $(i',\pi'\alpha) \approx (i,\pi\alpha)$)
\item
$\approx$ is an equivalence relation with a finite index
\item
$\Theta$ respects the equivalence: if $(i_1,\pi_1) \approx (i_2,\pi_2)$ then
$\Theta(i_1,\pi_1) = \Theta(i_2,\pi_2)$
\end{itemize}
\end{definition} 
An AMRS $\langle Ind,\Pi, \Theta, \approx \rangle$ is well-typed if for
every $(i,\pi) \in \Pi$,
$\Theta(i,\pi) \neq \top$ and if $(i,\pi f) \in \Pi$
then $Approp(f,\Theta(i,\pi)) \isdef$ and $Approp(f,\Theta(i,\pi))
\subsumes \Theta(i,\pi f)$.
It is totally well typed if, in addition, for every $(i,\pi)
\in \Pi$, if $Approp(f,\Theta(i,\pi))\isdef$ then $(i,\pi f) \in \Pi$.
The {\bf length} of an AMRS $A$ is $\len{A} = |Ind_A|$.
We use $\emptymrs$ to denote the empty AMRS, too, where
$Ind_{\emptymrs} = \emptyset$ and $\Pi_{\emptymrs} = \emptyset$ (so
that $\len{\emptymrs} = 0$).

The closure operations $Cl$ and $Eq$ are naturally extended to AMRSs:
If $A$ is a pre-AMRS then $Cl(A)$ is the least extension of $A$ that
is prefix- and fusion-closed, and $Eq(A)$ is the least extension of
$A$ to a pre-AMRS in which $\approx$ is an equivalence relation.
In addition, $Ty(\amrs{}) = \langle Ind,\Pi,\Theta',\approx \rangle$
where $\Theta'(i,\pi) = \bigsqcup_{(i',\pi') \approx (i,\pi)}
\Theta(i',\pi')$. 
The partial order $\preceq$ is extended to AMRSs: $\amrs{A} \preceq
\amrs{B}$ iff $Ind_A = Ind_B, \Pi_A \subseteq \Pi_B,
\approx_A \subseteq \approx_B$ and for every $(i,\pi) \in \Pi_A,
\Theta_A(i,\pi) \subsumes \Theta_B(i,\pi)$.
In the rest of this chapter we overload the symbol `$\subsumes$' so that
it denotes subsumption of AMRSs as well as MRSs.

AMRSs, too,  can be related to concrete ones in a natural way:
If $\sigma = \langle \Qbar,G\rangle$ is a MRS then $Abs(\sigma) =
\langle Ind_{\sigma},\Pi_\sigma, \Theta_\sigma, \approx_\sigma
\rangle$ is defined by:
\begin{itemize}
\item
$Ind_{\sigma} = \seq{|\Qbar|}$
\item
$\Pi_\sigma = \{(i,\pi) \mid \delta(\qbar_i,\pi)\isdef\}$
\item
$\Theta_\sigma(i,\pi) = \theta(\delta(\qbar_i,\pi))$
\item 
$(i,\pi_1)
\approx_\sigma (j,\pi_2)$ iff $\delta(\qbar_i,\pi_1) = \delta(\qbar_j,\pi_2)$
\end{itemize}
It is easy to see that $Abs(\sigma)$ is an AMRS. In particular, notice
that for every $i \in Ind_{\sigma}$ there exists a path $\pi$ such
that $(i,\pi) \in \Pi_{\sigma}$ since for every $i,
\delta(\qbar_i,\epsilon) \isdef$.
The reverse operation, $Conc$, can be defined in a similar manner.

AMRSs are used to represent ordered collections of AFSs. However, due to the
possibility of value sharing among the constituents of AMRSs, they are
not sequences in the mathematical sense, and the notion of
sub-structure has to be defined in order to relate them to AFSs.

\begin{definition}[Sub-structures]
Let $A=\amrs{A}$; let $Ind_B$ be a finite (contiguous) subset of
$Ind_A$; let $n+1$ be the index of the first element of $Ind_B$.
The {\bf
sub-structure} of $A$ induced by $Ind_B$ is an AMRS $B=\amrs{B}$
such that:
\begin{itemize}
\item
$(i-n,\pi) \in \Pi_B$ iff $i \in Ind_B$ and $(i,\pi) \in A$
\item
$\Theta_B(i-n,\pi) = \Theta_A(i,\pi)$ if $i \in Ind_B$
\item
$(i_1-n,\pi_1) \approx_B (i_2-n,\pi_2)$ iff  $i_1 \in Ind_B, i_2 \in
Ind_B$ and $(i_1,\pi_1) \approx_A (i_2,\pi_2)$
\end{itemize}
\end{definition} 
A sub-structure of $A$ is obtained by selecting a subsequence of the
indices of
$A$ and considering the structure they induce. 
Trivially, this structure is an AMRS.
We use $A^{j..k}$ to refer to the sub-structure of $A$ induced by
$\{ j, \ldots, k \}$.
If $Ind_B = \{i\}$, $A^{i..i}$ can be identified with an AFS, denoted $A^i$.

The notion of concatenation has to be defined for AMRSs, too. Notice
that by definition, concatenated AMRSs cannot share elements between them.
\begin{definition}[Concatenation]
The {\bf concatenation} of $A = \amrs{A}$ and $B=\amrs{B}$
of lengths $n_A, n_B$, respectively
(denoted by $A~\cdot~B$), is an AMRS $C=\amrs{C}$ such that
\begin{itemize}
\item
$Ind_C = \seq{n_A+n_B}$
\item
$\Pi_C = \Pi_A \cup \{(i+n_A,\pi) \mid (i,\pi) \in \Pi_B\}$
\item
$\Theta_C(i,\pi) = \left\{\begin{array}{lll}
			\Theta_A(i,\pi)	& \mbox{if}	& i \le n_A\\
			\Theta_B(i-n_A,\pi)	& \mbox{if}	& i > n_A
			  \end{array}\right.$
\item
$\approx_C \; = \; \approx_A \cup \{((i_1+n_A,\pi_1),(i_2+n_A,\pi_2)) \mid
(i_1,\pi_1) \approx_B (i_2,\pi_2)\}$
\end{itemize}
\end{definition} 
As usual, $A \cdot \emptymrs = \emptymrs \cdot A = A$.

We now extend the definition of unification to AMRSs: we want to allow
the unification of two $AMRSs$, according to a specified set of
indices.  Therefore, one operand is a pair consisting of an AMRS and a
set of indices, specifying some elements of it. The second operand is
either an AMRS or an AFS, considered as an AMRS of length~1. Recall
that due to reentrancies, other elements of the first AMRS can be
affected by this operation. Therefore, the result of the unification
is a new AMRS. We refer to AMRS unification as ``unification in
context'' in the sequel to emphasize the effect that the operation
might have on elements that are not directly involved in it.

\begin{definition}[Unification of AMRSs]
Let $A = \amrs{A}$ be an AMRS. Let $B = \amrs{B}$ be an AMRS (if $B$
is an AFS it is interpreted as an AMRS of length 1).
Let $J$ be a set of indices such that $J \subseteq Ind_A$.
Let $f(i) = i$ if $B$ is an AMRS, $f(i) = 1$ if $B$ is an AFS.
$(A,J) \unif B$ is defined if $B$ is an AMRS and $J \subseteq Ind_B$,
or if $B$ is an AFS and $|J| = 1$; in any case, it is the
AMRS $C'=Ty(Eq(Cl(\amrs{C})))$, where
\begin{itemize}
\item
$Ind_C = Ind_A$
\item
$\Pi_C = \Pi_A \cup \{(i,\pi) \mid i \in J$ and $(f(i),\pi) \in \Pi_B$
\item
$\Theta_C(i,\pi) = \left\{ \begin{array}{ll}
		\Theta_A(i,\pi)		& 
		\mbox{if $i \not\in J$} \\
		\Theta_A(i,\pi) \unif \Theta_B(f(i),\pi)	& 
		\mbox{if $i \in J$ and $(i,\pi) \in \Pi_A$ and
$(f(i),\pi) \in \Pi_B$} \\
		\Theta_A(i,\pi) 	& 
		\mbox{if $i \in J$ and $(i,\pi) \in \Pi_A$ and
$(f(i),\pi) \not\in \Pi_B$} \\
		\Theta_B(f(i),\pi)		& 
		\mbox{if $i \in J$ and $(i,\pi) \not\in \Pi_A$ and
$(f(i),\pi) \in \Pi_B$} \\
			\end{array}
		\right.$
\item
$\approx_C \, = \, \approx_A \cup \{((i_1,\pi_1),(i_2,\pi_2)) \mid
i_1,i_2 \in J$ and $(f(i_1),\pi_1) \approx_B (f(i_2),\pi_2)\}$
\end{itemize}
The unification fails if there exists some pair
$(i,\pi) \in \Pi_{C'}$ such that $\Theta_{C'}(i,\pi) = \top$.
\end{definition} 


Many of the properties of AFSs, proven in the previous section, hold
for AMRSs, too. In particular, if $A,B$ are AMRSs then so is $(A,J)
\unif B$ if it is defined, $\len{(A,J) \unif B} = \len{A}$ and $(A,J)
\unif B \subsumed A$.  Also, for every two AMRSs $A,B$, $(A,\{1 \ldots
\len{A}\}) \unif B = A$ iff $B^{1 \ldots \len{A}} \subsumes A$.

The linear representation of TFSs, suggested in section~\ref{linear-rep}, is
naturally extended to MRSs: a multi-term is a sequence of terms, where
the scope of tags is extended to the entire sequence. 

\section{Rules and Grammars}
\label{rules-and-grammars}
We define rules and grammars over a fixed set \words\ of words (in
addition to the fixed sets \feats\ and \types). We use $w$ to refer to
elements of \words, $w_i$ to refer to strings over \words. We assume
that the lexicon associates with every word $w_i$ a set of feature
structures $Cat(w_i)$, its {\bf category},\footnote{$Cat(w_i)$ is a
singleton if $w_i$ is unambiguous.}  so we can ignore the terminal words
and consider only their categories. The input for the parser,
therefore, is a sequence\footnote{We assume that there is no
reentrancy among lexical items.} of sets of TFSs rather than a string
of words.

\begin{definition}[Pre-terminals]
Let $w=w_1\ldots w_n \in \words^{*}$.
$PT_w(j,k)$ is 
defined iff $1 \le j, k \le n$, in which case it is the set of AMRSs
$Abs(\langle A_j, A_{j+1}, \ldots, A_k \rangle)$ where $A_i \in
Cat(w_i)$ for $j \le i \le k$.
If $j>k$ then $PT_w(j,k) = \{\emptymrs\}$.
We omit the subscript $w$ when it is clear from the context.
\end{definition} 

\begin{lemma}
\label{lemma:pt-concat}
If $w = w_1 \cdots w_n$, $1 \le i \le j \le k \le n$, $A \in
PT_w(i,j)$ and $B \in PT_w(j+1,k)$ then $A \cdot B \in PT_w(i,k)$.
\end{lemma}
\proof{ An immediate corollary of the definition.}

\begin{definition}[Rules]  
A {\bf rule} is a MRS of length $n>0$ with a distinguished last
element. If $\langle A_1,\ldots,A_{n-1},A_n\rangle$ is a rule then
$A_n$ is its {\bf head}\footnote{This use of {\em head} must not be
confused with the linguistic one, the core features of a phrase.}  and
$\langle A_1,\ldots,A_{n-1}\rangle$ is its {\bf body}.\footnote{Notice
that the traditional direction is reversed and that the head and the
body need not be disjoint.}  We write such a rule as $\langle
A_1,\ldots,A_{n-1} \Rightarrow A_n\rangle$.  In addition, every
category of a lexical item is a rule (with an empty body). We assume
that such categories don't head any other rule.
\end{definition} 
Notice that the definition supports $\epsilon$-rules, i.e., rules with
null bodies.

\begin{definition}[Grammars] 
A {\bf grammar} $G=(\rules,A_s)$
is a finite set of rules $\rules$ and a {\bf start symbol} $A_s$ that is a TFS.
\end{definition} 

Figure~\ref{fig:grammar} depicts an example grammar (we use AVM
notation for this rule; tags such as $\tag{1}$ denote
reentrancy). While this example grammar has no linguistic pretensions,
it might be viewed as generating simple sentences in which the
predicates are headed by transitive and intransitive verbs. The type
hierarchy on which the grammar is based is omitted here.  A discussion
of the methodological status of the start symbol appears later on in
this section, prior to the definition of {\em languages}.
\begin{figure}[hbt]
Initial symbol:
{\scriptsize
\[
\begin{tfs}{phrase}
	CAT: & \begin{tfs}{s} \end{tfs}
\end{tfs}
\]
}
Lexicon:
{\tiny
\[
\begin{array}{ccc}
\mbox{John} & \mbox{her} & \mbox{loves} \\
\begin{tfs}{word}
	CAT:	& \begin{tfs}{n} \end{tfs} \\
	AGR:	& \begin{tfs}{agr}
			PER:	& \begin{tfs}{3rd} \end{tfs} \\
			NUM:	& \begin{tfs}{sg} \end{tfs} 
			\end{tfs} \\
	SEM:	& \begin{tfs}{sem} PRED: & \begin{tfs}{john}\end{tfs} \end{tfs}
\end{tfs}
&
\begin{tfs}{word}
	CAT:	& \begin{tfs}{n} CASE: & \begin{tfs}{acc} \end{tfs}\end{tfs} \\
	AGR:	& \begin{tfs}{agr}
			PER:	& \begin{tfs}{3rd} \end{tfs} \\
			NUM:	& \begin{tfs}{sg} \end{tfs} 
			\end{tfs} \\
	SEM:	& \begin{tfs}{sem} PRED: & \begin{tfs}{she} \end{tfs} \end{tfs}
\end{tfs}
&
\begin{tfs}{word}
	CAT:	& \begin{tfs}{v} \end{tfs} \\
	AGR:	& \begin{tfs}{agr}
			PER:	& \begin{tfs}{3rd} \end{tfs} \\
			NUM:	& \begin{tfs}{sg} \end{tfs} 
			\end{tfs} \\
	SEM:	& \begin{tfs}{sem} PRED: & \begin{tfs}{love} \end{tfs} \end{tfs}
\end{tfs}
\end{array}
\]
}
Rules:
{\tiny
\begin{eqnarray}
\label{rule:snpvp}
\begin{tfs}{sign}
	CAT:	& \begin{tfs}{n} CASE: & \begin{tfs}{nom}\end{tfs} \end{tfs} \\
	AGR:	& \tag{1} \\
	SEM:	& \begin{tfs}{sem} PRED: & \tag{3} \end{tfs}
\end{tfs}
\begin{tfs}{sign}
	CAT: 	& \begin{tfs}{v} \end{tfs} \\
	AGR:	& \tag{1}\\
	SEM:	& \tag{2}
\end{tfs}
&\longrightarrow &
\begin{tfs}{phrase}
	CAT:	& \begin{tfs}{s} \end{tfs} \\
	AGR:	& \tag{1}\\
	SEM:	& \tag{2}\begin{tfs}{sem}
			ARG1: & \tag{3}
		\end{tfs}
\end{tfs}
\\
\label{rule:vpvnp}
\begin{tfs}{sign}
	CAT:	& \begin{tfs}{v} \end{tfs} \\
	AGR:	& \tag{1} \\
	SEM:	& \tag{2}
\end{tfs}
\begin{tfs}{sign}
	CAT:	& \begin{tfs}{n} CASE: & \begin{tfs}{acc}\end{tfs} \end{tfs} \\
	SEM:	& \begin{tfs}{sem} PRED: \tag{3} \end{tfs}
\end{tfs}
&\longrightarrow &
\begin{tfs}{phrase}
	CAT:	& \begin{tfs}{v} \end{tfs} \\
	AGR:	& \tag{1} \\
	SEM:	& \tag{2}\begin{tfs}{sem} ARG2: & \tag{3} \end{tfs}
\end{tfs}
\end{eqnarray}
}
\mycaption{An example grammar}{המגודל קודקד}
\label{fig:grammar}
\end{figure}

For the following discussion we fix a particular grammar
$G=(\rules,A_s)$.  We define a {\em derivation} relation over AMRSs as
the basis for defining the {\em language} of TFS-based
grammars. Checking whether two given AMRSs $A$ and $B$ stand in the
derivation relation is accomplished by the following steps: first, an
element of $A$ has to be selected; this element has to unify with the
head of some rule $\rho$; then, a sub-structure of $B$ is selected;
this substructure has to unify with the body of $\rho$. All
unifications are done in context, so that other components of the
AMRSs involved may be affected, too. Moreover, there must be some way
to record the effects of successive unifications; to this end,
derivation is defined only for pairs of AMRSs that are already ``as
specific as needed''; that is to say, if the rule adds any information
to the AMRSs, this information already has to be recorded in them in
order for them to be related by derivation. This is why, in the
definition below, we use an AMRS $R$ that is {\em at least as specific
as} some rule $\rho$, and not $\rho$ itself, to guide the
derivation. This is also why the definition requires that all the
unifications do not add information. {\em strong derivation} is the
relation that holds between such AMRSs; another relation, {\em
derivation}, relaxes that requirement by allowing two AMRSs to be
related even if they contain only part of the information that is
required for strong derivation to hold.

Since elements of AMRSs involve indices that denote their linear
position in the sequence of roots, the operation of {\em replacing}
some element in one AMRS with a sub-structure, whose length might be
greater than one, becomes notationally complicated. Conceptually,
though, it resembles very much the replacement of some symbol with a
sequence of symbols in context-free derivation, or the replacement of
the selected goal (that unifies with the head of some rule) with the
body of the rule, in Prolog SLD-resolution. One main difference in our
definition is that we do not carry substitutions through sequences of
derivations; rather, we treat all the pairs in a derivation sequence as if
the appropriate substitutions have already been applied to them
(recall that members of these pairs are ``as specific as needed'').

\begin{figure}[hbt]
\center
\fbox{
\psfig{figure=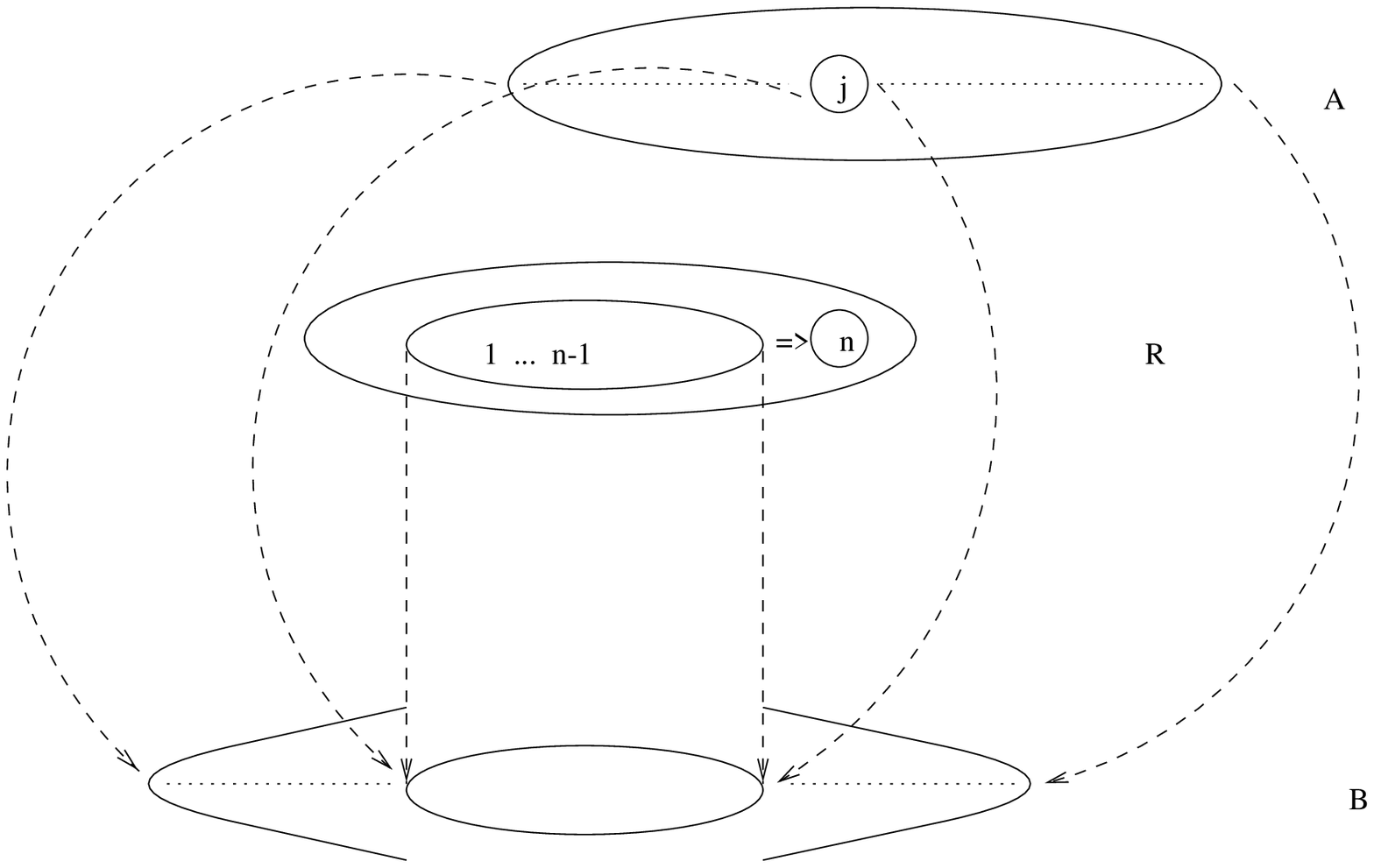,width=12cm}
}
\mycaption{Strong derivation}{הקזח הריזג}
\label{fig:derivation-def}
\end{figure}

\begin{definition}[Strong Derivation]
An AMRS $A=\amrs{A}$ whose length is $k$ {\bf strongly derives} an AMRS $B$
(denoted $A \derives B$) iff
\begin{itemize}
\item
there exist a rule $\rho \in \rules$ and an AMRS $R \subsumed
Abs(\rho)$ (with $\len{R} = n$), such that:
\item
some element of $A$ unifies with the head of $R$, and some
sub-structure of $B$ unifies with the body of $R$; namely, there exist
$j \in Ind_A$ and $i \in Ind_B$ such that:

\begin{tabular}{ll}
$A = (A,\{j\}) \unific R^n$,	&
$B^{i \ldots i+n-2} = (B,\{i \ldots i+n-2\}) \unif R^{1 \ldots n-1}$,	\\
$R = (R,\{n\}) \unific A^j$,	&
$R = (R,\{1 \ldots n-1\}) \unif B^{i \ldots i+n-2}$
\end{tabular}
%
\item
$B$ is the replacement of the $j$-th element of $A$ with the body of
$R$; namely, let
\[f(i) = \left\{ \begin{array}{ll}
			i	& \mbox{if $1 \le i < j$}\\
			i+n-2	& \mbox{if $j < i \le k$}
		    \end{array}\right., \hspace{0.7cm}
  g(i) = i+j-1\; \mbox{if}\; 1 \le i < n
\] 
then $B = Ty(Eq(Cl(\amrs{B'})))$, where
\begin{itemize}
	\item
	$Ind_{B'} = \langle 1,\ldots,k+n-2 \rangle$
	\item
	$(i,\pi) \in \Pi_{B'}$ iff $
 		\left\{ \begin{array}{llll}
			i = f(i') 		& \mbox{and}	&
			(i',\pi) \in \Pi_{A}	& \mbox{or}\\
			i = g(i') 	& \mbox{and} 	& 
			(i',\pi) \in \Pi_{R}
 			\end{array}
 		\right.$
	\item
	$\Theta_{B'}(i,\pi) = 
 		\left\{ \begin{array}{lll}
			\Theta_{A}(i',\pi) & \mbox{if} & i = f(i')\\
			\Theta_{R}(i',\pi) & \mbox{if} & i = g(i')
 			\end{array}
 		\right.$
	\item
	$(i_1,\pi_1) \approx_{B'} (i_2,\pi_2)$ if
	\begin{itemize}
	\item
	$i_1 = f(i'_1)$ and $i_2 = f(i'_2)$ and 
	$(i'_1,\pi_1) \approx_{A} (i'_2,\pi_2)$, or
	\item
	$i_1 = g(i'_1)$ and $i_2 = g(i'_2)$ and 
	$(i'_1,\pi_1) \approx_{R} (i'_2,\pi_2)$, or
	\item
	$i_1 = f(i'_1)$ and $i_2 = g(i'_2)$ and
	there exist $\pi_1,\pi_2,\pi_3$ such that
	$(i'_1,\pi_1) \approx_{A} (j,\pi_3)$ and $(n,\pi_3) \approx_{R}
	(i'_2,\pi_2)$, or
	\item
	$i_1 = g(i'_1)$ and $i_2 = f(i'_2)$ and
	there exist $\pi_1,\pi_2,\pi_3$ such that
	$(i'_1,\pi_1) \approx_{R} (j,\pi_3)$ and $(n,\pi_3) \approx_{A}
	(i'_2,\pi_2)$
	\end{itemize} 
\end{itemize}
\end{itemize}
The reflexive transitive closure of `$\derives$', denoted
`$\derivess$', is defined as follows: $A \derivess A''$ if
$A = A''$ or if there exists $A'$ such that
$A \derives A'$ and $A' \derivess A''$.
\end{definition} 
Intuitively, $A$ strongly derives $B$ through some AFS $A^j$ in $A$,
if some rule $\rho \in \rules$ licenses the derivation.  $A^j$ is
unified with the head of the rule, and if the unification succeeds,
the (possibly modified) body of the rule replaces $A^j$ in $B$.  The
definition is graphically demonstrated in figure~\ref{fig:derivation-def}.

%
\begin{lemma}
\label{lemma:specify}
If $A \derives B$ and $A \subsumes A'$ then there exists $B'$ such that
$B \subsumes B'$ and $A' \derives B'$.
\end{lemma}
\proof{}
$A \derives B$, therefore there exists a rule $\rho \in \rules$, an
AMRS $R \subsumed Abs(\rho)$ and an index $j$ such that $A$ unifies
with the head of $R$, and $B$ is obtained by replacing $A^j$ with the
body of $R$. $A$ and $B$ are already ``as specific as needed''; thus,
since $A \subsumes A'$ and $A = (A,\{j\}) \unif R^n$, $A' = (A',\{j\})
\unif R^n$. Hence there exists $R' \subsumed R$ such that $R' = (R',
\{n\}) \unif A'^j$, $A'$ unifies with its head and $B'$ is obtained by
replacing the $j$-th element of $A'$ with its body.

\begin{lemma}
\label{lemma:sspecify}
If $A \derivess B$ and $A \subsumes A'$ then there exists $B'$ such that
$B \subsumes B'$ and $A' \derivess B'$.
\end{lemma}
\proof{ By induction on the derivation sequence and lemma~\ref{lemma:specify}.}

\begin{lemma}
\label{lemma:concat}
If $A^{1\ldots k} \derivess B$ and $A^{k+1} \derivess C$ then
$A^{1\ldots k+1} \derivess B \cdot C$.
\end{lemma}
\proof{}
The derivation is obtained by applying first the derivation steps that
derive $B$ from $A^{1 \ldots k}$ and then those that derive $C$ from
$A^{k+1}$. Since $A^{1\ldots k} \derivess B$, $A$ is ``as specific as
needed'' and the application of the derivation steps from $A$ to $B$
does not affect the applicability of the derivations step to $C$.

\begin{lemma}
\label{lemma:complete}
If $A \subsumed Abs(\rho)$ for some $\rho \in \rules$ of length $n$
then $A^n \derives A^{1 \ldots n-1}$.
\end{lemma}
\proof{ Immediate from the definition of derivation.}

There are various definitions in the literature for the {\em language}
that is defined by a grammar $G$ expressed in a unification-based
grammar formalism. For example,~\cite{restriction,shieber92} do not
include a start symbol in the grammar at all, and define $L(G)$ as the
set of strings derivable from {\em some} feature structure.
In~\cite{deductive-parsing} a start symbol is defined (notated {\em
goal axiom}), and $L(G)$ is defined as the set of strings that are
derivable from some generalization of the start symbol, i.e., from
some feature structure that subsumes it. \cite{sikkel}, on the other
hand, assumes that a specific feature {\em cat} is present in every
feature structure (the value of which simulates non-terminals in a
context-free ``underlying'' grammar), and uses this feature to single
out the start symbol: $L(G)$ is the set of strings that are derivable
from some feature structure in which the {\em cat} feature is $S$ (the
start symbol of the underlying context-free grammar). A similar
definition is given by~\cite{haas}: $L(G)$ is the set of strings
derivable from the start symbol, where the start symbol is a {\em
constant} (that is, an atomic feature structure).

There is a good motivation to employ a start symbol: the grammar
writer might want to specify a certain criterion for the permissible
strings in the language, for example, that they are all {\em
sentences}. Moreover, it makes sense to include in the language such
strings that are not derived directly by the start symbol, but rather
by a TFS that is related to the start symbol. For example, the grammar
writer might state that only TFSs with a {\em cat} feature valued {\em
S} are permissible, meaning that every TFS that {\em is subsumed} by
the start symbol (that is, contains all the information it encodes) is
a sentence. However, such a definition prevents the incorporation of
{\em subsumption test} (see section~\ref{subsumption-test} below) into
the parsing, since the correctness of the computation can not be
maintained.

Due to these consideration we chose a relaxed condition on the start
symbol in our definition of languages. We define a derivation relation
between AMRSs in a way that allows the initial symbol of the grammar
to derive a sequence of lexical entries even if the actual (strong)
derivation is between a TFS that unifies with the start symbol and a
more specific instance of the pre-terminals.

\begin{definition}[Derivation]
An AMRS $A$ {\bf derives} an AMRS $B$ ($A \aderives B$) iff there
exist AMRSs $A',B'$ such that $(A,\seq{\len{A}}) \unif A' \neq \top$,
$B \subsumes B'$ and $A' \derivess B'$.
\end{definition} 


\begin{definition}[Language]
The {\bf language} of a grammar $G$
is  $L(G) = \{w=w_1 \cdots w_n \in \words^*\mid Abs(A_s) \aderives B$ for some
$B \in PT_w(1,n)\}$.
\end{definition} 
Figure~\ref{fig:derivation} depicts a derivation of the
string ``John loves her'' with respect to the example grammar. The
scope of reentrancy tags is limited to one MRS, of course, but we use
the same tags across different MRSs to emphasize the flow of
information during derivation.
This example shows that the
sentence ``John loves her'' is in the language of the example grammar,
since the derivation starts with a TFS that is more specific than the
initial symbol and ends in a specification of the lexical entries of
the sentences' words.
\begin{figure}[hbt]
{\tiny
\[
\begin{array}{ccc}
&
\begin{tfs}{phrase}
	CAT:	& \begin{tfs}{s} \end{tfs} \\
	AGR:	& \tag{1}\begin{tfs}{agr}
			PER:	& \begin{tfs}{3rd} \end{tfs} \\
			NUM:	& \begin{tfs}{sg} \end{tfs} 
			\end{tfs}\\
	SEM:	& \tag{2}\begin{tfs}{sem}
			PRED: & \begin{tfs}{love}\end{tfs}\\
			ARG1: & \tag{3}\begin{tfs}{john}\end{tfs}\\
			ARG2: & \tag{4}\begin{tfs}{she}\end{tfs}
		\end{tfs}
\end{tfs}
& \stackrel{\mbox{(\ref{rule:snpvp})}}{\longrightarrow}
\\
\\
\begin{tfs}{sign}
	CAT:	& \begin{tfs}{n} CASE: & \begin{tfs}{nom}\end{tfs} \end{tfs} \\
	AGR:	& \tag{1} \begin{tfs}{agr}
			PER:	& \begin{tfs}{3rd} \end{tfs} \\
			NUM:	& \begin{tfs}{sg} \end{tfs} 
			\end{tfs}\\
	SEM:	& \begin{tfs}{sem} 
			PRED: & \tag{3} \begin{tfs}{john}\end{tfs}
		\end{tfs}
\end{tfs} &
\begin{tfs}{sign}
	CAT: 	& \begin{tfs}{v} \end{tfs} \\
	AGR:	& \tag{1}\\
	SEM:	& \tag{2}\begin{tfs}{sem}
			PRED: & \begin{tfs}{love}\end{tfs}\\
			ARG1: & \tag{3}\\
			ARG2: & \tag{4}\begin{tfs}{she}\end{tfs}
		\end{tfs}
\end{tfs}
& \stackrel{\mbox{(\ref{rule:vpvnp})}}{\longrightarrow}
\\
\\
\begin{tfs}{word}
	CAT:	& \begin{tfs}{n} CASE: & \begin{tfs}{nom} \end{tfs}\end{tfs} \\
	AGR:	& \tag{1}\begin{tfs}{agr}
			PER:	& \begin{tfs}{3rd} \end{tfs} \\
			NUM:	& \begin{tfs}{sg} \end{tfs} 
			\end{tfs} \\
	SEM:	& \begin{tfs}{sem} 
			PRED: & \tag{3}\begin{tfs}{john}\end{tfs} 
			\end{tfs}
\end{tfs}
&
\begin{tfs}{word}
	CAT:	& \begin{tfs}{v} \end{tfs} \\
	AGR:	& \tag{1}\begin{tfs}{agr}
			PER:	& \begin{tfs}{3rd} \end{tfs} \\
			NUM:	& \begin{tfs}{sg} \end{tfs} 
			\end{tfs} \\
	SEM:	& \tag{2}\begin{tfs}{sem} 
			PRED: & \begin{tfs}{love} \end{tfs} 
			\end{tfs}
\end{tfs}
&
\begin{tfs}{word}
	CAT:	& \begin{tfs}{n} CASE: & \begin{tfs}{acc} \end{tfs}\end{tfs} \\
	AGR:	& \begin{tfs}{agr}
			PER:	& \begin{tfs}{3rd} \end{tfs} \\
			NUM:	& \begin{tfs}{sg} \end{tfs} 
			\end{tfs} \\
	SEM:	& \begin{tfs}{sem} 
			PRED: & \tag{4}\begin{tfs}{she} \end{tfs} 
			\end{tfs}
\end{tfs} 
\\
\\
\mbox{\normalsize John}	& \mbox{\normalsize loves} & \mbox{\normalsize her}
\end{array}
\]
}
\mycaption{A leftmost derivation}{תילאמש הריזג}
\label{fig:derivation}
\end{figure}

\section{Parsing as Operational Semantics}
\label{parsing}
Parsing is the process of determining whether a given string belongs to
the language defined by a given grammar, and assigning a structure to
the permissible strings. Various parsing algorithms exist for various
classes of grammars. In this section we formalize and explicate some
of the notions of~\cite[chapter 13]{carp92}. We give direct
definitions for rules, grammars and languages, based on our new notion
of AMRSs. This
presentation is more adequate to current TFS-based systems
than~\cite{haas,deductive-parsing}, that use first-order 
terms. Moreover, it does not necessitate special, ad-hoc features and
types for encoding trees in TFSs as~\cite{shieber92} does.
We don't assume any explicit context-free backbone for the grammars, as
do~\cite{lfg}~or~\cite{sikkel}.

The parsing algorithm we describe is a pure bottom-up one that makes
use of a chart to record edges.  The formalism we presented is aimed
at being a platform for specifying grammars in HPSG, which is
characterized by employing a few very general rules (or rule
schemata); selecting the rules that are applicable in every step of
the process requires unification anyhow.  Therefore we choose a
particular parsing algorithm that does not make use of top down
predictions but rather assumes that every rule might be applied in
every step. This assumption is realized by initializing the chart with
predictive edges for every rule, in every position.

As is well known (see, e.g.,~\cite{lloyd}), the meaning of a logic
program $P$ can be specified algebraically as the least fix-point
(lfp) of the {\em immediate consequence} operator $T_P$ of the
program.  A similar approach can be applied to a context-free grammar
$G$, such that $L(G)$ equals (a projection of) the least fix-point of
an analogous {\em immediate derivation} operator, $T_G$.
Let $G=(V,T,P,S)$ be a context-free grammar.\footnote{We assume a
normal form, where for $A \rightarrow \alpha \in P$, either $\alpha
\in T$ or $\alpha \in V^{*}$.} Let $I \subseteq V \times T^{*}$.
Define $T_G(I) = \{\langle A,w \rangle \mid A \rightarrow w \in P, w
\in T\} \cup \{\langle A,w_1 \cdots w_k \rangle \mid A \rightarrow A_1
\cdots A_k \in P, \langle A_i,w_i \rangle \in I, 1 \le i \le k\}$.
Then the least fix-point of $T_G$ is the union over $A \in V$ of
$\{\langle A,w \rangle \mid w \in L_A(G)\}$.

In a sense, computing the lfp of $T_G$ corresponds to computing the
language generated by $G$. Parsing, then, amounts to checking if the
input $w$ is in the language.  This process induces an inherently
inefficient computation: since $w$ is given, it can be used to
optimize the computation. This is achieved by defining $T_{G,w}$, a
{\em parsing step} operator, which is dependent on the input sentence
$w$. The set of items $I$ has to be extended, too: an item is a triple
$[i,A,j]$ where $0 \le i,j, \le n$ ($n$ being the length of $w$) and
$A \in V$.  Informally, an item $[i,A,j]$ represents the existence of
a derivation for the symbol $A$ to a substring of $w$, namely $w_i
\ldots w_j$.  $w \in L_S(G)$ if and only if $[1,S,n] \in
lfp(T_{G,w})$, so that parsing now amounts to computing the least
fixed point of $T_{G,w}$, which is more efficient, and then checking
whether the appropriate item is in the lfp.

We now return to TFS-based formalisms and define $T_{G,w}$ for a
TFS-based grammar $G$, thus providing means for defining the meaning
of $G$.  A computation is triggered by some input string of words
$w=w_1 \cdots w_n$ of length $n \ge 0$. For the following discussion
we fix a particular grammar $G=(\rules,A_s)$ and a particular input
string $w$ of length $n$.  A {\em state} of the computation is a set
of {\em items}, and states are related by a transition relation. The
presentation below corresponds to a pure bottom-up parsing algorithm,
as it is both simple and efficient.

\begin{definition}[Items]
An {\bf item} is a four-tuple $[i,A,j,c]$, where $i,j \in \ns$,
$i \le j$, $A$ is an AMRS and $c$ is either \act, in which
case the item is {\bf active}, or \comp, in which case it is {\bf complete}.
Let \items\ be the collection of all items.
\end{definition} 
If $[i,A,j,c]$ is an item, we say that $A$ {\em spans} the input from
position $i+1$ to position $j$ (inclusive). $A$ can be seen as a
representation of a {\em dotted rule}, or {\em edge}: during parsing
all generated items are such that $A$ is (possibly more specific
than) a prefix of some grammar rule. The notion of items usually
employs edges that contain entire rules, whereas we only use prefixes
of rules. This difference is not essential, and in an actual
implementation of a parser that is induced by $T_{G,w}$, edges indeed
include a reference to the rule on which they rely.

In what follows we define $T_{G,w}$, a parsing operator that
corresponds to (bottom-up) chart parsing. However, it is possible to
characterize algebraic operators that correspond to other parsing
schemas as well.
\setcounter{equation}{0}
\begin{definition}
\label{def:tgw}
Let $T_{G,w}: 2^{\items} \onto 2^{\items}$ be a transformation on sets of
items, where for $I \in \items$,$x \in T_{G,w}(I)$ iff either
\begin{equation}\begin{array}{l}
\label{dm}
\exists \rho \in \rules, Abs(\rho) =R= A_1,\ldots,A_{m-1} \Rightarrow A_m, m > 1 \\
\exists k < m-1	\\
\exists \alpha \in I, \alpha =
[i_{\alpha},A_{\alpha},j_{\alpha},\act], \len{A_{\alpha}} = k	\\
\exists \beta \in I, \beta = [i_{\beta},A_{\beta},j_{\beta},\comp],
\len{A_{\beta}} = 1	\\
j_{\alpha} = i_{\beta}	\\
B = (R,\{ 1 \ldots k \}) \unif A_{\alpha}	\\
C = (B,\{k+1\}) \unific A_{\beta}	\\
x = [i_{\alpha},C^{1 \ldots k+1}, j_{\beta},\act]
\end{array}\end{equation}
or
\begin{equation}\begin{array}{l}
\label{complete}
\exists \rho \in \rules, Abs(\rho) =R= A_1,\ldots,A_{m-1} \Rightarrow A_m, m > 1 \\
\exists \alpha \in I, \alpha =
[i_{\alpha},A_{\alpha},j_{\alpha},\act], \len{A_{\alpha}} = m-1	\\
C = (R,\{ 1 \ldots m-1 \}) \unif A_{\alpha}	\\
x = [i_{\alpha},C^m, j_{\alpha},\comp]
\end{array}\end{equation}
or
\begin{equation}\begin{array}{l}
\label{predict}
\exists i, 0 \le i \le n	\\
x = [i,\emptymrs,i,\act]
\end{array}\end{equation}
or
\begin{equation}\begin{array}{l}
\label{fact}
\exists \rho \in \rules, \len{\rho} = 1	\\
\exists i, 0 \le i \le n	\\
x = [i,Abs(\rho),i,\comp]
\end{array}\end{equation}
or
\begin{equation}\begin{array}{l}
\label{scan}
w = w_1,\ldots,w_n, n \ge 1	\\
\exists i, 0 < i \le n	\\
x = [i-1,Abs(A_i),i,\comp], A_i \in Cat(w_i)
\end{array}\end{equation}
\end{definition}

Cases~\ref{dm}~and~\ref{complete} perform the operation known as {\em
completion}:~\ref{dm} moves the dot one position along the body of a
rule, and~\ref{complete} creates a complete item once the dot reaches
the end of the body. Case~\ref{predict} corresponds to the {\em prediction}
operation, whereas case~\ref{scan} corresponds to {\em scanning}.
Case~\ref{fact} handles $\epsilon$-rules, i.e., rules with null
bodies, and creates complete items that span a null substring of the
input sentence. Notice that cases~\ref{predict}~and~\ref{fact} are
independent of the argument $I$ and therefore add the same items in
every application of $T_{G,w}$. Case~\ref{scan} is also independent of
the argument, but is dependent on the input sentence $w$.

The operator $T_{G,w}$, on which the algebraic semantics of TFS-based
grammars is based, naturally induces an operational semantics for such
formalisms: once the operator is shown to be continuous, a
computational process that corresponds to the iterative application of
$T_{G,w}$ computes the set of items in the least fix-point of the
operator. This process can be thought of as an analog of bottom-up,
chart-based parsing: the chart is initialized with predictions for
every rule in every position (by operation~\ref{predict}) and with
complete edges for every input word (by operation~\ref{scan}). Then,
operations~\ref{dm}~and~\ref{complete} are used to apply the grammar
rules using the chart, in an unspecified order. We prove below that
the process is indeed analogous to parsing $w$ with respect to $G$.

\begin{theorem}
$T_{G,w}$ is monotone: if $I_1 \subseteq I_2$ then $T_{G,w}(I_1) \subseteq
T_{G,w}(I_1)$.
\end{theorem}
\proof{}
Suppose $I_1 \subseteq I_2$. If $x \in
T_{G,w}(I_1)$ then $x$ was added by one of the five operations;
operations~\ref{predict},~\ref{fact}~and~\ref{scan} add the same
items every time $T_{G,w}$ is applied, and thus $x \in T_{G,w}(I_2)$,
too. If $x$ was added by operation~\ref{dm}, then there exist items
$\alpha, \beta$ in $I_1$ to which this operation applies. Since $I_1
\subseteq I_2$, $\alpha, \beta$ are in $I_2$, too, and hence $x \in
T_{G,w}(I_2)$, too. The same applies for operation~\ref{complete}. In
any case, $x \in T_{G,w}(I_2)$ and hence $T_{G,w}(I_1) \subseteq
T_{G,w}(I_2)$.

\begin{theorem}
$T_{G,w}$ is continuous: if $I_i, i \ge 0$ is a chain, then
$T_{G,w}(\bigcup_i I_i) = \bigcup_i T_{G,w}(I_i)$.
\end{theorem}
\proof{}
First, $T_{G,w}$ is monotone. Second, let $I = I_0 \subseteq I_1
\subseteq \ldots$ be a chain of items. If $x \in T_{G,w}(\bigcup_{i
\ge 0} I_i)$ then there exist $\alpha,\beta \in \bigcup_{i \ge 0} I_i$
as required, due to which $x$ is added. Then there exist $i,j$ such
that $\alpha \in I_i$ and $\beta \in I_j$. Let $k$ be the maximum of
$i,j$. Then $\alpha,\beta \in I_k$, $x \in T_{G,w}(I_k)$ and hence $x
\in \bigcup_{i \ge 0} T_{G,w}(I_i)$.\\ If $x \in \bigcup_{i \ge 0}
T_{G,w}(I_i)$ then there exists some $i$ that $x \in T_{G,w}(I_i)$.
$I_i \subseteq \bigcup_{i \ge 0} I_i$ and since $T_{G,w}$ is monotone,
$T_{G,w}(I_i) \subseteq T_{G,w}(\bigcup_{i \ge 0} I_i)$, and hence $x
\in T_{G,w}(\bigcup_{i \ge 0} I_i)$. Therefore $T_{G,w}$ is continuous.

\begin{corollary}
The least fix-point of $T_{G,w}$ can be obtained by iteratively computing
$I_{m+1} = T_{G,w}(I_m)$, starting from $I_0 = \emptyset$ and stopping when
a fix-point is reached.
\end{corollary}
\proof{ By Tarski-Knaster theorem, the lfp exists since $T_{G,w}$ is
monotone; By Kleene's theorem, since  $T_{G,w}$ is continuous, the lfp
can be obtained by applying the operator iteratively, starting from $\emptyset$.}

\begin{definition}[Algebraic meaning]
The {\bf meaning} of a grammar $G$ with respect to an input sentence
$w$ is the least fix-point of the operator $T_{G,w}$.
\end{definition}

\begin{definition}[Computation]
The {\bf $w$-computation} triggered by $w \in \words^{*}$ is the
infinite sequence of sets of items $I_i, i \ge 0$, such that $I_0 =
\emptyset$ and for every $m \ge 0$, $I_{m+1} = T_{G,w}(I_m)$. The
computation is {\bf terminating} if there exists some $m \ge 0$ for
which $I_m = I_{m+1}$ (i.e., a fix-point is reached in finite
time). The computation is {\bf successful} if there exists some $m$
such that $[0,A,n,\comp] \in I_m$, where $\len{A} = 1$ and $A \unif
Abs(A_s) \neq \top$;
otherwise, the computation fails.
\end{definition}
Notice that we check whether the generated items are {\em unifiable}
with the initial symbol, in accordance with the definition of {\em
languages}. If the initial symbol of the grammar is interpreted
differently when languages are defined, a corresponding modification
has to be made in the condition for {\em success}.

\section{Proof of Correctness}
\label{correctness}
In this section we show that parsing, as defined above, is (partially)
correct. First, the algorithm is {\em sound}: a $w$-computation
succeeds only if $w \in L(G)$; second, it is {\em complete}: if $w \in
L(G)$, it triggers a successful $w$-computation.  Computations are not
guaranteed to terminate, but we show that termination is assured for a
certain subset of the grammars that are {\em off-line parsable}.  We
discuss off-line parsability in section~\ref{olp}.

\subsection{Soundness}
In what follows we fix a particular $w$-computation $I_0, I_1, \ldots$,
triggered by some input $w=w_1 \cdots w_n$.
\begin{lemma}
\label{lemma:length1}
If $[i,A,j,\comp] \in I_l$ for some $l$ then $\len{A} = 1$.
\end{lemma}
\proof{}
By definition of $T_{G,w}$, complete items are generated by
operations~\ref{complete},~\ref{predict}~and~\ref{fact}. All these
operations add items in which the AMRS is of length~1.

\begin{lemma}
\label{lemma:rexists}
If $[i,A,j,\act] \in I_l$ for some $l$ and $\len{A} =k > 0$ then
there exists $\rho \in \rules$ such that $Abs(\rho)^{1 \ldots
k} \subsumes A$.
\end{lemma}
\proof{ By induction on $l$.}
If $l=0$ then $I_l = \emptyset$ and the proposition holds vacuously.
Assume that the proposition holds for every $l' < l$. Suppose that $x =
[i,A,j,\act] \in I_l$ and $A \neq \emptymrs$. Then $x$ must have been
added by operation~\ref{dm} (dot movement). Then $x = [i_{\alpha},C^{1
\ldots k+1},j_{\beta},\act]$ where $C = ((R,\{1 \ldots k\}) \unif
A_{\alpha}),\{k+1\}) \unif A_{\beta}$, namely $C \subsumed R$ and thus $C^{1
\ldots k+1} \subsumed R^{1 \ldots k+1}$.

\begin{theorem}[Parsing invariant (a)]
If $[i,A,j,c] \in I_l$ and $i<j$ then there exist $B \in PT_w(i+1,j)$
and $A' \subsumed B$
such that $A \derivess A'$. If $i=j$ then $A \derivess \emptymrs$.
\end{theorem}
\proof{ By induction on $l$.}\\
If $l=0$ then $I_l = \emptyset$ and the proposition holds vacuously.\\
Assume that the proposition holds for every $l' < l$. Suppose that $x =
[i,A,j,c] \in I_l$. Then $x$ must have been added by one of the
operations. Consider each case separately:
\begin{description}
\item[\ref{dm}. dot movement:]
$x = [i_{\alpha},C^{1 \ldots k+1},j_{\beta},\act]$ where there exist
$\alpha,\beta \in I_{l-1}$ as required and $C = (B,\{k+1\}) \unif
A_{\beta}$, $B = (R,\{1 \ldots k\}) \unif A_{\alpha}$. By the
induction hypothesis, there exist $A'_{\alpha},B_{\alpha}$ such that
$A_{\alpha} \derivess A'_{\alpha}$ and $A'_{\alpha} \subsumed
B_{\alpha} \in PT(i_{\alpha} + 1, j_{\alpha})$. Also, there exist
$A'_{\beta},B_{\beta}$ such that $A_{\beta} \derivess A'_{\beta}$ and
$A'_{\beta} \subsumed B_{\beta} \in PT(j_{\alpha} + 1, j_{\beta})$.
$B^{1 \ldots k} = (R,\{1 \ldots k \}) \unif A_{\alpha}$; if $k>0$,
$A_{\alpha} \neq \emptymrs$ and by lemma~\ref{lemma:rexists}
$A_{\alpha} \subsumed R$, hence $B^{1 \ldots k} = A_{\alpha}$. If
$k=0$, $B^{1 \ldots k} =\emptymrs = A_{\alpha}$. Hence $B^{1..k}
\derivess A'_{\alpha}$. $C^{1..k} \subsumed B^{1..k}$, and by
lemma~\ref{lemma:sspecify} there exists $A''_{\alpha} \subsumed
A'_{\alpha}$ such that $C^{1..k} \derivess A''_{\alpha}$. In the same
way, there exists $A''_{\beta} \subsumed A'_{\beta}$ such that
$C^{k+1} \derivess A''_{\beta}$. By lemma~\ref{lemma:concat}, $C^{1
\ldots k+1} \derivess A''_{\alpha} \cdot A''_{\beta}$. 
But $A''_{\alpha} \cdot A''_{\beta} \subsumed A'_{\alpha} \cdot
A'_{\beta} \subsumed B_{\alpha} \cdot B_{\beta}$, and since
$B_{\alpha} \in PT(i_{\alpha} + 1, j_{\alpha})$ and $B_{\beta} \in
PT(j_{\alpha} + 1, j_{\beta})$, by lemma~\ref{lemma:pt-concat} $B_{\alpha}
\cdot B_{\beta} \in PT(i_{\alpha} + 1, j_{\beta})$.
The cases in which $i_{\alpha} = j_{\alpha}$ or
$i_{\beta} = j_{\beta}$ are trivial.
\item[\ref{complete}. completion:]
$x = [i_{\alpha},C^{m},j_{\alpha},\comp]$ where $C = (R, \{1 \ldots
m-1\}) \unif A_{\alpha}$ and there exist an abstract rule $R$ and an item
$\alpha \in I_{l-1}$ as required, and (by lemma~\ref{lemma:rexists})
$A_{\alpha}^{1 \ldots m - 1} \subsumed R^{1 \ldots m - 1}$. If
$i_{\alpha} < j_{\alpha}$ then by the induction hypothesis, there
exist $A'_{\alpha},B_{\alpha}$ such that $A_{\alpha} \derivess
A'_{\alpha}$ and $A'_{\alpha} \subsumed B_{\alpha} \in PT(i_{\alpha} +
1, j_{\alpha})$. $C = (R, \{1 \ldots m-1\}) \unif A_{\alpha}$, hence
$C^{1 \ldots m-1} = A_{\alpha}$ and thus $C^{1 \ldots m-1} \derivess
A_{\alpha}$.
From lemma~\ref{lemma:complete}, $C^m \derives C^{1 \ldots m-1}$, and
thus $C^m \derivess A'_{\alpha}$. 
If $i_{\alpha} = j_{\alpha}$ then $A_{\alpha} \derivess \emptymrs$ and
hence $C^m \derivess \emptymrs$.
\item[\ref{predict}. prediction:]
$x = [i,\emptymrs,i,\act]$ and $PT(i+1,i) = \emptyset$.
\item[\ref{fact}. $\epsilon$-rules:]
$x = [i,Abs(\rho),i,\comp]$ and $PT(i+1,i) = \emptyset$.
\item[\ref{scan}. scanning:]
$x = [i-1,Abs(A_i),i,\comp]$ where $A_i \in Cat(w_i)$, and $Abs(A_i)
\derivess Abs(A_i)$ trivially. $Abs(A_i) \in PT(i+1,j)$ by definition.
\end{description}

\begin{theorem}
If a computation, triggered by $w$, is successful, then $w \in L(G)$.
\end{theorem}
\proof{}
If a computation is successful then there exists some $m \ge 0$ such
that $x=[0,A,n,\comp] \in I_m$ where $\len{A} = 1$ and $A \unif Abs(A_s)
\neq \top$. By the parsing invariant, $A \derivess A'$ for some $A'
\subsumed B \in PT_w(1,n)$. Hence $Abs(A_s) \aderives B$ and $w \in L(G)$.

\subsection{Completeness}
The following theorem shows that one derivation step, licensed by a rule
$Abs(\rho)$ of length $r$, corresponds to $r+1$ applications of $T_{G,w}$,
starting with an item that predicts the rule and advancing the dot $r$
times, until a complete item for that rule is generated.

\begin{theorem}[Parsing invariant (b)]
If $A \derivess A'$ and $A' \subsumed B \in PT_w(i+1,j)$ then for
every $k$, $0 < k \le \len{A}$, there exists $l_k$ such that
$[i_k,C_k,j_k,\comp] \in I_{l_k}$, where $C_k \subsumes A^k$, $i_1 =
i$, $j_{\len{A}} = j$ and $j_k = i_{k+1}$ if $k<\len{A}$.
\end{theorem}
\proof{}
By induction on $l$, the number of derivation steps from $A$ to $A'$:\\ 
If $l=0$, $A = A' \subsumed B$. Since $B \in PT_w(i+1,j)$, $B =
Abs(A_{i+1}) \cdot \ldots \cdot Abs(A_j)$ where $A_k \in Cat(w_k)$ for
$i+1 \le k \le j$. The scanning operation (\ref{scan}) of $T_{G,w}$
adds appropriate items whenever it is applied. \\ Assume that $A
\derives D \derivess B \subsumed PT_w(i+1,j)$ and the proposition
holds for $D$ and $B$. By the induction hypothesis, for every $k$, $0
< k \le \len{D}$, there exists $l_k$ such that $[i_k,C_k,j_k,\comp]
\in I_{l_k}$, where $C_k \subsumes A^k$.  Suppose that $A \derives D$
through a rule $\rho$ of length $r$ by expanding $A^y$ to $D^{x \ldots
x+r-1}$. Then the following sequence of items is generated, where for
every $m$, $C_{1 \ldots m}
\subsumes D^{x \ldots x+m-1}$, and $C_r \subsumes A^y$:

\begin{tabular}{llll}
$[i,\emptymrs,i,\act]$	& $\in$	& $I_1$	& by prediction (\ref{predict})\\
$[i_1,C_1,j_1,\comp]$	& $\in$	& $I_{l_1}$	& by the induction hypothesis\\
$[i,C_1,j_1,\act]$	& $\in$	& $I_{l_1}$	& by dot movement (\ref{dm})\\
$[i_2,C_2,j_2,\comp]$	& $\in$	& $I_{l_2}$	& by the induction hypothesis\\
$[i,C_{1 \ldots 2},j_2,\act]$	& $\in$	& $I_{max(l_1,l_2)}$	& by dot movement (\ref{dm})\\
\vdots\\
$[i,C_{1 \ldots r-1},j_{r-1},\act]$	& $\in$	& $I_{max(l_1,\ldots,l_{r-1})}$	& by dot movement (\ref{dm})\\
$[i,C_{r},j,\comp]$	& $\in$	& $I_{max(l_1,\ldots,l_{r-1})+1}$ & by completion (\ref{complete})
\end{tabular}

Items are generated by the dot movement~(\ref{dm}) operation since
the conditions for its application obtain: it is easy to see that the
indices ($i,j$) match; in addition, if for some $m$, $C_{1 \ldots m}
\subsumes D^{x \ldots x+m-1}$, and for every $k$, $C_k \subsumes A^k$,
then there exists $C_{1 \ldots m+1}$ that is obtained by unifying
some $R \subsumed Abs(\rho)$ first with $C_{1 \ldots m}$ and then with
$C^{m+1}$, such that $C_{1 \ldots m+1} \subsumes D^{x \ldots x+m}$ as
required. Therefore, by induction on $m$ it can be shown that all the
items that result from dot movement are indeed generated. Finally, the
completion~(\ref{complete}) operation is applicable and (since $A
\derives D$) we have $C_r \subsumes A^y$.

\begin{theorem}
If $w \in L(G)$ then the computation triggered by $w$ is successful.
\end{theorem}
\proof{}
$w \in L(G)$, hence $Abs(A_s) \aderives B$, where $B \in PT_w(1,n)$. Hence
there exist $A',B'$ such that $A' \unif Abs(A_s) \neq \top, B' \subsumed B$
and $A' \derivess B'$. By the parsing invariant, there exists $l$ such
that $[0,C,n,\comp] \in I_l$ where $C \subsumes A'$. Hence $C \unif
Abs(A_s) \neq \top$, and therefore the computation is successful.

\subsection{Subsumption Check}
\label{subsumption-test}
To assure efficient computation and eliminate redundant items, many
parsing algorithms employ a mechanism called {\em subsumption check}
(see, e.g.,~\cite{shieber92,sikkel}) to filter out certain generated
items. We introduce this mechanism below and show that it doesn't
effect the correctness of the computation.

Define a (partial) order over items: $[i_1,A_1,j_1,c_1] \preceq
[i_2,A_2,j_2,c_2]$ iff $i_1=i_2,j_1=j_2, c_1 = c_2$ and $A_1 \subsumes
A_2$. Modify the ordering on sets of items as follows: $I_1 \preceq
I_2$ iff for every $x_1 \in I_1$ there exists $x_2 \in I_2$ such that
$x_2 \preceq x_1$. Sets of items are no longer ordered by inclusion,
but rather by a weaker condition that only requires the existence of a
more general item (in the higher set) for every item (in the lower set).

The subsumption filter is realized by modifying $T_{G,w}$: $x \in
T_{G,w}(I)$ only if there does not exist any item $x' \in T_{G,w}(I)$
such that $x' \preceq x$. Namely, for all items that span the same
substring and have the same status (\act\ or \comp), only the most
general one is preserved across successive applications of $T_{G,w}$.
Given the new ordering of sets of items, it can be shown that this
modification does not harm neither monotonicity nor continuity, and
hence every computation is guaranteed to reach a least
fix-point. Obviously, the soundness of the computation is also
maintained.  More interestingly, completeness is preserved, too:
recall that the parsing invariant (b) states that if $A \derivess A'
\subsumed B$ then for every $k$ some item $[i_k,C_k,j_k,\comp]$ is
generated such that $C_k \subsumes A^k$. Since the subsumption test
only leaves out an item if a more general one exists, the invariant
still holds and hence the correctness of the computation is
guaranteed. Notice that if $L(G)$ would have been defined as the set
of strings that are derivable from the start symbol itself, the
subsumption check might have removed crucial items, and the computation
could cease to be correct.

\subsection{Termination}
\label{olp}
It is well-known (see, e.g.,~\cite{parsing-as-deduction,johnson88})
that unification-based grammar formalisms are Turing-equivalent, and
therefore decidability cannot be guaranteed in the general case.
However, for grammars that satisfy a certain restriction, termination
of the computation can be proven.  We make use of the well-foundedness
result (section~\ref{sec:well-founded}) to prove that parsing is
terminating for {\em off-line parsable} grammars.

{\em Off-line parsability} was introduced by~\cite{lfg} and adopted
by~\cite{parsing-as-deduction}, according to which ``A grammar is
off-line parsable if its context-free skeleton is not infinitely
ambiguous''. As~\cite{johnson88} points out, this restriction
(defined in slightly different terms) ``ensures that the number of
constituent structures that have a given string as their yield is
bounded by a computable function of the length of that string''.  The
problem with this definition is demonstrated by~\cite{haas}: ``Not every
natural unification grammar has a context-free backbone''. 

A context-free backbone is inherent in LFG, due to the separation of
c-structure from f-structure and the explicit demand that the
c-structure be context-free.  However, this notion is not well-defined
in HPSG, where phrase structure is encoded within feature structures
(indeed, HPSG itself is not well-defined in the formal language
sense).  Such a backbone is certainly missing in Categorial Grammar,
as there might be infinitely many categories. \cite{shieber92}
generalizes the concept of off-line parsability but doesn't prove that
parsing with off-line parsable grammars is terminating. We use an
adaptation of his definition below and provide a proof.

To overcome this problem, \cite{haas} uses a different restriction:
``A grammar is depth-bounded if for every $L>0$ there is a $D>0$ such
that every parse tree for a sentential form of $L$ symbols has depth
less than $D$''. \cite{shieber92} generalizes it and we use an
adaptation of his definition below.
\begin{definition}[Finite-range decreasing functions]
A function $F:D\rightarrow D$, where $D$ is a partially-ordered
set, is {\bf finite-range decreasing} (FRD) iff the range of $F$
is finite and for every $d\in D,F(d)\preceq d$.
\end{definition}

\begin{definition}[Strong off-line parsability]
A grammar is {\bf strongly off-line parsable} iff there exists an
FRD-function $F$ from AMRSs to AMRSs (partially ordered by subsumption)
such that for every string $w$ and different AMRSs $A,B$ such
that $A \derives B$, if $A \derives PT_w(i+1,j)$ and
$B \derives PT_w(i+1,j)$ then $F(A) \neq F(B)$.
\end{definition}
Strong off-line parsability guarantees that any particular sub-string
of the input can only be spanned by a finite number of AMRSs:
if a grammar is strongly off-line parsable, there can not exist an
infinite set $S$ of AMRSs, such that for some $0 \le i \le j \le |w|$,
$s \derives PT_w(i+1,j)$ for every $s\in S$.  If such a set existed,
$F$ would have mapped its elements to the set $\{F(s) \mid s \in
S\}$. This set is infinite since $S$ is infinite and $F$ doesn't map
two different items to the same image, and thus the finite
range assumption on $F$ is contradicted.

As \cite{shieber92} points out, ``there are non-off-line-parsable
grammars for which termination holds''.  We use below a more general
notion of this restriction: we require that $F$ produce a different
output on $A$ and $B$ only if they are incomparable with
respect to subsumption. We thereby extend the class of grammars for
which parsing is guaranteed to terminate (although there still remain
decidable grammars for which even the weaker restriction doesn't hold).

\begin{definition}[Weak off-line parsability]
A grammar $G$ is {\bf weakly off-line parsable} iff there exists an
FRD-function $F$ from AMRSs to AMRSs (partially ordered by subsumption)
such that for every string $w$ and different AMRSs $A,B$ such
that $A \derives B$, if $A \derives PT_w(i+1,j)$,
$B \derives PT_w(i+1,j)$, $A \not\sqsubseteq B$ and
$B \not\sqsubseteq A$, then $F(A) \neq F(B)$.
\end{definition}
Clearly, strong off-line parsability implies weak off-line parsability.
However, as we show below, the inverse implication does not hold.

We now prove that weakly off-line parsable grammars guarantee
termination of parsing in the presence of acyclic AMRSs. We prove that
if these conditions hold, only a {\em finite} number of different items can
be generated during a computation. The main idea is the following: if
an infinite number of different items were generated, then an infinite
number of different items must span the same sub-string of the input
(since the input is fixed and finite). By the parsing invariant, this
would mean that an infinite number of AMRSs derive the same sub-string
of the input. This, in turn, contradicts the weak off-line parsability
constraint.

\begin{theorem}
\label{theorem:olp}
If $G$ is weakly off-line parsable and AMRSs are acyclic then every
computation terminates.
\end{theorem}
\proof{}
Fix a computation triggered by $w$ of length $n$. We claim that there
is only a finite number of generated items. Observe that the
indices that determine the span of items are bounded: $0 \le i \le j
\le n$. It remains to show that only a finite number of AMRSs are
generated. Let $x=[i,A,j,c]$ be a generated item. Suppose another item
is generated where only the AMRS is different: $x'=[i,B,j,c]$ and $A
\neq B$. If $A \sqsubseteq B$, $x'$ will not be preserved because of
the subsumption test.  There is only a finite number of AMRSs $A'$
such that $A' \sqsubseteq A$ (since subsumption is well-founded for
acyclic AMRSs). Now suppose $A \not\sqsubseteq B$ and $B
\not\sqsubseteq A$.  By the parsing invariant (a) there exist $A',B'$ such
that $A \derivess A' \in PT_w(i+1,j)$ and $B \derivess B' \in
PT_w(i+1,j)$.  Since $G$ is off-line parsable, $F(A) \neq F(B)$. Since
the range of $F$ is finite, there are only finitely many items with
equal span that are pairwise incomparable.  Since only a finite number
of items can be generated and the computation uses a finite number of
operations, the least fix-point is reached within a finite number of steps.

The above proof relies on the well-foundedness of subsumption, and
indeed termination of parsing is not guaranteed by weak off-line
parsability for grammars based on cyclic TFSs. Obviously, cycles can
occur during unification even if the unificands are acyclic. However,
it is possible (albeit costly, from a practical point of view) to spot
them during parsing. Indeed, many implementations of logic programming
languages, as well as of unification-based grammars (e.g.,
ALE~\cite{ale}) do not check for cycles. If cyclic TFSs are allowed,
the more strict notion of strong off-line parsability is needed. Under
the strong condition the above proof is applicable for the case of
non-well-founded subsumption as well.

To exemplify the difference between strong and weak off-line
parsability, consider a grammar $G$ that contains 
the following single rule:
\[
\tag{1}\begin{tfs}{t}f: & \bot\end{tfs}
\Rightarrow
\begin{tfs}{t}f: & \tag{1}\end{tfs}
\]
and the single lexical entry, $w_1$, whose category is:
\[
Cat(w_1)= \begin{tfs}{t}f: & \bot\end{tfs}
\]
This lexical entry can be derived by an infinite number of TFSs:
\[
\ldots \derives
\begin{tfs}{t}f: & \begin{tfs}{t}f: & \begin{tfs}{t}f: & \bot\end{tfs}\end{tfs}\end{tfs} \derives
\begin{tfs}{t}f: & \begin{tfs}{t}f: & \bot\end{tfs}\end{tfs} \derives
\begin{tfs}{t}f: & \bot\end{tfs} =
Cat(w_1)
\]
%
It is easy to see
that no FRD-function can distinguish (in pairs) among these TFSs, and
hence the grammar is not strongly off-line parsable. 
The grammar
is, however, {\em weakly\/} off-line parsable: since the TFSs that
derive each lexical entry form a subsumption chain, the antecedent of
the implication in the definition for weak off-line parsability never
holds; even trivial functions such as the function that returns the
empty TFS for every input are appropriate FRD-functions. Thus parsing
is guaranteed to terminate with this grammar.

It might be claimed that the example rule is not a part of any grammar
for a natural language. It is unclear whether the distinction between
weak and strong off-line parsability is relevant when ``natural''
grammars are concerned. Still, it is important when the formal,
mathematical and computational properties of grammars are
concerned. We believe that a better understanding of formal properties
leads to a better understanding of ``natural'' grammars as well.
Furthermore, what might be seem un-natural today can be common practice
in the future.


\chapter{\amalia\ -- An Abstract Machine for Linguistic Applications}
\label{sec:amalia}
This chapter details the design and implementation of the abstract
machine. Section~\ref{framework} presents the formal language in which
specifications are input, including the type hierarchy, the grammar
and the lexicon. The machine is explained incrementally:
section~\ref{m0} describes its core engine, dedicated to the
unification of two feature structures. The engine is enveloped with
control structures to accommodate for parsing in
section~\ref{machine-parsing}. Optimizations and extensions are
discussed in section~\ref{sec:optimizations}, and the actual implementation
of the machine, including some comparative performance analyses, are
given in section~\ref{sec:implementation}.

\section{The Input Language}
\label{framework}
\subsection{Type Specification}
A program (i.e., a grammar) must specify the type hierarchy and the
appropriateness specification. We adopt ALE's syntax \cite{ale} for
this specification: it is a sequence of statements of the form:
\begin{center}
$t$ {\mf sub} $[t_1,t_2,\ldots,t_n]$ {\mf intro} $[f_1:r_1,\ldots,f_m:r_m]$.
\end{center}
where ${t,t_1,\ldots,t_n,r_1,\ldots,r_m}$ are types,
$f_1,\ldots,f_m\;$ are features and $n,m \ge 0$.  If $m=0$, the
`intro' part is omitted.  This statement, which is said to {\em
characterize} $t$, means that ${t_1,\ldots,t_n}$ are all -- and the
only -- (immediate) subtypes of ${t}$ (i.e., for every $i, 1 \le i \le
n, {t} \subsumes {t_i}$), and that ${t}$ has the features
$f_1,\ldots,f_m$ appropriate for it. Moreover, these features are {\em
introduced} by $t$, i.e., they are not appropriate for any type $t'$
such that $t' \sqsubset t$. Finally, the statement specifies that
$Approp(t,f_i) = r_i$ for every $i$.  Each type (except $\top$ and
$\bot$) must be characterized by exactly one statement. The {\em
arity} of a type $t$, $Ar(t)$, is the number of features appropriate
for it.

The full subsumption relation is the reflexive transitive closure of
the immediate relation determined by the characterization
statements. If this relation is not a bounded complete partial order,
the specification is rendered invalid. The same is true in case it is
not an appropriateness specification.

We use the type hierarchy in figure~\ref{hier} as a running example, where
\verb+bot+ stands for $\bot$. The type $\top$ 
is systematically omitted from type specifications.
\begin{figure}[hbt]
\begin{minipage}{6.2cm}
\begin{verbatim}
bot sub [g,d].
  g sub [a,b] intro [f3:d].
    a sub [c] intro [f1:bot].
      c sub [] intro [f4:bot].
    b sub [c,e] intro [f2:bot].
      e sub [].
  d sub [d1,d2].
    d1 sub [].
    d2 sub [].
\end{verbatim}
\end{minipage} \hfill
\begin{minipage}{5.8cm}
\setlength{\unitlength}{0.012500in}%
\begin{picture}(160,131)(280,440)
\put(320,475){\line( 2,-1){ 40}}
\put(415,475){\line(-2,-1){ 40}}
\put(285,515){\line( 3,-4){ 15}}
\put(335,515){\line(-1,-2){ 10}}
\put(300,555){\line(-3,-4){ 15}}
\put(320,555){\line( 3,-4){ 15}}
\put(375,555){\line(-3,-4){ 15}}
\put(410,515){\line( 1,-4){  5}}
\put(440,515){\line(-2,-5){ 10}}
\put(360,440){\makebox(0,0)[lb]{\smash{\mf}bot}}
\put(280,520){\makebox(0,0)[lb]{\smash{\mf}a[f1:bot]}}
\put(335,520){\makebox(0,0)[lb]{\smash{\mf}b[f2:bot]}}
\put(300,560){\makebox(0,0)[lb]{\smash{\mf}c[f4:bot]}}
\put(380,560){\makebox(0,0)[lb]{\smash{\mf}e}}
\put(400,520){\makebox(0,0)[lb]{\smash{\mf}d1}}
\put(440,520){\makebox(0,0)[lb]{\smash{\mf}d2}}
\put(415,480){\makebox(0,0)[lb]{\smash{\mf}d}}
\put(300,480){\makebox(0,0)[lb]{\smash{\mf}g[f3:d]}}
\end{picture}
\end{minipage}
\mycaption{An example type hierarchy}{המגודל םיסופיט תייכראריה}
\label{hier}
\end{figure}

\subsection{Rules and Grammars}
A {\em grammar} consists of a non-empty set of {\em rules} and a set
of {\em lexical entries\/} which associate a feature structure with
every word. Each rule is a sequence of feature structures of length
greater than 1, with possible reentrancies among its elements, and a
designated (last) element that is the rule's {\em head}. The rest of
the elements in a rule form its {\em body}. We use multi-terms for
representing rules and lexical entries; however, we employ a simple
description language in which such terms are expressible, compatible
to the ALE input language \cite{ale}. 

ALE's description language for feature structures is based on a logical
language developed by~\namecite{kasperounds}, extended to accommodate
types, where path sharing is replaced by the notion of variables due
to~\namecite{smolka}. A TFS is described as a conjunction of
specifications that might include its type, its features, along with
their values, or a variable (whose name begins with a capitalized
letter) that refers to it. Multiple occurrences of the same variable
denote reentrancy. The syntax for specifying rules consists of a rule
name, the reserved word `\verb+rule+', a description for the rule's
head, the reserved symbol `\verb+===>+' and then the elements of the
rule's body, each preceded by the reserved word `\verb+cat>+'. Lexical
items consist of the word itself, the symbol `\verb+--->+' and a
description. ALE's descriptions can include several other features that
are not supported by \amalia, most notably disjunctions and inequations.
Refer to~\namecite{ale} for a formal definition of ALE's input language.

As an example, the specification of the example grammar of
figure~\ref{fig:grammar} is depicted in figure~\ref{fig:ale-grammar}
below. Notice that ALE's syntax places the head of rules {\em
before\/} the body. Comments are preceeded by `\verb+%+'.
\begin{figure}[hbt]
\begin{verbatim}
%%%**********************  Grammar Rules
%grammar

s_np_vp rule
(phrase,cat:s,agr:Agr,sem:(Sem,arg1:SubjSem))
===>
cat> (cat:(n,case:nom),agr:Agr,sem:pred:SubjSem),
cat> (cat:v,agr:Agr,sem:Sem).

np_v_np rule
(phrase,cat:v,agr:Agr,sem:Sem,sem:arg2:ObjSem)
===>
cat> (cat:v,agr:Agr,sem:Sem),
cat> (cat:(n,case:acc),sem:pred:ObjSem).

%%%**********************  Lexical Entries
%lexicon

john --->
(word,cat:n,agr:(per:third,num:sg),sem:pred:john).

her --->
(word,cat:(n,case:acc),agr:(per:third,num:sg),sem:pred:she).

loves --->
(word,cat:v,agr:(per:third,num:sg),sem:pred:love).
\end{verbatim}
\mycaption{An example grammar in ALE format}{המגודל קודקד}
\label{fig:ale-grammar}
\end{figure}

\section{A TFS Unification Engine}
\label{m0}
\subsection{First-Order Terms vs.\ Feature Structures}
\label{sec:fs-vs-fot}
While TFSs resemble first-order terms (FOTs) in many aspects, it is
important to note the differences between them. Most importantly,
while FOTs are essentially trees, with possibly shared leaves, TFSs
are directed graphs, within which variables can occur anywhere.
Moreover, our system doesn't rule out cyclic structures, so that
infinite terms can be represented, too.  Two FOTs are mutually
consistent only if they have the same functor and the same
arity. TFSs, on the other hand, can be unified even if their types
differ (as long as they have a non-degenerate LUB). Moreover, their
arity can differ, and the arity of the unification result can be
greater than that of any of the unificands.  Consequently, many
diversions from the original WAM were necessary in our design. In the
following sections we try to emphasize the points where such
diversions were made. We assume familiarity with basic WAM concepts in
this section.

\subsection{Processing Scheme}
\amalia's engine is designed for unifying two TFSs: a {\em
program} and a {\em query}.  Many queries (representing input Natural
Language phrases) can be executed with respect to a given program (the
grammar of the Natural Language). The program is compiled only once to
produce machine instructions.  Each query is compiled before its
execution; the resulting code is executed prior to the execution of
the compiled program.  Execution of the instructions, produced by
compiling a query, builds a graph representation of the feature
structure denoted by the query in the machine's memory. The processing
of a program produces code that, during run-time, unifies the feature
structure denoted by the program with a query already resident in
memory.  The result of the unification is a new TFS, represented as a
graph in the machine's memory.  In what follows we interleave the
description of the machine, the TFS language it is designed for and
the compilation of programs in this language. In
section~\ref{machine-parsing}, queries are extended to sequences of TFSs,
representing input strings, and programs are extended to sets of
rules, i.e., grammars.

\subsection{Memory Representation of Feature Structures}
The major operation of \amalia's engine is feature structure unification;
therefore, the major data structure of the machine is aimed at storing
feature structures in an efficient way.  Following the WAM, we use a
global, one-dimensional array of data cells called {\mf HEAP}.  A
global register {\mf H} points to the top element of {\mf HEAP}.  Data
cells are tagged: STR cells represent nodes, and store their types,
while REF cells represent arcs, and contain the address of their
targets.\footnote{A third tag of heap cells, VAR, in introduced in
section~\ref{sec:var-cells}.} The number of arcs leaving a node of
type $t$ is $Ar(t)$, fixed due to total well-typedness.  Hence, we can
keep the WAM's convention of storing all the outgoing arcs from a node
consecutively following the node. Given a type $t$ and a feature $f$
that is appropriate for $t$, the position of the arc corresponding to
$f$ ($f$-arc) in any TFS of type $t$ can be statically determined; the
value of $f$ can be accessed in one step.  This is a major difference
from the approach presented in \cite{prl7}; it leads to a more
time-efficient system without harming the elegance of the machine
design.

Computations performed on the machine involve TFS unification, during
which new structures are built on top of the heap and existing
structures might be modified.  It is important to note that STR cells
differ from their WAM analogs in that they can be dereferenced when a
type is becoming more specific. In such cases, a chain of REF cells
leads to the dereferenced STR cell.  Thus, if a TFS is modified, only
its STR cell has to be changed in order for all pointers to it to
`feel' the modification automatically. The use of self-referential REF
cells is different, too: there are no real (Prolog-like) variables in
our system, and such cells stand for features whose values are
temporarily unknown.

One cell is required for every node and arc, so for representing a
graph of $n$ nodes and $m$ arcs, $n+m$ cells are needed. Of course,
during unification nodes can become more specific and a chain of REF
cells is added to the count, but the length of such a chain is bounded
by the depth of the type hierarchy and path compression during
dereferencing cuts it occasionally. As an example, figure~\ref{heap1}
depicts a possible heap representation of the TFS
\nterm{b(b(\tag{1}d,\tag{1}),d)}.

\begin{figure}[hbt]
\centering
\begin{tabular}{|l|c|c|c|c|c|c|c|c|} \hline
        address: & 1 & 2 & 3 & 4 & 5 & 6 & 7 & 8 \\ \hline
        tag:     & STR&REF&REF&STR&REF&REF&STR&STR\\ \hline
        contents:& b & 4 & 8 & b & 7 & 7 & d & d \\ \hline
\end{tabular}
        \mycaption{Heap representation of the TFS \nterm{b(b(\tag{1}d,\tag{1}),d)}}{ןורכיזב תוינוכת הנבמ לש גוציי}
        \label{heap1}
\end{figure}

\subsection{Flattening Feature Structures}
Before processing a TFS, its linear representation is transformed to a
set of ``equations'', each having a flat (nesting free) format, using
a set of {\em registers} $\{X_i\}$ that store {\em addresses} of TFSs
in memory.  A register {\mf Reg[$j$]} is associated with each tag
\tag{j} of a normal term (recall that a term is normal only if all
its types are tagged).  The flattening algorithm is straight-forward
and similar to the WAM's. The order of the equations correspond to a
depth-first, postorder search of the term, where new registers are
allocated for sub-terms before the sub-terms are
processed. Figure~\ref{eqs} depicts examples of the equations
corresponding to two TFSs.

\begin{figure}[hbt]
\centering
\begin{tabular}{|c|l|}  \hline
Linear representation:  & Set of equations      \\ \hline
\nterm{a(\tag{3}d1,\tag{3})}    & $X1 = a(X2,X2)$       \\
                                & $X2 = d1$             \\ \hline
\nterm{b(b(\tag{1}d,\tag{1}),d)}        & $X1 = b(X2,X3)$       \\
                                & $X2 = b(X4,X4)$       \\
                                & $X4 = d$              \\
                                & $X3 = d$              \\ \hline
\end{tabular}
\mycaption{Feature structures as sets of equations}{תואוושמ לש הצובקכ תוינוכת הנבמ}
\label{eqs}
\end{figure}

\subsection{Processing of a Query}
When processing an equation of the form $X_{i_0} = t(X_{i_1}, X_{i_2},
\ldots)$, representing part of a query, two different kinds of
instructions are generated. The first is {\mf put\_node t/n,
$X_{i_0}$}, where $n = Ar(t)$. Then, for every argument $X_{i_j}$, an
instruction of the form {\mf put\_arc $X_{i_0}$, $j$, $X_{i_j}$} is
generated. Execution of the {\mf put\_node} instruction creates a
representation of a node of type $t$ on top of the heap and stores its
address in $X_{i_0}$; it also increments {\mf H} to leave space for
the arcs of the newly created node. Execution of the subsequent {\mf
put\_arc} instructions fills this space with REF cells.

In order for {\mf put\_arc} to operate correctly, the registers it
uses must be initialized. Since only {\mf put\_node} sets the
registers, one way of ensuring correctness is having all {\mf
put\_node} instructions executed before any {\mf put\_arc} instruction
is. Hence, the machine maintains two separate streams of instructions,
one for {\mf put\_node} and one for {\mf put\_arc}, and executes all
elements of the first before moving to the other. This compilation
scheme is called for by the cyclic character of TFSs: as explained in
\cite{prl7}, the original single-streamed WAM scheme would fail on
cyclic terms.  

The effect of the two instructions is given in figure~\ref{put-inst}.
We use syntax similar to that of~\cite{wam} for describing the effect
of instructions; in particular, the arguments of an instruction are
listed succeeding its mnemonic. We use `{\mf <STR,t>}' to denote an
STR cell of type $t$, and `{\mf <REF,a>}' to denote a REF cell
pointing to the address $a$.  Figure \ref{exmpl-code} lists the result
of compiling the term \nterm{b(b(\tag{1}d,\tag{1}),d)}. When this code
is executed (first the {\mf put\_node} instructions, then the {\mf
put\_arc} ones), the resulting representation of the TFS in memory is
the one shown above in figure~\ref{heap1}.

\begin{program}{put-inst}{The effect of the {\mf put} instructions --
{\mf put}}{
put\_node t/n,$X_i$ $\equiv$            \\
\label{inst:put-node}
                \> HEAP[H] \get <STR,t>;        \\
                \> $X_i$ \get H;        \\
                \> H \get H + n + 1;    \\
\\
put\_arc $X_i$,offset,$X_j$ $\equiv$            \\
\label{inst:put-arc}
                \> HEAP[$X_i$+offset] \get <REF,$X_j$>; \\
}
\end{program}
\begin{figure}[hbt]
{\mf
\begin{verbatim}
put_node b/2,X1        % X1 = b(
   put_arc X1,1,X2     %        X2,
   put_arc X1,2,X3     %           X3)
put_node b/2,X2        % X2 = b(
   put_arc X2,1,X4     %        X4,
   put_arc X2,2,X4     %           X4)
put_node d/0,X4        % X4 = d
put_node d/0,X3        % X3 = d
\end{verbatim}
}
\mycaption{Compiled code for the query \nterm{b(b(\tag{1}d,\tag{1}),d)}}{התליאש רובע הנוכמ תודוקפ}
\label{exmpl-code}
\end{figure}

\subsection{Compilation of the Type Hierarchy}
\label{compile-th}
One of the reasons for the efficiency of our implementation is that it
performs a major part of the unification during compile-time: the type
unification.  The WAM's equivalent of this operation is a simple
functor and arity comparison. It is due to the nature of a typed
system that this check has to be replaced by a more complex
computation.  Efficient methods were suggested for performing LUB
computation at run time, relying on efficient encoding of types
(see~\cite{lattice-ops}). We compute LUBs only once, at compile time,
using a simple transitive closure computation.  Since type unification
adds information by returning the features of the unified type, this
operation builds new structures, in our design, that reflect this
addition. Moreover, the WAM's special register S is here replaced by a
stack. S is used by the WAM to point to the next sub-term to be
matched against, but in our design, as the arity of the two terms can
differ, there might be a need to hold the addresses of more than one
such sub-term.  These addresses are stored in the stack (more details
and an example are given below).

When the type hierarchy is processed, the (full) subsumption relation
is computed and checked for bounded-completeness (using a
straight-forward implementation of Warshall
algorithm~\cite{warshall}).  Then, a table is generated which
stores, for every two types $t_1,t_2$, the least upper bound $t = t_1
\unif t_2$.  In addition, this table lists also the arity of $t$, its
features and their ``origin'': whether they are inherited from $t_1$,
$t_2$, both or none of them.  Figure~\ref{fig:tables} graphically
depicts the LUB and appropriateness tables generated for the running
example type hierarchy.
\begin{figure}[hbt]
\center
\begin{tabular}{||l|l|l|l|l|l|l|l|l|l||c|l||} \hline
$t$ & \multicolumn{9}{c||}{least upper bound} & $Ar$ & $Approp$ \\ \hline
     &$\bot$  &g    &d    &a    &b    &c    &e    &d1   &d2   & & \\ \hline
$\bot$  &$\bot$  &g    &d    &a    &b    &c    &e    &d1   &d2   & 0 &   \\ \hline 
g    &g    &g    &     &a    &b    &c    &e    &     &     & 1 &f3:d  \\ \hline
d    &d    &     &d    &     &     &     &     &d1   &d2   & 0 &    \\ \hline
a    &a    &a    &     &a    &c    &c    &     &     &     & 2 &f3:d,f1:$\bot$ \\ \hline
b    &b    &b    &     &c    &b    &c    &e    &     &     & 2 &f3:d,f2:$\bot$ \\ \hline
c    &c    &c    &     &c    &c    &c    &     &     &     & 4 &f3:d,f1:$\bot$,f4:$\bot$,f2:$\bot$   \\ \hline
e    &e    &e    &     &     &e    &     &e    &     &     & 2 &f3:d,f2:$\bot$ \\ \hline              
d1   &d1   &     &d1   &     &     &     &     &d1   &     & 0 &   \\ \hline
d2   &d2   &     &d2   &     &     &     &     &     &d2   & 0 &   \\ \hline
\end{tabular}
\mycaption{Type unification tables}{םיסופיט תדחאה תולבט}
\label{fig:tables}
\end{figure}

Out of this table a series
of abstract machine language functions are generated. The functions
are arranged as a two-dimensional array called {\mf unify\_type},
indexed by two types $t_1,t_2$. Each such function receives one
parameter, the address of a TFS on the heap. Recall that the machine
is designed to unify two feature structures, one of which is part of a
{\em program}, represented as code, and the other, which is part of
the {\em query}, already resides on the heap. Each {\mf unify\_type}
function receives the address of this second unificand as a
parameter. When executed, it builds on the heap a skeleton for the
unification result: an STR cell of the type $t_1 \unif t_2$, and a REF
cell for each appropriate feature of it.

Consider {\mf unify\_type[t1,t2](addr)} where {\mf addr} is the
address of some TFS, $A$ (of type $t_2$), in memory.  Let $t = t_1
\unif t_2$, and let $f$ be some feature appropriate for $t$.  If $f$
is inherited from $t_2$ only, the value of the REF cell in the
skeleton result is simply set to point to the $f$-arc in $A$.  In this
case, a {\mf build\_ref $i$} instruction is generated, where $i$ is
the position of the feature $f$ in $t_2$.  If $f$ is inherited from
$t_1$ only, a self-referential REF cell is created in the result. But
an indication that the actual value for this cell is yet to be
determined must be recorded.  This is done by means of the global
stack S, every element of which is a pair {\mf <action,addr>}, where
{\mf action} is either `copy' or `unify'. In the case we describe, the
action is `copy' and the address is that of the REF cell.  Thus, the
instruction that is generated is {\mf build\_self\_ref}.

If $f$ is appropriate for both $t_1$ and $t_2$, a REF cell with the
address of the $f$-arc in $A$ is created, and a `unify' cell is pushed
onto the stack.  The generated instruction is {\mf
build\_ref\_and\_unify $i$}, where $i$ is the position of $f$ in $t$.
Finally, if $f$ is introduced by $t$, a VAR cell is created, with
$t'=Approp(t,f)$ as its value, by the instruction {\mf build\_var t'}
(VAR cells are explained in section~\ref{sec:var-cells}).

As an example, we list in figure~\ref{comp-code} the resulting code
for the unification the two types {\type a} and {\type b} of the
running example. Since $a\unif b=c$, the first instruction of the
function is {\mf build\_str c}. For every feature that is appropriate
for $c$ an instruction is generated according to the rules described
above. Finally, a {\mf return} instruction completes the function.
\begin{program}{comp-code}{{\mf unify\_type[a,b]}}{
XXXX \= XXXXXXXXXXXXXXXXXXXXXXXX \= \kill
unify\_type[a,b] (b\_addr) \\
\> build\_str(c);               \> \% since $a \unif b = c$ \\
\> build\_self\_ref;            \> \% the value of f1 is yet unknown.   \\
\> build\_ref(1);               \> \% f2 is the first feature of b,     \\
\> build\_ref\_and\_unify(2);   \> \% f3 is the second, and it still    \\
\>                              \> \% has to be unified with a. \\
\> build\_var(bot);             \> \% f4 is a new structure.    \\
\> return;                      \\
}
\end{program}

This example code is rather complex; often the code is much simpler:
for example, when $t_2$ is subsumed by $t_1$, nothing has to be
done. As another example, if $t_1$ is subsumed by $t_2$, then only
additional features of the program term have to be added to $A$.  For
each such feature, a {\mf unify\_feat $i$} instruction is generated,
where $i$ is the position of the feature. Another case is when $t_1$
and $t_2$ are not compatible: {\mf unify\_type[t1,t2]} returns `fail'.
This leads to a call to the function {\mf fail}, which aborts the
unification.\footnote{The notion of failure is elaborated in
section~\ref{sec:failure}; rather than aborting all operations,
failure will indicate the need in backtracking to an alternative
solution.} The effect of the type unification instructions is given in
figure~\ref{type-unif-insts}. The special purpose register ADDR is
used for passing the parameter; the exact details of control transfer
mechanisms, including the effect of {\mf return}, are straight-forward
and won't be specified here.
\begin{figure}[hbt]
\center
\fbox{\mf
\begin{minipage}[t]{6cm}
\begin{tabbing}
\Tabs
\label{inst:unify-type}
build\_str t $\equiv$           \\
                \> HEAP[H] \get <STR,t>;        \\
                \> H \get H + 1;        \\
build\_ref\_and\_unify i $\equiv$               \\
                \> HEAP[H] \get <REF,ADDR+i+1>; \\
        \> push(unify,H);       \\
                \> H \get H + 1;        \\
build\_ref i $\equiv$           \\
                \> HEAP[H] \get <REF,ADDR+i+1>; \\
                \> H \get H + 1;        
\end{tabbing}
\end{minipage} 
\hspace{1cm}
\begin{minipage}[t]{5cm}
\begin{tabbing}
\Tabs
build\_self\_ref i $\equiv$             \\
                \> HEAP[H] \get <REF,H>;        \\
        \> push(copy,H);        \\
                \> H \get H + 1;        \\
build\_var t $\equiv$           \\
                \> HEAP[H] \get <VAR,t>;        \\
                \> H \get H + 1;        \\
unify\_feat i $\equiv$          \\
        \> push(unify,ADDR+i+1);        \\
\end{tabbing}
\end{minipage}
}
\mycaption{The effect of the type unification instructions}{םיסופיט תדחאה תודוקפ}
\label{type-unif-insts}
\end{figure}

\subsection{Processing of a Program}
The program is stored in a special memory area, the {\mf CODE} area.
Unlike the WAM, in our framework registers that are set by the
execution of a query are not helpful when processing a program. The
reason is that there is no one-to-one correspondence between the
sub-terms of the query and the program, as the arities of the TFSs can
differ.  The registers are used, but (with the exception of $X_1$)
their old values are not retained during execution of the program.

Three kinds of machine instructions are generated when processing a
program equation of the form {$X_{i_0}$ =
t($X_{i_1}$,\ldots,$X_{i_n}$)}. The first one is {\mf get\_structure
t/n,$X_{i_0}$}, where $n = Ar(t)$.  For each argument $X_{i_{j}}$ of
$t$ an instruction of the form {\mf unify\_variable $X_{i_{j}}$} is
generated if $X_{i_{j}}$ is first encountered; if it was already
encountered, {\mf unify\_value $X_{i_{j}}$} is generated.  For
example, the machine code that results from compiling the program
\nterm{a(\tag{3}d1,\tag{3})} is depicted in figure~\ref{program-code}.
The implementation of these three instructions is given in
figure~\ref{unif-insts}.

\begin{figure}[hbt]
{\mf
\begin{verbatim}
get_structure a/2,X1    % X1 = a(
unify_variable X2        %        X2,
unify_value X2           %           X2)
get_structure d1/0,X2    % X2 = d1
\end{verbatim}
}
\mycaption{Compiled code for the program \nterm{a(\tag{3}d1,\tag{3})}}{תינכות רובע הנוכמ תודוקפ}
\label{program-code}
\end{figure}

\begin{program}{unif-insts}{Implementation of the get/unify
instructions}{
get\_structure t/n,$X_i$ $\equiv$       \\
\label{inst:get-structure}
        \> addr \get deref($X_i$); $X_i$ \get addr;             \\
        \> case HEAP[addr] of           \\
        \>\> <REF,addr>:        \>\>\>\>\>\>\>\> \% 
                                        uninstantiated cell     \\
        \>\>\> HEAP[H] \get <STR,t>;    \\
        \>\>\> bind(addr,H);    \>\>\>\>\>\>\> \%       
                                        HEAP[addr] \get <REF,H> \\
        \>\>\> for j \get 1 to n do HEAP[H+j] \get <REF,H+j>    \\
        \>\>\> for j \get n downto 1 do push(<copy,H+j>);       \\
        \>\>\> H \get H + n + 1;        \\
        \>\> <STR,t'>:  \>\>\>\>\>\>\>\> \% 
                                        a node  \\
        \>\>\> if (unify\_type[t,t'](addr) = fail) then fail;   \\
\\
unify\_variable $X_i$ $\equiv$          \\
\label{inst:unify-variable}
        \> <action,addr> \get pop();    \\
        \> $X_i$ \get addr;             \\
\\
unify\_value $X_i$ $\equiv$             \\
\label{inst:unify-value}
        \> <action,addr> \get pop();    \\
        \> case action of               \\
        \>\> copy: HEAP[addr] \get HEAP[$X_i$]; \\
        \>\> unify: if (unify(addr,$X_i$) = fail) then fail;            \\
}
\end{program}

The {\mf get\_structure} instruction is generated for a TFS $A_p$ (of
type $t$) which is associated with a register $X_i$. Execution of this
instruction matches $A_p$ against a TFS $A_q$ that resides in memory,
using $X_i$ as a pointer to $A_q$. Since $A_q$ might have undergone
some type inference (for example, due to previous unifications caused
by other instructions), the value of $X_i$ must first be
\label{fun:deref}
dereferenced. This is done by the function {\mf deref} which follows a
chain of REF cells until one that does not point to another, different
REF-cell, is reached. The address of this cell is the value it
returns.

The dereferenced value of $X_i$, {\mf addr}, can either be a
self-referential REF cell or an STR cell. In the first case, the TFS
has to be built by executing the program. A new TFS is being built on
top of the heap (using code similar to that of {\mf put\_structure})
with {\mf addr} set to point to it.  For every feature of this
structure, a `copy' item is pushed onto the stack.  The second case,
in which $X_i$ points to an existing TFS of type $t'$, is the more
interesting one.  An existing TFS has to be unified with a new one
whose type is $t$. Here the pre-compiled {\mf unify\_type[t,t']} is
invoked.

To readers familiar with the WAM, the {\mf unify\_variable}
instruction resembles very much its WAM analog, in the {\em read mode}
of the latter. There is no equivalent of the WAM's {\em write mode} as
there are no real variables in our system. However, in {\mf
unify\_value} there is some similarity to the WAM's modes, where the
`copy' action corresponds to write mode and the `unify' action to read
mode. In this latter case the function {\mf unify} is called, just
like in the WAM.  This function (figure~\ref{unify-code}) is based
upon {\mf unify\_type}.  In contrast to {\mf unify\_type}, the two TFS
arguments of {\mf unify} reside in memory, and full unification is
performed. The first difference is the reason for removing an item
from the stack S and using it as a part of the unification process;
the second is realized by recursive calls to {\mf unify} for subgraphs
of the unified graphs. Notice that the function returns immediately if
its arguments point to the same address, and binds its arguments
otherwise. This guarantees correctness even in the face of cyclic
structures.
\begin{program}{unify-code}{The code of the {\mf unify} function}{
function unify(addr1,addr2:address): boolean;                   \\
\label{fun:unify}
begin                                                           \\
        \> addr1 \get deref(addr1); addr2 \get deref(addr2);    \\
        \> if (addr1 = addr2) then return(true);                \\
        \> if (HEAP[addr1] = <REF,addr1>) then                  \\
        \>\> bind(addr1,addr2); return(true);                   \\
        \> if (HEAP[addr2] = <REF,addr2>) then                  \\
        \>\> bind(addr2,addr1); return(true);                   \\
        \> t1 \get HEAP[addr1].type; t2 \get HEAP[addr2].type;  \\
        \> if (unify\_type[t1,t2](addr2) = fail) then return (false);   \\
        \> for i \get 1 to Ar(t1) do                            \\
        \>\> <action,addr> \get pop();                          \\
        \>\> case action of                                     \\
        \>\>\> copy: HEAP[addr] \get <REF,addr1+i>;             \\
        \>\>\> unify: if (not (unify (addr,addr1+i)))           \\
        \>\>\>\>then return(false);     \\
        \> bind(addr1,addr2);                                   \\
        \> return(true);                                        \\
end;                                                            
}
\end{program}

When a sequence of instructions that were generated for some TFS is
successfully executed on some query, the result of the unification of
both structures is built on the heap and every register $X_i$ stores
the value of its corresponding node in this graph. The stack S is
empty.

\section{Parsing}
\label{machine-parsing}
The previous section delineated the core engine of \amalia; this
section shows how it is extended with control instructions to
accommodate for {\em parsing}.  This constitutes the major difference
between \amalia\ and abstract machines that were devised for variants
of Prolog: computations performed on \amalia\ amount to parsing with
respect to the input grammar, as opposed to SLD resolution.

The parsing process described in section~\ref{parsing} above is a
generic, abstract one: there is no specification of the order in which
new items are computed during each application of $T_{G,w}$. When
designing the control mechanisms of the machine, several parameters
have to be explicated and their values determined. In what follows we
describe how the machine (and a compiler for it) are designed to allow
for efficient implementation of parsing, that is, computation of the
least fix-point of $T_{G,w}$ for a given grammar $G$ and an input
string of lexical elements $w$ of length $n$. The control modules of
\amalia\ are motivated by the abstract process of chart parsing and
are not, in general, inspired by the specific TFS-based formalism that
we deal with. For example, the machine can also be used for parsing
with respect to ``plain'' context-free grammars.

Notice that cases~\ref{predict},~\ref{fact}~and~\ref{scan} of
$T_{G,w}$ are independent of the argument $I$ and add the same items
in every application of the operator. Therefore, when computing the
least fix-point of the operator, they are computed only {\em once},
when the process is initiated. Cases~\ref{dm}~and~\ref{complete} are
more interesting. We treat completion as a special case of dot
movement where the dot is moved from the penultimate position to the
final one, so that completion can take place immediately after the
final application of dot movement. Dot movement creates an item on
the basis of two items in $I$, an active one $[l,A,m,\act]$ and a
complete one $[m,B,r,\comp]$, where $l \le m \le r$.\footnote{We
use $l,m,r$ for left,mid and right, respectively.}  Since for every
off-line parsable grammar the number of items that span a particular
substring of the input is finite, it is possible first to generate the
items spanning $(m,r)$, for all $m > l$, and all the items spanning
$(l,m)$, where $m<r$, and only then the items spanning $(l,r)$. This
is the invariant underlying our design. When generating items that
span $(l,r)$, the active items that span $(l,m)$ are combined with
complete items that span $(m,r)$, where $m$ decreases from $r-1$ to
$l$. We use a {\em chart} to store generated items, since they may be
used more than once.

\subsection{A Parsing Algorithm}
\label{parsing-alg} 
A chart of size $n$ is a data structure that can be accessed by a key
that is a triple $(l,m,r)$, where $1 \le r \le n$, $0 \le l \le r-1$
and $l \le m \le r-1$. Given these restrictions, a chart of
size $n$ can be accessed by $\sum_{r=1}^{n} \sum_{l=0}^{r-1}
\sum_{m=l}^{r-1} 1 = \sum_{r=1}^{n} \sum_{l=0}^{r-1} (r-l) =
\sum_{r=1}^{n} {r(r-1)}/2 = {(n^3-n)}/6$ different keys. Keys are linearly
ordered as follows: $(l,m,r) \prec (l',m',r')$ iff $r < r'$ or $((r =
r')$ and $(l > l'))$ or $((r = r')$ and $(l = l')$ and $(m < m'))$.
Each element of the chart is a pair of {\em chart entries}. Such pairs
are accessed by the coordinates of the key: the element indexed by
$(l,m,r)$ is a pair indexed by $\langle (l,m),(m,r) \rangle$.
Additionally, if two elements have matching sub-keys then it is
required that the corresponding elements of the pairs be identical:
the element indexed by $(l,m)$ in $(l,m,r)$ must be identical to the
element indexed by $(l,m)$ in $(l,m,r')$.  Therefore, even though
there exist ${(n^3-n)}/6$ different keys by which $2
\times{(n^3-n)}/6$ chart entries can be accessed, there are only
${n(n+1)}/2$ different chart entries.

Each chart entry contains two {\em lists of edges}: active and
complete. Each list, both active and complete ones, is a sequence of
{\em edges} along with a specified {\em current} edge,
on which the
following operations are defined:
\begin{description}
\item[{\mf new(list)}:]
return an empty list;
\item[{\mf add\_edge(list,edge)}:]
add {\mf edge} to the end of {\mf list};
\item[{\mf init(list)}:]
set the {\em current} edge of {\mf list} to be the first edge, if
there is one;
\item[{\mf advance(list)}:]
set the {\em current} edge in {\mf list} to be the next edge, if there
is one.
\item[{\mf current(list)}:]
return the {\em current} edge of {\mf list};
\item[{\mf exhausted(list)}:]
return {\mf true} iff the {\em current} element in {\mf list} is undefined;
\end{description}

An {\em edge\/} can be either {\em active\/} or {\em complete}. An
active edge stems from some grammar rule; it contains a part that was
already scanned, and a part that is left to be seen. The position
between the two parts is indicated by the location of the {\em dot}. A
complete edge is a result of scanning an entire body of some rule, and
constructing the rule's head.

\begin{figure}[hbt]
\begin{center}{a. Building chart entries}\end{center}
\begin{minipage}{9cm}
{\mf
\begin{tabbing} \Tabs
Procedure main  \\
for right \get 0 to n do        \\
\>      for left-1 \get right downto 0 do       \\
\>      \>      build\_chart\_entry (left,right)        \\
\end{tabbing}
}
\end{minipage}
\hfill
\begin{minipage}{4.5cm}
\psfig{figure=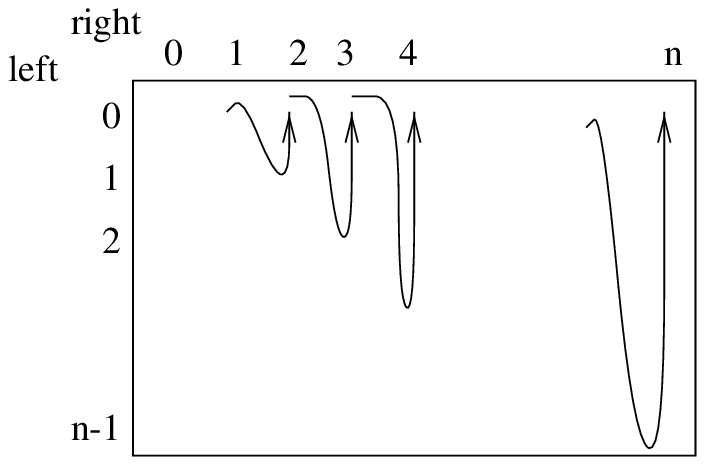,width=4cm}
\end{minipage}
\begin{center}{b. Constructing one chart entry}\end{center}
\begin{minipage}{9cm}
{\mf
\begin{tabbing} \Tabs
Procedure build\_chart\_entry (left,right)      \\
for mid \get right-1 downto left do     \\
\>      combine(left,mid,right)\\
\end{tabbing}
}
\end{minipage}
\hfill
\begin{minipage}{4.5cm}
\psfig{figure=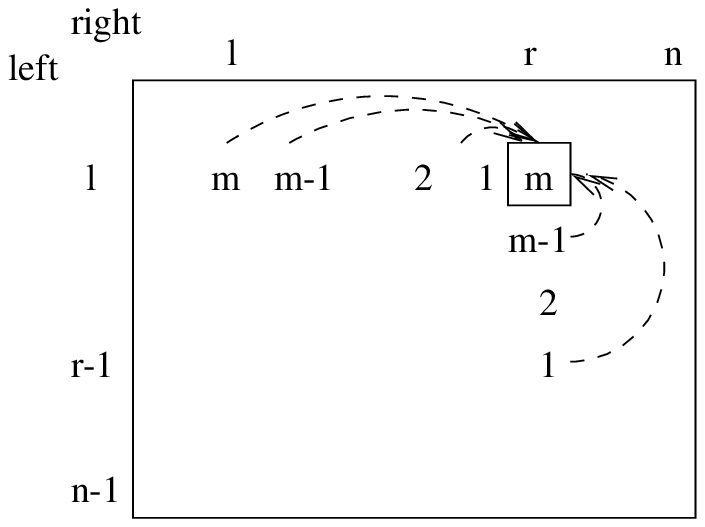,width=4cm}
\end{minipage}
\begin{center}{c. Combining two entries}\end{center}
\begin{minipage}{9cm}
{\mf
\begin{tabbing} \Tabs
Procedure combine(left,mid,right)       \\
for every active edge e1 in chart[left,mid]     \\
\>      for every complete edge e2 in chart[mid,right] \\
\>      \>      e \get dot\_movement(e1,e2)     \\
\>      \>      chart[left,right] \get chart[left,right] $\cup$ \{e\}\\
\end{tabbing}
}
\end{minipage}
\mycaption{Parsing -- informal description}{ילמרופ אל רואית - חותינ}
\label{parse-alg}
\end{figure}

The parsing process is outlined informally in figure~\ref{parse-alg}:
(a) shows the order in which chart entries are
constructed;
(b) shows the order in which chart entries are
scanned to construct the {\mf [left,right]} entry;
combination of chart entries is performed as described in
(c).
The heart of the process is {\em dot movement}, which creates a new
edge $e$ by unifying the TFS that immediately succeeds the dot in an
active edge $e1$ with some complete edge $e2$.

The last part of the process requires some precaution: application of dot
movement to an active edge in $(l,m)$ and a complete one in
$(m,r)$ results in a new edge in $(l,r)$. This edge might
be complete; in the special case where $l=m$, the complete edge
that is thus created is added to $(l,r) = (m,r)$. Such
edges can now be combined with active edges in $(l,m)$
again. Notice that the situation occurs only when $l=m$. The only
way an active edge in the $(l,l)$ entry of the chart (that is,
an edge with the dot in the initial position) can become complete is
if the rule on which the edge is based is of length $2$ (a unit
rule). Therefore, for unit rules a special treatment is required:
first, active edges that originate from unit rules are placed {\em
before} edges that stem from other rules within the same chart entry.
This guarantees that complete edges that result from application of
dot movement on (an active edge that stems from) a unit rule are
constructed {\em before} they are needed. The only problem left is the
possibility of more than one unit edge in the same chart entry. Such
rules are required to be ordered by the grammar writer, such that if
$\rho_1$ can ``feed'' $\rho_2$, $\rho_1$ precedes $\rho_2$ in the
grammar.\footnote{Such an ordering must always exist for off-line
parsable grammars.}

The advantage of this parsing algorithm is that it is simple; in
particular, it does not require an {\em agenda}. In the above
description we assumed that the chart is initialized with complete
edges for the lexical entries of the input words, and with active
edges -- with the dot in the initial position -- for the rules.

Following this informal description, several observations about the
parsing process can be made:
\begin{itemize}
\item
After the edges in the $[l,r]$ chart entry are constructed, no more
edges will be added to entries that were constructed earlier;
\item
When complete edges of column $c$ are used, no complete edges of
columns $1,\ldots,c-1$ will be used any more. Consequently, at every
point during the process, the complete edges in chart entries whose
second index is less than $r$ will not be used any more;
\item
active edges may be used over and over again.
\end{itemize}
These observations guide the specification of the machine
architecture, which is divided into data structures and machine
instructions.

\subsection{Data Structures}
The {\em chart} is represented as a two-dimensional array, indexed by
two integers, where {\mf chart[l,r]} refers to a {\em chart
entry}. Each chart entry contains two {\em lists of edges}, which are
referred to as {\mf chart[l,r].active} and {\mf chart[l,r].complete}.

A {\em complete edge} is represented in memory as a structure {\mf e}
containing an address, {\mf e.addr}, of some {\mf HEAP} cell that is
the root of a feature structure.  An {\em active edge} is represented
as a structure containing a pointer, {\mf e.label}, and a set of {\em
register values}, {\mf e.regs}. An active edge always stems from some
grammar rule; it records a major part of the state of the machine at
some given time.  {\mf e.label} is the address (in {\mf CODE}) of the
first instruction in the code that is generated for the TFS
immediately following the {\em dot} of the edge; {\mf e.regs} records
the values of the machine registers.  The registers represent the part
of the edge prior to the dot, whereas the pointer represents the part
following it.\footnote{The machine's {\em stack\/} is always empty
after the execution of a code that was generated for a feature
structure. Therefore, the stack does not have to be included in the
state.} During a computation of \amalia, edges are repeatedly stored
in the chart and loaded from it. When an edge {\mf e} is loaded, {\mf
e.label} is used to determine the next instruction that is to be
executed; this implies an implicit branch whenever an active edge is
loaded (see the effect of the {\mf call} instruction below). The
auxiliary function {\mf make\_edge} creates an edge from its
components (an address and a set of values for the registers).

Special purpose registers record the current values of the chart
indices {\mf LEFT, RIGHT} and {\mf MID} and the input length {\mf LEN}.
Like all the machine's data structures, the control structures are
initialized before every execution of a program: the registers {\mf
LEFT, MID, RIGHT} and {\mf LEN} are set to $0$ and {\mf new} is
applied to every chart entry. The control structures are affected by
the execution of machine instructions that are generated for both the
program and the query, as explained below.

\subsection{Compilation}
Compilation of a grammar produces code that, when executed, realizes
the parsing process with respect to the grammar and an input query
(that is a sequence of TFSs) representing the Natural Language
input. In this section we describe the compilation scheme in terms of
the resulting code.

First, the chart is initialized. Three sets of edges have to be
inserted to the chart, in correspondence to the three cases of
$T_{G,w}$ (definition~\ref{def:tgw}) that are independent of the
input: complete edges for lexical items, active edges (with the dot in
the initial position) for rules and complete edges for
$\epsilon$-rules.

The first set, corresponding to case~\ref{scan} of $T_{G,w}$, is
inserted to the chart by processing the query. Recall that processing
a query results in the generation of machine code that is executed
prior to the execution of the program code. When the query is composed
of more than one feature structure, the generated code contains {\mf
proceed} instructions that are inserted after every feature structure
in the query. The effect of {\mf proceed} is given in
figure~\ref{proceed}; the parameter $X_i$ is the register that points
to the root of the feature structure that was just created (that is,
the first register mentioned in the code that immediately precedes the
{\mf proceed} instruction).
\begin{program}{proceed}{The effect of the {\mf proceed} instruction}{
proceed $X_i$ $\equiv$  \\
\>      LEN \get LEN + 1;       \\
\>      add\_edge(make\_edge($X_i$,null),chart[LEN-1,LEN].complete);
}
\end{program}
Processing of the query results in edges that are added to the
$[i-1,i]$ diagonal of the chart; in addition, the value of {\mf LEN}
is set to the length of the input. 

The second set, corresponding to case~\ref{predict}, is added by means
of specific machine instructions, {\mf put\_rule l}, that are
generated for each of the rules in the input grammar, where $l$ is the
label of the first instruction of the compiled rule.  {\mf put\_rule}
adds an active edge for the rule starting at address $L$, with no
registers bindings, to the $[i,i]$ entries of the chart
(figure~\ref{put-rule}).  Those instructions are placed by the
compiler {\em prior} to any other instruction of the program.
\begin{program}{put-rule}{The effect of {\mf put\_rule}}{
put\_rule $L$ $\equiv$  \\
\>      for $i \get 0$ to LEN do        \\
\>      \>      add\_edge(make\_edge($L$,null), chart[i,i].active);
}
\end{program}

Rules of length one ($\epsilon$-rules, or {\em empty categories}),
corresponding to case~\ref{fact}, are processed by the compiler in a
special way, that is described in section~\ref{sec:empty-cats}.

The main product of the compiler is the code that corresponds to dot
movement and completion (cases~\ref{dm}~and~\ref{complete}).  For a
rule of the form: $A_1, A_2, \ldots, A_n \Rightarrow A_0$ the compiler
generates the code given in figure~\ref{rule-code}, where $r_i$ is the
index of the first register that is mentioned in the code for $A_i$,
and where $X$ is the first register mentioned in the code for
$A_0$. The effect of this code is discussed in section~\ref{effect}.
\begin{program}{rule-code}{Compiled rule}{
XXXX \= XXXXXXXXXXXXXXXXX \=  XXXXXX \= \kill
     \> put\_rule $L_1$ \>
; add active edges to the main diagonal\\
$L_1$:  \> load\_fs $r_1$\\
     \> \hspace{1cm}  [ (program) code for $A_1$ ] \\
     \> copy\_active\_edge      \> $L_2$\\
\>      $\vdots$        \\
$L_i$:  \> load\_fs $r_i$\>
; $X_{r_i}$ \get ADDR\\
     \> \hspace{1cm}  [ (program) code for $A_i$ ] \\
     \> copy\_active\_edge \> $L_{i+1}$ ; add edge to chart and return    \\
\>      $\vdots$        \\
$L_n$:  \> load\_fs $r_n$\\
     \> \hspace{1cm}  [ (program) code for $A_n$ ] \\
     \> \hspace{1cm}  [ (query) code for $A_0$ ] \\
     \> copy\_complete\_edge    \> $X$ \\
}
\end{program}

Controlling the order in which chart entries are constructed 
is independent of the grammar. Consequently, the compiler generates a
few pieces of identical code for {\em every} grammar.  On the basis of
the generated code for one rule, the code for a grammar consisting of
$k$ rules is given in figure~\ref{grammar-code}. 
\begin{program}{grammar-code}{Compiled code for a grammar with $k$
rules}{
XXXXXX \= XXXXXX \= XXXXXXX \= \kill
        \>      \> put\_rule $L_1$\\
        \>      \>$\vdots$      \\
        \>      \> put\_rule $L_k$\\
        \> first\_key           \\
$L_{start}$:    \> next\_key            \\
$L_{act}$:      \> tst\_active\_edges $L_{end}$         \\
$L_{comp}$:     \> tst\_complete\_edges $L'_{comp}$             \\
        \> load\_machine\_state \\
        \> call \\
        \> next\_complete\_edge $L_{comp}$      \\
$L'_{comp}$:    \> next\_active\_edges $L_{act}$                \\
$L_{end}$:      \> check\_key $L_{start}$       \\
        \> end\_program         \\
$L_1$:  \>      \> code for 1st rule    \\
        \>      \>$\vdots$      \\
$L_k$:  \>      \> code for $k$-th rule \\
}
\end{program}

An important observation regarding the control flow of a compiled
program has to be made here. In general it is impossible to determine, during
compile time, the order in which machine instructions will be
executed. Indeed, some control instructions (in particular, the
key manipulation instructions) resemble ordinary conditional branches.
However, other instructions ({\mf call, copy\_active\_edge} and {\mf
copy\_complete\_edge}) `hide' implicit branches to addresses that
are only known at run time. Further details are given below.

To conclude this section, figure~\ref{fig:comp-prog} depicts (part of)
the code that was generated by the compiler for the example grammar of
figure~\ref{fig:ale-grammar}. Lines 2-11 contain the constant, grammar
independent code; lines 12-49 list the code that was generated for the
first rule; and the code on the right side was generated for the
lexical entries of the words ``john'' and ``love''.
\begin{figure}[hbt]
\center
{\scriptsize
\begin{minipage}[t]{6cm}
\begin{verbatim}
 0        put_rule L6 (1)
 1        put_rule L8 (2)
 2        first_key
 3 L1   : next_key
 4 L2   : tst_active_edges L5
 5 L3   : tst_complete_edges L4
 6        call
 7        next_complete_edge L3
 8 L4   : next_active_edge L2
 9 L5   : check_key L1
10        end_of_program
11 L6   : load_fs X1
12              get_structure sign, X1
13              unify_variable X2
14              unify_variable X3
15              unify_variable X4
16              get_structure n, X2
17              unify_variable X5
18              get_structure nom, X5
19              get_structure agr, X3
20              unify_variable X6
21              unify_variable X7
22              get_structure per, X6
23              get_structure num, X7
24              get_structure sem, X4
25              unify_variable X8
26              unify_variable X9
27              unify_variable X10
28              get_structure atom, X8
29              get_structure atom, X9
30              get_structure atom, X10
31        copy_active_edge L7
32 L7   : load_fs X11
33              get_structure sign, X11
34              unify_variable X12
35              unify_value X3
36              unify_variable X13
37              get_structure v, X12
38              get_structure sem, X13
39              unify_variable X14
40              unify_value X8
41              unify_variable X15
42              get_structure atom, X14
43              get_structure atom, X15
44              put_node phrase, X16
45              put_node s, X17
46              put_arc X16,1,X17
47              put_arc X16,2,X3
48              put_arc X16,3,X13
49        copy_complete_edge 16
\end{verbatim}
\end{minipage} \hspace{1cm}
\begin{minipage}[t]{5cm}
\begin{verbatim}
 0 L91  :       put_node word, X1
 1              put_node n, X2
 2              put_node case, X5
 3              put_node agr, X3
 4              put_node third, X6
 5              put_node sg, X7
 6              put_node sem, X4
 7              put_node john, X8
 8              put_node atom, X9
 9              put_node atom, X10
10              put_arc X1,1,X2
11              put_arc X1,2,X3
12              put_arc X1,3,X4
13              put_arc X2,1,X5
14              put_arc X3,1,X6
15              put_arc X3,2,X7
16              put_arc X4,1,X8
17              put_arc X4,2,X9
18              put_arc X4,3,X10
19        proceed X1

40 L93  :       put_node word, X1
41              put_node v, X2
42              put_node agr, X3
43              put_node third, X5
44              put_node sg, X6
45              put_node sem, X4
46              put_node love, X7
47              put_node atom, X8
48              put_node atom, X9
49              put_arc X1,1,X2
50              put_arc X1,2,X3
51              put_arc X1,3,X4
52              put_arc X3,1,X5
53              put_arc X3,2,X6
54              put_arc X4,1,X7
55              put_arc X4,2,X8
56              put_arc X4,3,X9
57        proceed X1
\end{verbatim}
\end{minipage}
}
\mycaption{Compiled code obtained for the example grammar}{המגודה קודקד רובע הנוכמ תודוקפ}
\label{fig:comp-prog}
\end{figure}

\subsection{Effect of the Machine Instructions}
\label{effect}
This section details the effect of the control module machine
instructions. While the effect of each instruction is given
independently of its context, we assume throughout the description
that a sequence of instructions is present that were generated by the
compiler for some grammar.

The instructions {\mf first\_key}, {\mf next\_key} and {\mf
check\_key}, constantly generated for every grammar,
are aimed at implementing the outermost control flow during parsing.
Motivated by the invariant stated in section~\ref{parsing-alg} above,
these instructions form a loop that causes \amalia\ to build all
the necessary chart entries in the order specified there. The body of
the loop contains code whose effect corresponds to the procedure {\mf
Combine} of figure~\ref{parse-alg}.  The
effect of the key-manipulation instructions is depicted in
figure~\ref{key-insts}.
\begin{program}{key-insts}{The effect of the key-manipulation
instructions}{
first\_key $\equiv$     \\
\>      RIGHT \get 0;   \\
\>      LEFT \get -1;   \\
\>      MID \get -1;    \\
\\
next\_key $\equiv$      \\
\>      MID \get MID-1; \\
\>      if (MID < LEFT) then    \\
\>      \>      LEFT \get LEFT - 1;     \\
\>      \>      if (LEFT < 0) then      \\
\>      \>      \>      RIGHT \get RIGHT + 1;   \\
\>      \>      \>      LEFT \get RIGHT - 1;    \\
\>      \>      MID \get RIGHT - 1;     \\
\>      init(chart[LEFT,MID].active);   \\
\>      init(chart[MID,RIGHT].complete);        \\
\\
check\_key $l$ $\equiv$ \\
\>      if (RIGHT $\neq$ LEN or LEFT $\neq$ 0 or MID $\neq$ LEFT) then  \\
\>      \>      branch $l$;     \\
}
\end{program}

Every iteration of the main loop corresponds to the combination of two
chart entries, taking the active edges from the entry indexed by the
values of {\mf LEFT} and {\mf MID} and the complete edges from the
entry indexed by the values of {\mf MID} and {\mf RIGHT}. To loop over
all the active edges (in the designated chart entry), two instructions
are used: {\mf tst\_active\_edges} and {\mf next\_active\_edge}. 
The instructions {\mf tst\_complete\_edges} and {\mf
next\_complete\_edge} scan the complete edges in the chart entry that
is indexed by the values of {\mf MID} and {\mf RIGHT}.

The effect of these four instructions is given in
figure~\ref{fig:edge-insts}.  
The two instructions that loop over the
active edges are straight-forward; the other two are quite similar, with a
few differences. First, {\mf tst\_complete\_edges} initializes the
list of complete edges in the {\mf [MID,RIGHT]} chart entry. Second,
{\mf next\_complete\_edge} calls the auxiliary function {\mf
reset\_trail}, whose purpose will be discussed presently.

\begin{program}{fig:edge-insts}{The effect of the edge traversal
instructions}{
tst\_active\_edges $l$ $\equiv$ \\
\>      if exhausted(chart[LEFT,MID].active) then       \\
\>      \>      branch $l$;     \\
\\
next\_active\_edge $l$ $\equiv$ \\
\>      advance(chart[LEFT,MID].active) \\
\>      branch $l$      \\
\\
$l$: tst\_complete\_edges $l'$ $\equiv$ \\
\>      if exhausted(chart[MID,RIGHT].complete) then    \\
\>      \>      init(chart[MID,RIGHT].complete);        \\
\>      \>      branch $l'$;    \\
\\
next\_complete\_edge $l$ $\equiv$       \\
\>      reset\_trail;   \\
\>      advance(chart[MID,RIGHT].complete); \\
\>      branch $l$;     
}
\end{program}

{\mf call} (figure~\ref{fig:call-inst}) sets the stage for the
operation of dot movement.  Let {\mf e1} be the current active edge in
the {\mf [LEFT,MID]} entry of the chart, end {\mf e2} -- the current
complete edge in the {\mf [MID,RIGHT]} entry.  The address in the code
area of the next instruction to be executed is stored in {\mf
e1.label}; this code is to be executed on the feature structure
pointed to by {\mf e2.addr}. In other words, the code that was
generated for the next element of {\mf e1}, viewed as a procedure
call, is to be executed on the complete edge {\mf e2}, whose address
is viewed as a parameter.  {\mf call} loads the registers' values from
{\mf e1}, and saves {\mf e2.label} in the special purpose register
{\mf ADDR}, which is used for passing parameters. Then, it saves the
address to return to (that is, the address of the instruction
following the {\mf call}) in a special purpose stack of return
addresses and branches to {\mf e1.label}

\begin{program}{fig:call-inst}{The effect of the {\mf call}
instruction}{
$l$: call $\equiv$      \\
\>      registers \get current(chart[LEFT,MID].active).regs;    \\
\>      ADDR \get current(chart[MID,RIGHT].complete).addr;      \\
\>      push\_return\_addr($l+1$);      \\
\>      branch  current(chart[LEFT,MID].active).label;\\
}
\end{program}

The first instruction that is executed in a `procedure' is {\mf
load\_fs $X$}, which loads the value stored in {\mf ADDR} onto the
register $X$.  Then, the instructions that were generated for this
part of the rule are executed in order, thus unifying the TFS
immediately after the dot (in the active edge) with the TFS that $X_r$
points to. If the unification succeeds, control flows to the {\mf
copy\_active\_edge} instruction that adds the newly created MRS to the
chart, and returns the control to the address stored in the stack of
return addresses. If the entire body of the rule was consumed, the last
instruction is {\mf copy\_complete\_edge} (see
figure~\ref{rule-code}), which adds the newly created {\em complete\/}
edge to the chart and returns. The auxiliary function {\mf copy\_mrs}
copies the MRS accessible from the current registers on top of the
heap. {\mf copy\_fs $X$} copies the feature structure rooted in $X$ on
top of the heap. The effect of these three instructions is depicted in
figure~\ref{fig:rule-insts}.
\begin{program}{fig:rule-insts}{The effect of the {\mf copy}
instructions}{
load\_fs $r$ $\equiv$   \\
\>      $X_r$ \get ADDR;        \\
\\
copy\_active\_edge $l$ $\equiv$ \\
\>      copy\_mrs;      \\
\>      add\_edge(make\_edge($l$,registers),chart[LEFT,RIGHT].active);  \\
\>      branch pop\_return\_address();  \\
\\
copy\_complete\_edge $X$ $\equiv$       \\
\>      copy\_fs($X$);  \\
\>      add\_edge(make\_edge(X,null),chart[LEFT,RIGHT].complete);       \\
\>      branch pop\_return\_address();  \\
}
\end{program}
 
When the code that is associated with some program feature structure
is executed, the heap is modified. Sometimes the same code has to be
executed on several TFSs (since one active edge might be combined with
several complete ones). If the unification fails, that is, {\mf fail}
is called, the heap must be restored to its original form. To this end
a new data structure is introduced: the {\em trail}. It is an array
whose contents are pairs of the form {\mf <address,value>}, which
record modifications to {\mf HEAP} cells. Pairs are being added to the
\label{fun:bind}
trail by means of the {\mf bind} function, whenever the value of a
heap cell is modified.  If all the unifications are successful, and
control flows naturally to {\mf next\_complete\_edge}, the trail is
reset using the auxiliary function {\mf reset\_trail}.

\label{sec:failure}
Consider now the case where some unification fails. The effect of the
{\mf fail} function has to be modified: failure of a ``local''
unification no longer means termination of the program; rather, it
indicates the need to try different edges to combine. Failure can be
detected during the execution of any of the instructions in the {\em
program} code. In this case, the previous bindings are undone, using a
call to {\mf unwind\_trail}, and the stack is initialized using {\mf
reset\_stack}. Then, a branch is made to the last {\mf
tst\_complete\_edges} instruction executed. This instruction's address
is stored in the special purpose register {\mf RETURN\_ADDR}. The
definition of {\mf fail} is given in figure~\ref{fail}.
\begin{program}{fail}{The {\mf fail} function}{
procedure fail; \\
\>      unwind\_trail;          \\
\>      reset\_stack;           \\
\>      branch pop\_return\_address();
}
\end{program}

The WAM uses a trail to undo `side effects' on the stack and the heap
upon backtracking to a choice point (see~\cite[chapter 4.2]{wam}). In
\amalia\ no backtracking is performed and so the trail could have
been eliminated. Notice that after execution of program code, the newly
created edge (whether active or complete) is copied onto the heap. A
different strategy could have been chosen, in which the active edge is
copied {\em prior} to the execution of the program code. In this case,
all that has to be done upon failure is restoring the value of the
heap pointer {\mf H}, so that the cells that were used by
(ineffective) instructions can be re-used. While the gain in this
strategy is that no trail is needed, it doesn't seem to be too
effective: active edges would have to be copied before they are used,
which means that many MRSs will be copied in vain. Since copying
is one of the most time-consuming operations, we opt for the method
described above.

\section{Optimizations and Extensions}
\label{sec:optimizations}
\subsection{Lazy Evaluation of Feature Structures}
\label{sec:var-cells}
One of the drawbacks of maintaining total structures is that when two
TFSs are unified, the values of features that are introduced by the
unified type have to be built.  For example, {\mf unify\_type[a,b]}
(figure~\ref{comp-code}) has to build a TFS of type {\type bot}, which
is the value of the $f4$ feature of type {\type c}. This is expensive
in terms of both space and time; the newly built structure might not
be used at all. Therefore, it makes sense to defer it.

To optimize the design in this aspect, a new kind of heap cells,
VAR-cells, is introduced. A VAR cell whose contents is a type {\type
t} stands for the most general TFS of type {\type t}. VAR cells are
generated by the various {\mf unify\_type} functions for introduced
features; they are expanded only when the explicit values of such
features are needed: either during the execution of {\mf
get\_structure}, where the dereferenced value is a VAR cell, or during
{\mf unify}.\footnote{The effect of {\mf get\_structure} and the
definition of {\mf unify} are modified in a straight-forward way to
accommodate VAR cells.} In both cases the TFS has to be built, by means of
executing the pre-compiled function {\mf build\_most\_general\_fs}
with the contents of the VAR cell as an argument. This function (which
is automatically generated by the type hierarchy compiler) builds a
TFS of the designated type on the heap, with VAR cells instead of REF
cells for the features. These cells will, again, only be expanded when
needed. We thus obtain a lazy evaluation of TFSs that weakly resembles
G\"otz's notion of {\em unfilled feature structures}
(\cite{thilo:master}). Moreover, we gain another important property,
namely that our type hierarchies can now contain loops, since
appropriateness loops can only cause non termination when introduced
features are fully constructed. This approach might not be
applicable in the presence of type constraints, which are currently
not supported by \amalia.

\subsection{Partial Descriptions}
\label{partial-desc}
\amalia\ requires that its input be total: both grammar rules
and lexical entries are required to consist of totally well-typed
feature structures. This requirement might be problematic for the
grammar writer, who might prefer to specify only partial information.
To this end, the compiler employs a pre-processor that performs type
inference on the partial input; the result of this processing is
almost total, but partiality is maintained in certain cases.

Recall that a normal term consists of a tag, a type and a sequence of
arguments, each of which is a normal term. Whenever some sub-term is
the most general term of its type, it is substituted by the type
name only. Using the running example of figure~\ref{hier}, the term
$a(bot,d())$ can be replaced by the term $a$. 

When the compiler encounters such a partial description, it creates
one of the following two instructions: {\mf put\_var t/n, $X_i$}, if
the type $t$ is part of a query code, or {\mf get\_var t/n, $X_i$}, if
it is part of a program code. {\mf put\_var} is very similar to {\mf
put\_node}, with two differences: it creates a VAR-cell, rather than
an STR-cell, on the heap; and it does not leave space for REF-cells,
as there won't be any. {\mf get\_var} is the analog of {\mf
get\_structure}, but is much simpler: it uses {\mf
build\_most\_general\_fs} to create the most general feature structure
of type $t$ on top of the heap, and then calls {\mf unify} to unify this
newly created TFS with the one that is pointed to by $X_i$. Thus,
partial descriptions in the input result in a more efficient code, and
consequently in a faster, more space-economic processing.

\subsection{Empty Categories}
\label{sec:empty-cats}
The presence of empty categories ($\epsilon$-rules) in a grammar
causes both theoretical and practical problems. There is a current
trend in HPSG of avoiding empty categories altogether, due to
theoretical linguistic and cognitive reasons (see,
e.g.,~\cite{sag-fodor94}). From a computational point of view, such
categories always cause considerable efficiency degradation.

\amalia\ is designed to support empty categories as an inherent part of
the input grammars. Empty categories are processed by the compiler at
compile time. Each category is matched against every element $i$ in the
body of every rule $r$, and if the unification succeeds, a new rule is
created: this rule consists of $r$, modified by the effects of
the unification, in which the $i$-th element is removed. This process
can be shown to yield an equivalent grammar, if it terminates.

However, for certain grammars, the process will never terminate, since
it can lead to the creation of new empty categories (when it is
applied to rules with just one element in their bodies). A typical
example would be the rule
\[
\tag{2}\begin{tfs}{list} \end{tfs}
\Longrightarrow 
\begin{tfs}{list}
        hd: & \begin{tfs}{a} \end{tfs}  \\
        tl: & \tag{2}
\end{tfs} 
\]
When it is applied to the empty category $\begin{tfs}{elist}
\end{tfs}$, a new empty category is created: 
\[
\begin{tfs}{list}
        hd: & \begin{tfs}{a} \end{tfs}  \\
        tl: & \begin{tfs}{elist} \end{tfs}
\end{tfs}
\]
This new empty category can, in turn, be unified with the head of the
rule, etc.
To eliminate such infinite loops and to maintain efficiency even in
face of empty categories, the compiler limits their application: new
rules, that were obtained by applying some empty category to an
original grammar rule, can not by applied to other empty categories.
This implies that a single grammar rule cannot derive two empty
categories. Since usually empty categories are designed to operate in
a very limited context, this seems to be a reasonable compromise.

\subsection{Lexical Ambiguity}
The lexicon associates every word $w$ with a set of feature structures
$Cat(w)$. If this set contains more than one element, $w$ is said to
be {\em ambiguous}. \amalia\ processes the lexicon at compile time: to
every input word $w_i$ the lexicon assigns a normal term, which is
transformed to machine instructions. If $w_i$ is ambiguous the lexicon
assigns it several normal terms. The code that is generated for these
terms is regular query code; however, the instruction that separates
the code of one term from the code of another, if both are associated
with the same word, is {\mf same\_word} instead of {\mf proceed}. The
only difference between the two instructions is that the former does
not increment the value of the special purpose register {\mf LEN}. At
run time, {\mf proceed} causes the machine to search for the next
lexical entry, whereas {\mf same\_word} does not. Thus, the execution
of the code that was generated for an ambiguous lexical entry $w$ causes
several complete edges to be inserted into the chart, one for each
element of $Cat(w)$. 

\subsection{Functional Attachments}
While the phrase structure grammar organization underlying our design
is usually appropriate for constructing grammars for natural
languages, there is sometimes need in computations that are not easily
expressed using the formalism. Although contemporary grammatical
formalisms tend to be highly declarative in nature, grammar writers
might find it useful to resort to some mechanism that enables simple
computations to be executed without the full power of the grammatical
formalism. ALE supports this need to the fullest, by incorporating a
complete system of definite clause attachments to grammar rules.
Basically, this is a version of a Prolog-like programming language,
where the basic units are TFSs rather than FOTs.

\amalia\ does not include such a module. As a limited solution, we
implemented a small set of functions that can be used by the grammar
writer; these functions are executed during the parsing process and
their results might be integrated with the parsing.

As an example, consider the pre-defined function {\mf append}. It
receives two parameters, which must be lists of TFSs, and returns a
list consisting of their concatenation. The grammar writer can use
{\mf append} by integrating it in the grammar: following the body of
any rule a goal of the form
`\verb+goal> append(L1,L2,L3)+'
can be placed. The variables \verb+L1+ and \verb+L2+ must be associated with
lists, and after the goal is executed, the variable \verb+L3+ will be bound
to the concatenation of the input lists. Now, \verb+L3+ can be used in
the head of the rule.

Since parsing is performed bottom-up, goals can only be placed {\em
after\/} all the elements of the body of a rule; their input
parameters must be instantiated, and the output parameter can only be
used in the head of the rule. Currently, only a small number of
functions (mainly for handling lists and sets) are integrated into
\amalia, but more can be easily added.  It must be noted, though, that
this situation is very different from ALE, in which the user can
define just {\em any\/} definite clause relation.

\section{Implementation}
\label{sec:implementation}
\amalia\ is implemented as a complete grammar development system,
containing a compiler from the ALE input language to the abstract
machine language, an interpreter for the machine instructions, a
simple debugger for the machine language and a graphical user
interface (GUI) that eases the process of grammar design and
debugging. The major part of the software is written in $C$; the
compiler is written using {\em yacc\/} and {\em lex}, and the
graphical user interface is implemented using {\em
Tcl/Tk\/}~\cite{tcltk}. The system is implemented on a Sun Sparc
station under the Solaris operating system.

The system was tested with a wide variety of grammars, mostly
adaptations of existing ALE grammars. It is important to note that
\amalia\ does not provide the wealth of input specifications ALE does.
Some of ALE's features that are not included in \amalia\ include lexical
rules, free use of definite clause attachments and disjunctive descriptions.
On the other hand, development of grammars in \amalia\ is made easier
due to the GUI and its improved performance over ALE. 
A complete description of \amalia's implementation, its deviations
from ALE's input language and a complete users' guide,
is given in~\namecite{amalia-man}.

To compare \amalia\ with ALE we have used a few benchmark
grammars. Both systems were used to compile the same grammar and to
parse the same strings. We shortly describe below each of the
grammars, and summarize the results of a performance comparison of
\amalia\ and ALE in figure~\ref{fig:performance}.  All times are in
seconds; in ALE we measured the time for the first result to be
displayed, and in \amalia\ -- the time for all the results.

The first grammar is an early version of the HPSG-based Hebrew grammar
described in the next chapter. It consists
of 4 rules and one empty category; the type hierarchy contains 84
types and 32 features, and the lexicon contains 13
words. 
%
%
The second grammar is an  HPSG-based grammar for a subset (emphasizing
relative clauses) of the
Russian language, developed by Evgeniy Gabrilovich and Arkady Estrin. 
It consists
of 8 rules and 76 lexical entries; the type hierarchy contains 151
types and 31 features. 
%
The third example is a simple grammar generating the language $\{a^n
b^n \mid n>0\}$. While the execution times for this simple grammar are
less important, the differences in compilation time indicate a major
advantage in using \amalia\ for instructional purposes; in such cases
grammars are compiled over and over again, while they are usually
executed only a few times. 

\begin{figure}[hbt]
\center
\begin{tabular}{|l||r|r|} \hline
task    & \multicolumn{1}{l|}{ALE} & \multicolumn{1}{l|}{\amalia} \\ \hline
\multicolumn{3}{|c|}{Grammar 1} \\ \hline
Compilation                     & 35.0  & 1.4   \\ \hline
Parsing, 6 words, 2 results     & 0.5   & 0.5   \\ \hline
Parsing, 10 words, 8 results    & 3.2   & 0.8   \\ \hline
Parsing, 14 words, 125 results  & 140.0 & 9.0   \\ \hline
\multicolumn{3}{|c|}{Grammar 2} \\ \hline
Compilation                     & 68.0  & 2.3   \\ \hline
Parsing, 2 words, 2 results     & 0.5   & 0.8   \\ \hline
Parsing, 4 words, 2 results     & 2.4   & 0.9   \\ \hline
Parsing, 7 words, 2 results     & 5.1   & 1.1   \\ \hline
Parsing, 8 words, 2 results     & 7.8   & 1.2   \\ \hline
Parsing, 12 words, 2 results    & 17.0  & 1.5   \\ \hline
\multicolumn{3}{|c|}{Grammar 3} \\ \hline
Compilation     & 6.5   & 0.2   \\ \hline
Parsing, n=4    & 0.1   & 0.2   \\ \hline
Parsing, n=8    & 0.8   & 0.3   \\ \hline
Parsing, n=16   & 2.8   & 1.1   \\ \hline
Parsing, n=32   & 26.0  & 16.0  \\ \hline
\end{tabular}
\mycaption{Performance comparison of \amalia\ and ALE}{םיעוציב תאוושה}
\label{fig:performance}
\end{figure}

\chapter{An HPSG-based Grammar for Hebrew}
\label{sec:hebrew}

\renewcommand{\topfraction}{0.9}
\newcounter{lexample}
\setcounter{lexample}{0} 
\newenvironment{lexample}[3]{\begin{verse}\refstepcounter{lexample}
	\begin{minipage}{1cm}
		(\thelexample)
	\end{minipage}\
        \begin{tabular}{llllllllllllllllllllllllllllll}
        #1\\
        #2\\
	\multicolumn{30}{l}{``#3''}\end{tabular}}{\end{verse}}
\newcommand{\ha}{{\sf ha-}}
\newcommand{\koll}{{\sf koll}}
\newcommand{\sepr}{{\sf sepr}}
\newcommand{\sparim}{{\sf sparim}}
\newcommand{\gadol}{{\sf gadol}}
\newcommand{\banglit}{{\sf b-'anglit}}
\newcommand{\ze}{{\sf ze}}
\newcommand{\sheni}{{\sf \$eni}}
\newcommand{\shlosha}{{\sf \$lo\$a}}
\newcommand{\shlosht}{{\sf \$lo\$t}}
\newcommand{\qaniti}{{\sf qaniti}}
\newcommand{\exad}{{\sf 'exxad}}
\newcommand{\axat}{{\sf 'axxat}}
\newcommand{\axadim}{{\sf 'xadim}}

\newcommand{\nismak}{{`nismak'}}

\newcommand{\elist}{\langle \rangle}
\newcommand{\fsthe}{
 \begin{tfs}{phrase}
  qstore: & \{\}\\
  synsem: & \begin{tfs}{synsem}
  \end{tfs} 
 \end{tfs} 
}       

In order to test the validity of the abstract machine and its
appropriateness for designing HPSG-based grammars, we have devised a
small-scale grammar for a fragment of the Hebrew language, based upon
the principles of HPSG as stipulated in~\namecite{hpsg2}. It must be
emphasized that the main objective of the grammar design was to verify
the machine, and therefore its linguistic contributions are minor.
Still, it might serve as the starting point for the construction of a
larger scale, broad coverage grammar for the language.

The Hebrew script uses a character set that differs from the one that
appears on an ordinary keyboard. The script is highly ambiguous, as
most of the vowels are not written; furthermore, many particles (prepositions,
articles and conjunctions) are attached (in the script) to the words
succeeding them. Since the problem of morphological analysis of
Hebrew, even when represented in the Hebrew script, is practically
solved, we have decided in this work to use a transcription of Hebrew,
known as {\em Phonemic Script}\footnote{This script was accepted as a
standard number ISO-DIS 259-3.}~\cite{ornan86,ornan94,ornan95}.
First, it uses only symbols that appear on any computer keyboard;
second, it is unambiguous, similarily to average European languages.

We first list (section~\ref{sec:schemas}) some of the major HPSG
schemata that serve to combine different kinds of phrases, along with
their adaptation to our needs. Section~\ref{sec:nps}
describes the structure of noun phrases, and we concentrate in
section~\ref{sec:def} on the status of the definite article in Hebrew.
Section~\ref{sec:smikut} briefly discusses noun-noun constructs. The
complete grammar is listed in appendix~\ref{app:grammar}.

\section{Phrase Structure Schemata}
\label{sec:schemas}
HPSG ``rules'' are organized as a set of {\em principles\/} that
set constraints on the properties of well-formed phrases, along with a
set of {\em ID schemata\/} that license certain phrase structures. The
schemata are independent of the categories of the involved phrases;
they state general conditions for the construction of larger phrases
out of smaller ones, according to the function of the sub-phrases.
In~\cite{hpsg2} six schemata are listed; we have adopted four of them
in our grammar. 

ID schemata only {\em license\/} certain phrase combinations. They do
not specify {\em all\/} the constraints imposed on the involved
sub-phrases, as these are articulated by the principles. However, in a
system that is based on phrase-structure rules (e.g., ALE) the
principles and the schemata must be interleaved: each rule encodes not
only the phrase structure, but also constraints imposed by the grammar
principles. 

Consider, for example, the {\em head-subject\/} schema of HPSG, which
states that a phrase with an empty {\em subj\/} list can be
constructed by combining a (head) phrase, whose
{\em subj\/} list is of length~1, with a (subject) phrase. Nothing in
the schema relates the subject to the head; it is the {\em
subcategorization principle\/} that requires that the subject be
unifiable with the single element in the head's {\em subj\/} list.
Furthermore, the {\em head feature principle\/} requires that the
values of the {\em head\/} features in both the phrase itself and its
head sub-phrase be identical. The first rule listed below
(figure~\ref{fig:schema1}) combines these constraints: it states that
a phrase can be constructed out of two sub-phrases, the subject and
the head, where the first element (the value of the {\mf hd} feature)
in the {\em subj\/} list of the head is token-identical to the subject
(through the use of the {\mf Subj} variable), and the {\mf head}
features of the phrase and its head are token-identical (through the
use of the {\mf Head} variable).

\begin{description}
\item[Subject-Head schema]
Most importantly, this schema licenses the combination of a subject
with a predicate to form a sentence. The properties of the subject are
taken from the {\em subj\/} feature of the head daughter. The schema
is listed in figure~\ref{fig:schema1}.
%
\item[Head-Complement schema]
The rest of the complements, other than the subject, are combined with
the head by the head-complement schema. Once again, the appropriate
complements are determined by the head and are specified as the
elements in the list {\em comps}, as shown in figure~\ref{fig:schema2}.
%
\item[Head-Marker Schema]
Markers are used to guarantee that a certain modifier combines only
once with a certain head. A typical example is quantifiers (such as
`every') modifying nouns. This schema is listed in figure~\ref{fig:schema4}.
%
\item[Head-Adjunct schema]
Adjuncts can be combined with the heads they modify over and over
again. In HPSG adjuncts select their heads -- it is the adjunct that
determines the features of the head it might be attached too, through
the value of the feature {\em mod\/}, as depicted in figure~\ref{fig:schema5}.
\end{description}

\begin{figure}[hbt]
\begin{verbatim}
% Schema 1 (ch. 9, p. 347)
% Subject - Head
subject_head rule 
(phrase,cat:(cat,head:Head,subj:e_list,comps:Comps,spr:Spr,marking:Marking),
       cont:Cont,conx:backgr:BM,qstore:QM)
===>
cat> % subject
(Subj,sign,cat:cat,cont:sem_obj,conx:backgr:BS,qstore:QS),	
cat> % head
(sign,cat:(cat,head:Head,subj:(hd:Subj,tl:e_list),
           comps:(Comps,e_list),spr:Spr,marking:Marking),
      cont:(Cont,sem_obj),conx:backgr:BH,qstore:QH),
goal> union(QS,QH,QM),
goal> union(BS,BH,BM).
\end{verbatim}
\mycaption{Subject-Head schema}{ןיערג-אשונ תמכס}
\label{fig:schema1}
\begin{verbatim}
% Schema 2 (ch. 9, p. 348)
% Head - Complement
head_complement rule
(phrase,cat:(cat,head:Head,subj:Subj,comps:Comps,spr:Spr,marking:Marking),
        cont:Cont,conx:backgr:BM,qstore:QM)
===>
cat> % head
(sign,cat:(cat,head:Head,subj:Subj,
           comps:(hd:Comp,tl:Comps),
           spr:Spr,marking:Marking),
      cont:Cont,conx:backgr:BH,qstore:QH),
cat> % complement
(Comp,sign,cat:cat,cont:sem_obj,conx:backgr:BC,qstore:QC),	
goal> union(BH,BC,BM),
goal> union(QH,QC,QM).
\end{verbatim}
\mycaption{Head-Complement schema}{םילשמ-ןיערג תמכס}
\label{fig:schema2}
\end{figure}
\begin{figure}[hbt]
\begin{verbatim}
% Schema 4 (ch. 1, p. 51)
% Marker - Head
marker_head rule
(phrase,cat:(cat,head:Head,subj:Subj,comps:Comps,spr:Spr,marking:Marking),
        cont:Cont,conx:backgr:BM,qstore:(elt:Elt,elts:Elts))
===>
cat> % marker
(word,cat:(cat,head:(mark,spec:HeadDtr),
           subj:list,comps:list,spr:list,marking:(Marking,marked)),
      cont:(Elt,quant,det:sem_det,restind:sem_obj),
      conx:backgr:BD,qstore:e_set),
cat> % head
(HeadDtr,sign,cat:(cat,head:Head,subj:Subj,comps:Comps,
                   spr:Spr,marking:unmarked),
              cont:Cont,conx:backgr:BH,qstore:Elts),
goal> union(BD,BH,BM).
\end{verbatim}	
\mycaption{Head-Marker schema}{ןמס-ןיערג תמכס}
\label{fig:schema4}
\begin{verbatim}
% Schema 5 (ch. 9, p. 403)
% Head - Adjunct
head_adjunct rule
(phrase,cat:Cat,cont:Cont,conx:backgr:BM,qstore:QM)
===>
cat> % head
(HeadDtr,sign,cat:Cat,cont:sem_obj,conx:backgr:BH,qstore:QH),
cat> % adjunct
(sign,cat:head:(adj,defness:defness,mod:HeadDtr),
      cont:Cont,conx:backgr:BA,qstore:QA),
goal> union(BH,BA,BM),
goal> union(QH,QA,QM).
\end{verbatim}
\mycaption{Head-Adjunct schema}{חפסנ-ןיערג תמכס}
\label{fig:schema5}
\end{figure}

\section{The Structure of Noun Phrases in Hebrew}
\label{sec:nps}
A noun phrase (NP) is a phrase that is headed by a
noun\footnote{Elliptic NPs might not contain a noun, but we don't
discuss ellipsis here.} (N), optionally modified or complemented by
various adjuncts. In this section we list the possible adjuncts and
briefly discuss their character. A more thorough discussion of
selected phenomena is provided in the next sections.  More Hebrew data
as well as further references can be found
in~\cite{ornan64,ornan79,glinert,shuly:master,dana}.

A {\em noun\/} is a word whose {\em head\/} feature has the type {\em
noun\/} and whose {\em cont\/} feature is of type {\em nom\_obj}. The
{\em head\/} feature of nouns carries an additional (boolean) feature, {\em
def}, which is explained in section~\ref{sec:heb-def} below. Hebrew
nouns are specified for {\em gender, number\/} and {\em
person}\footnote{Only pronouns are specified for person, other nouns
are inherently third person.}, and
these three features are listed as part of the {\em index\/} feature
of nouns. Figure~\ref{fig:ex-noun} depicts the lexical entry of the
common noun {\sf sepr} (book), where `$\elist$' represents an empty
list and `$\{ \}$' denote a set.
\begin{figure}[hbt]
\[
\begin{tfs}{word}
  phon: & \mbox{\sf sepr}\\
  cat: & \begin{tfs}{cat}
    head: & \begin{tfs}{noun}
      def: & -
    \end{tfs} \\
    subj: & \elist \\
    comps: & \elist \\
    spr: & \elist \\
    marking: & \begin{tfs}{marking} \end{tfs}
  \end{tfs}\\
  cont: & \begin{tfs}{nom\_obj}
    index: & \tag{2}\begin{tfs}{index}
      per: & \begin{tfs}{third} \end{tfs} \\
      num: & \begin{tfs}{sg} \end{tfs} \\
      gen: & \begin{tfs}{masc} \end{tfs} \\
    \end{tfs}\\
    restr: & \left\{ \begin{tfs}{psoa}
      nucleus: & \begin{tfs}{book} instance: & \tag{2} \end{tfs}
    \end{tfs} \right\}
  \end{tfs}\\
  qstore: \{ \}
\end{tfs}
\]
\mycaption{The lexical entry of the noun {\sf sepr}}{םצעה םש לש לויעה}
\label{fig:ex-noun}
\end{figure}

Hebrew is a relatively free constituent order language. Still, the
order of the NP elements is sometimes fixed. In particular, the
adjuncts can be strictly classified as either pre-head or post-head.
Within each category a default ordering exists, from which some
deviations are allowed. In the discussion below the adjucnts are
listed by this default ordering.

\paragraph{Pre-head adjuncts}
\begin{description}
\item[Determiners]
This is a closed class of words such as {\sf koll} (all/every), {\sf
robb} (most-of), {\sf kamma} (some) etc.
\item[Cardinal numbers]
Such as {\shlosha} (three). Cardinals appear in two forms: the regular
one and the \nismak\ form, discussed in section~\ref{sec:cardinals}.
\item[Definite article]
The definite article {\sf ha-} is separated from the other determiners
for reasons that are explicated in section~\ref{sec:def}.
\end{description}

\paragraph{Post-head adjuncts}
\begin{description}
\item[Nominal complement]
Hebrew allows a very elaborate system of nominal-nominal compounds. The
first nominal might be a noun or, rarely, an adjective; it is the syntactic
head of the compound, and it is morphologically marked. The second
nominal can be any NP. Compounds are discussed in
section~\ref{sec:smikut}.
\item[Adjectives]
Hebrew adjectives are marked for number, gender and definiteness, 
on which they must agree with the head noun. 
\item[Ordinal numbers]
Such as {\sf \$eni} (second).
\item[Possessives]
These include possessive pronouns such as {\sf \$elli} (mine) as well as
phrases ({\sf \$ell dan} -- Dan's).
\item[Prepositional phrases]
The rules that govern the combination of prepositional phrases to head
nouns in Hebrew are very similar to those in English.
\item[Subcategorized complements]
Certain nouns subcategorize for particular complements. For example,
verbal nouns such as {\sf racon} (wish) permit an
infinitival verb phrase as a complement. This is encoded in the list
of complements (the value of {\em comp}) in the lexical entries of the nouns.
\item[Relative clauses]
Are not covered in our grammar.
\end{description}

As mentioned above, a thourough and complete description of the
linguistic data is outside the scope of this work. The reader is
referred to, e.g.,~\cite{ornan64,ornan79,glinert,dana} for more details.

\section{The Status of the Definite Article in Hebrew}
\label{sec:def}
\subsection{The Data}
Hebrew marks {\em definiteness\/} in a way that differs a lot from
English (but resembles other Semitic language, notably Arabic, and
also modern Greek, as will be shown below). The definite article {\sf
ha-} in Hebrew attaches to {\em words}, not to phrases. It combines
with various kinds of nominals: common nouns, a few proper nouns,
adjectives, ordinal numbers, cardinal numbers and demonstratives.
Moreover, definite noun phrases in Hebrew are {\em polydefinite}: most
of the elements of the phrase are required to be explicitly
definite,
and there is a strict requirement that these elements {\em agree\/} on
definiteness for the phrase to be grammatical.
Hebrew does have {\em indefinite\/} articles (\exad, \axat,
\axadim), but their use is optional and not common. It is therefore
useful to view bare nominals (with no attached
definite article) by default as indefinite.
See examples~(\ref{ex:def1})~--~(\ref{ex:def8}) for some data.

\begin{minipage}{6.5cm}
\begin{lexample}
	{\ha & \sepr}
	{the & book}
	{the book}
\label{ex:def1}
\end{lexample}
\begin{lexample}
	{\ha & \sepr & \ha & \gadol}
	{the & book & the & big}
	{the big book}
\label{ex:def2}
\end{lexample}
\begin{lexample}
	{\ha & \sepr & \ha & \sheni}
	{the & book & the & second}
	{the second book}
\label{ex:def3}
\end{lexample}
\end{minipage}
\hspace{1cm}
\begin{minipage}{6cm}
\begin{lexample}
	{\sepr & (\exad)}
	{book & (one)}
	{a book}
\label{ex:def5}
\end{lexample}
\begin{lexample}
	{\sepr & \gadol& (\exad)}
	{book & big & (one)}
	{a big book}
\label{ex:def6}
\end{lexample}
\begin{lexample}
	{\sepr & \sheni}
	{book & second}
	{a second book}
\label{ex:def8}
\end{lexample}
\end{minipage}

\subsection{HPSG Approach}
HPSG (as formulated in~\cite{hpsg2}) uses two schemata to form simple
noun phrases (NPs): the head-marker schema combines a determiner (DET)
with a noun (N), and the head-adjunct schema combines any number of
adjectives (ADJs) with an NP. Nouns subcategorize for DET in English:
the lexical entry of a singular noun explicitly states an anticipation
for a determiner. The combination of DET-N results in a full NP; the
effect of the determiner is recorded in the semantics of the phrase as
the value of the QSTORE feature propagates from the determiner to the mother.
Adjectives `select' the NP they modify in the sense
that the NP is the contents of the MOD feature in the adjunct's lexical
entry. The head-adjunct schema treats the NP as the head and the ADJ as
the semantic head, so that the semantics of the phrase is inherited
from the adjunct. 

The HPSG account of~\namecite{hpsg2} would not be
appropriate for Hebrew due to the differences in the structure of NPs
in the two languages. Most notably, Hebrew nouns do not
subcategorize for determiners, for bare nouns qualify perfectly as
complete NPs, as shown in examples~(\ref{ex:def1})~-~(\ref{ex:def8})
above.  

An alternative construction is the HPSG analysis of modern
Greek NPs presented in~\cite{greek}. It appears that in Greek, too,
the definite article can attach to various kinds of nominals, and the
language exhibits both {\em monadic\/} definites and {\em
polydefinites}. Thus, all three phrases
in~(\ref{ex:greek1})~--~(\ref{ex:greek3}) are grammatical:\footnote{
The Greek examples are taken from~\namecite{greek}.}

\begin{lexample}
	{{\sf to} & {\sf kokino} & {\sf podilato}}
	{the & red & bike}
	{the red bike}
\label{ex:greek1}
\end{lexample}
\begin{lexample}
	{{\sf to} & {\sf kenurio} & {\sf to} & {\sf kokino} & {\sf podilato}}
	{the & new & the & red & bike}
	{the new red bike}
\label{ex:greek2}
\end{lexample}
\begin{lexample}
	{{\sf ta} & {\sf dio} & {\sf ta} & {\sf podilata} & {\sf ta} & {\sf kokina}}
	{the & two & the & bikes & the & red}
	{the two red bikes}
\label{ex:greek3}
\end{lexample}

\namecite{greek} concludes that the Greek definite article is not a regular
determiner, but constitutes a category of its own. It does not head the
phrase it occurs in (as was suggested by~\cite{netter94} for Germanic
languages); rather, it functions as an adjunct: it is optional, and it
selects the head it modifies by specifying this head as the value of the MOD
feature. Furthermore, the definite article marks the phrase it occurs
in as definite. This is achieved by introducing a new feature of
nominals, UNIQUE,\footnote{The specification of uniqueness has a
semantic contribution in addition to its syntactic marking, but we
suppress a complete discussion of semantics here.} whose (boolean) value
is `+' iff the nominal is definite. Naturally, the value of this
feature in the lexical entries of nominals is~`$-$' (since they are
indefinite by default).

\subsection{An Analysis of Hebrew Definites}
\label{sec:heb-def}
The analysis of~\namecite{greek} employs a non-quantificational
approach to the semantics of definites. The UNIQUE feature is a
semantic one (it is part of the CONTents of a phrase), and is the only
indication of the definiteness of the phrase. This is in contrast to
the approach of~\cite{hpsg2} that is based on Cooper Storage of
quantifiers. Whatever approach to semantics is taken, it is clear from
examples~(\ref{ex:def1})~-~(\ref{ex:def8}) that agreement in definiteness
among elements of the NP in Hebrew is a morpho-syntactic process, and
we account here for this component of the grammar only.

In contrast to Greek, Hebrew exhibits no cases of monadic definites,
so all we have to account for is the case of polydefinites. A major
observation here is that the definite article attaches only to {\em
words}. Therefore, it seems reasonable to account for definite article
combination by means of a {\em lexical rule\/} that creates a definite
nominal out of an indefinite one. For a detailed discussion of the
definite article in Modern Hebrew, see~\cite{ornan64}.


The {\em Definite Lexical Rule\/} (DLR) operates on various kinds of
nominals: nouns (e.g., \sepr), adjectives (e.g., \gadol), ordinals
(e.g., \sheni), demonstratives (e.g., \ze) and cardinals (e.g.,
\shlosha). In all categories its effect on the phonology is that of
prefixing it with {\sf ha-}.  To emphasize the fact that definiteness
agreement in Hebrew is not a semantic process, we add a boolean feature
DEF to the CATegory of nominals (rather than to their CONTent). The
DLR changes the value of the path SYNSEM$|$LOC$|$CAT$|$DEF
from~`$-$'~to~`+'. When the DLR operates on {\em adjuncts}, it
additionally changes the value of the path MOD$|$LOC$|$CAT$|$DEF in
the same manner. Thus it is guaranteed that definite adjectives, for
example, are not only specified as definite but also select definite
heads. The effect of the DLR when applied to a few nominals is
exemplified in figure~\ref{fig:dlr-effect}.
\begin{figure}[hbt]
{\footnotesize
\[
\begin{array}{ccc}
\begin{tfs}{word}
 phon: & \mbox{\sepr} \\
 cat: & \begin{tfs}{cat}
  head: & \begin{tfs}{noun} \end{tfs} \\
  subcat: & \elist \\
  def: & -
 \end{tfs} \\ 
 cont: & \begin{tfs}{nom\_obj}
  index: & \tag{1}\begin{tfs}{index}
   per: & \begin{tfs}{3rd} \end{tfs} \\
   num: & \begin{tfs}{sg} \end{tfs} \\
   gen: & \begin{tfs}{m} \end{tfs} 
  \end{tfs} \\ 
  restr: & \left\{ \begin{tfs}{book} inst: & \tag{1} \end{tfs} \right\}
 \end{tfs} 
\end{tfs} 
& \stackrel{DLR}{\longrightarrow} &
\begin{tfs}{word}
 phon: & \mbox{\ha\sepr} \\
 cat: & \begin{tfs}{cat}
  head: & \begin{tfs}{noun} \end{tfs} \\
  subcat: & \elist \\
  def: & +
 \end{tfs} \\ 
 cont: & \begin{tfs}{nom\_obj}
  index: & \tag{1}\begin{tfs}{index}
   per: & \begin{tfs}{3rd} \end{tfs} \\
   num: & \begin{tfs}{sg} \end{tfs} \\
   gen: & \begin{tfs}{m} \end{tfs} 
  \end{tfs} \\ 
  restr: & \left\{ \begin{tfs}{book} inst:  & \tag{1} \end{tfs} \right\}
 \end{tfs} 
\end{tfs} 
\\ \\
\begin{tfs}{word}
 phon: & \mbox{\gadol} \\
 cat: & \begin{tfs}{cat}
  head: & \begin{tfs}{adj} 
   mod: & \begin{tfs}{synsem}
   \end{tfs} 
  \end{tfs} \\ 
  subcat: & \elist \\
  def: & \tag{3}-
 \end{tfs} \\ 
 cont: & \begin{tfs}{nom\_obj}
  index: & \tag{1}\begin{tfs}{index}
   per: & \begin{tfs}{3rd} \end{tfs} \\
   num: & \begin{tfs}{sg} \end{tfs} \\
   gen: & \begin{tfs}{m} \end{tfs} 
  \end{tfs} \\ 
  restr: & \left\{ \begin{tfs}{big} inst: &  \tag{1} \end{tfs} \right\}
         \cup \tag{2}
 \end{tfs} 
\end{tfs} 
& \stackrel{DLR}{\longrightarrow} &
\begin{tfs}{word}
 phon: & \mbox{\ha\gadol} \\
 cat: & \begin{tfs}{cat}
  head: & \begin{tfs}{adj} 
   mod: & \begin{tfs}{synsem}
   \end{tfs} 
  \end{tfs} \\ 
  subcat: & \elist \\
  def: & \tag{3}+
 \end{tfs} \\ 
 cont: & \begin{tfs}{nom\_obj}
  index: & \tag{1}\begin{tfs}{index}
   per: & \begin{tfs}{3rd} \end{tfs} \\
   num: & \begin{tfs}{sg} \end{tfs} \\
   gen: & \begin{tfs}{m} \end{tfs} 
  \end{tfs} \\ 
  restr: & \left\{ \begin{tfs}{big} inst: &  \tag{1} \end{tfs} \right\}
         \cup \tag{2}
 \end{tfs} 
\end{tfs} 
\end{array}
\]
}
\mycaption{The effect of the Definite Lexical Rule}{עודייל ילקיסקלה קוחה}
\label{fig:dlr-effect}
\end{figure}

Once the process of adding the definite article is taking place in the
lexicon, the head-adjunct schema can remain intact.
Moreover, the agreement in definiteness between a nominal and its
adjuncts is stated in the lexical entry of the adjuncts, just like
agreement on number and gender is.

\label{sec:cardinals}
Cardinals introduce an irregularity to the analysis of definites. As
mentioned above, cardinals can combine with the definite article in
Hebrew. However, such constructs are used only in elliptic phrases. In
full noun phrases, when the head noun is present, the definiteness
agreement between the head noun and the cardinal number is realized in
a unique way: a definite noun does not combine with a definite
cardinal, but rather with {\em construct\/} form of the cardinal, 
\nismak. The absolute form of many other nominals have \nismak\ forms
that are used in noun-noun constructs (see
section~\ref{sec:smikut}). However, cardinals in this form are
implicitly definite, as they combine only with definite
NPs.\footnote{This rule has a few exceptions: the cardinal {\sf \$nei}
(two-\nismak) is combined with both definite and idefinite nouns; and
there are few indefinite nouns (such as {\sf me'ot} (hundreds) or {\sf
'lapim} (thousands)) that require \nismak\ cardinals. The phrases
preceded by `?' are marked, archaic forms.} The data are
summarized in examples~(\ref{ex:card1})~to~(\ref{ex:card5}) below.
\begin{lexample}
	{\shlosha & \sparim}
	{three & books}
	{three books}
\label{ex:card1}
\end{lexample}
\begin{lexample}
	{?\shlosha & \ha & \sparim}
	{three & the & books}
	{?the three books}
\label{ex:card2}
\end{lexample}
\begin{lexample}
	{\shlosht & \ha & \sparim}
	{three-\nismak & the & books}
	{the three books}
\label{ex:card3}
\end{lexample}
\begin{lexample}
	{?\shlosht & \sparim}
	{three-\nismak & books}
	{?three books}
\label{ex:card4}
\end{lexample}
\begin{lexample}
	{$*$\ha & \shlosha & \ha & \sparim}
	{the & three & the & books}
	{?the three books}
\label{ex:card5}
\end{lexample}
Notice that~(\ref{ex:card5}) is ungrammatical because the correct way of
marking definiteness of cardinal in Hebrew is by using the \nismak\
form, and not because the phrase {\sf \ha \shlosha} is ungrammatical.
Indeed, this last phrase can be used in elliptical structures such
as~(\ref{ex:card6}):
\begin{lexample}
	{\qaniti & \shlosha & \sparim. & \koll & \ha & \shlosha & \banglit}
	{I-bought & three & books. & All & the & three & in-English}
	{I bought three books. All three {\em of them are} in English}
\label{ex:card6}
\end{lexample}

The \nismak\ form of nominals is an inflection of their regular form,
and therefore is obtained as the outcome of a lexical rule. As far as
definiteness is concerned, when the DLR operates on an indefinite
cardinal, its output is the \nismak\ form rather than the regular
combination of {\sf ha-} and the cardinal. All other details remain
the same.  The definite {\sf ha-} is combined with cardinals by a
different mechanism that is not discussed here.

\section{Noun-Noun Constructs}
\label{sec:smikut}
Noun-noun compounds are constructed in a special way in Hebrew: the
head noun, which appears first in the compound, has a marked
morphological form\footnote{For many nouns in Hebrew, especially
among singular masculine and plural feminine, this form is
identical to the regular form.} -- \nismak. Most NPs can serve as the
adjunct of a compound.
Syntactically, the compound inherits all the features of the \nismak,
with the exception of {\em definiteness}, which is inherited from the
NP complement. Consider the following examples:
\begin{lexample}
	{\sf pirxei & \sf gann & \sf yapim}
	{flowers-pl-\nismak & garden-sg & beautiful-pl}
	{beautiful garden flowers}
\label{ex:smikut1}
\end{lexample}
\begin{lexample}
	{\sf pirxei & \sf ha- & \sf gann & \sf ha- & \sf yapim}
	{flowers-pl-\nismak & the & garden-sg & the & beautiful-pl}
	{the beautiful garden flowers}
\label{ex:smikut2}
\end{lexample}
In both examples the entire phrase is in plural, as can
be seen from the adjective, because the head noun {\sf pirxei} is in
plural. However,~(\ref{ex:smikut2}) is definite while~(\ref{ex:smikut1})
isn't, due to the definite article modifying the complement {\sf
gann}.

The process of compounding is recursive, as the resulting compound is
a legitimate NP for combining with some other `nismak' form. When more
than two nouns are combined, the resulting phrase might be (if the
nouns have the same gender and number) syntactically ambiguous:
example~(\ref{ex:ambig}) can be translated as ``my fat aunt's cow'' or
as ``my aunt's fat cow''.
\begin{lexample}
	{\sf parat & \sf dodati & \sf ha- & \sf \$mena}
	{cow-\nismak & my-aunt & the & fat-f}
	{my fat aunt's cow / my aunt's fat cow}
\label{ex:ambig}
\end{lexample}

The \nismak\ form is generated from the regular noun form by means of
the {\em \nismak\ lexical rule} (NLR). Apart from modifying the
phonology of the noun, this rule has a double effect. First, it adds a
subcategorized NP complement to the COMP list of the noun, to express
the expectation for an NP complement; second, it unifies the values of
the DEF feature of the noun and its newly added complement. When the noun
is complemented, the resulting phrase inherits the
definiteness from the adjunct. Figure~\ref{fig:nlr} depicts the
effect of applying the NLR to the noun {\sf praxim} (flowers).
\begin{figure}[hbt]
{
\begin{eqnarray*}
\lefteqn{%
\begin{tfs}{word}
 phon: & \mbox{\sf praxim} \\
 cat: & \begin{tfs}{cat}
  head: & \begin{tfs}{noun} \end{tfs} \\
  subcat: & \elist \\
  def: & -
 \end{tfs} \\ 
 cont: & \begin{tfs}{nom\_obj}
  index: & \tag{1}\begin{tfs}{index}
   per: & \begin{tfs}{3rd} \end{tfs} \\
   num: & \begin{tfs}{pl} \end{tfs} \\
   gen: & \begin{tfs}{m} \end{tfs} 
  \end{tfs} \\ 
  restr: & \left\{ \begin{tfs}{flower} inst: & \tag{1} \end{tfs} \right\}
 \end{tfs} 
\end{tfs} 
\stackrel{NLR}{\longrightarrow}
} \\
& & 
\begin{tfs}{word}
 phon: & \mbox{\sf pirxei} \\
 cat: & \begin{tfs}{cat}
  head: & \begin{tfs}{noun} \end{tfs} \\
  subcat: & \left\langle 
   \begin{tfs}{synsem}
     loc:cat:head: & \begin{tfs}{nominal} \end{tfs} \\
     loc:cat:def: & \tag{2}
   \end{tfs} 
            \right\rangle\\
  def: & \tag{2}\begin{tfs}{boolean} \end{tfs}
 \end{tfs} \\ 
 cont: & \begin{tfs}{nom\_obj}
  index: & \tag{1}\begin{tfs}{index}
   per: & \begin{tfs}{3rd} \end{tfs} \\
   num: & \begin{tfs}{pl} \end{tfs} \\
   gen: & \begin{tfs}{m} \end{tfs} 
  \end{tfs} \\ 
  restr: & \left\{ \begin{tfs}{flower} inst: & \tag{1} \end{tfs} \right\}
 \end{tfs} 
\end{tfs} 
\end{eqnarray*}
}
\mycaption{The effect of the \nismak\ lexical rule}{ךמסנל ילקיסקלה קוחה}
\label{fig:nlr}
\end{figure}

Notice that the lexical entry of the \nismak\ noun {\sf pirxei}, listed
in figure~\ref{fig:nlr}, does not specify any value for the DEF
feature. Hence, the DLR cannot be applied to {\sf pirxei}, as it only
applies for nominals that are specified as DEF~$-$. This corresponds
to the observation that \nismak\ nouns cannot be modified by the
definite article in Hebrew.
Once the \nismak\ lexical rule is applied to \nismak-form nouns, their
lexical entry specifies that they subcategorize for a nominal
complement. Noun-noun compounds can thus be constructed by the
head-complement schema (figure~\ref{fig:schema2}).

\chapter{Conclusion}
\label{sec:summary}
As linguistic formalisms become more rigorous, the necessity of well
defined semantics for grammars increases. We presented
an operational semantics for TFS-based formalisms, making use of an
abstract machine specifically tailored for this kind of
applications. In addition, we described a compiler for a general
TFS-based language. The compiled code, in terms of abstract machine
instructions, can be interpreted and executed on ordinary hardware.

We have formalized in this thesis the concepts of {\em grammars} and
{\em languages} for linguistic formalisms that are based on typed
feature structures, using the notion of {\em multi-rooted structures}
that generalize feature structures. We use multi-rooted structures for
representing grammar {\em rules} as well as (the equivalent of)
sentential forms that are generated during parsing. We described a
computational process that corresponds to parsing with respect to such
formalisms. We thus achieved two different specifications, namely a
declarative (derivation-based) one and an algebraic
(computation-based) one, for the semantics of those formalisms. Next,
we have proved that the two specifications coincide, namely that the
computational process induced by the algebraic specification is
correct with respect to the declarative specification. Finally, we
formally characterized a subset of the grammars, {\em off-line
parsable} ones, for which termination of parsing can be
guaranteed. Making use of the well-foundedness of the subsumption
relation, we proved that for every grammar in this class, parsing is
finitely terminating.

This view of parsing with typed feature structures is the basis for
the design of \amalia, an abstract machine specifically tailored for
executing code that is compiled from grammars. We detailed the
architecture of the machine, its data structures and instruction set,
along with the process of compilation of ALE grammars.
The use of abstract machine techniques results in highly efficient
processing.  The system was implemented and a comparison to ALE shows
a great improvement in both compilation and execution times.

The current implementation of \amalia\ is not fully compatible with
ALE. Several features of ALE are missing in our implementation, and
therefore a natural extension of this project would be to add them.
Most notably, \amalia\ doesn't support the use of lexical rules, which
are considered important for any reasonable grammar of natural
languages. ALE also includes a component of definite clauses over TFSs
which is missing in \amalia\ -- an interesting extension would be to
link the abstract machine with a WAM-like machine that can handle
definite clauses.

\amalia's current compiler is relatively basic, and several
optimizations might be introduced to it in the future. A major
optimization might be achieved by incorporating static (compile-time)
analysis of grammars. Several interesting questions, relating grammars
to computer programs, arise: for example, what is the equivalent of
{\em dead-code elimination}? How can concepts of {\em structured
programming\/} be transferred to grammars? Can {\em modules\/} be
defined for grammars, too?

A different line of improvements concerns the parsing algorithm
incorporated by \amalia. Currently only one, relatively simple,
algorithm is inherent to the machine. An interesting extension would
be to implement various algorithms, probably with user control over
them, and to experiment their time and space efficiency.

\amalia\ is currently being used as a platform for developing an HPSG
grammar for the Hebrew language. While this endeavor is still
underway, it serves as a realistic use of the system. The development
of the grammar already resulted in many improvements and extensions to
\amalia, and the system proved stable and reliable enough to support
it. We presented a partial HPSG-based grammar for a fragment of Hebrew,
concentrating on noun phrases. We hope that this endeavor will serve
as the basis for a more comprehensive, broad-coverage grammar of Hebrew.

\appendix
\chapter{List of Machine Instructions}
\label{inst-list}
The following table lists, for quick reference, the machine
instructions and functions, accompanied by a reference to the page in
the text in which they are described.

\begin{table}[h]
\begin{center}
\begin{minipage}{6cm}
\begin{tabular}{lr}
{\bf Query processing}          &				\\
{\mf put\_node t/n, $X_i$}      & \pageref{inst:put-node}       \\
{\mf put\_arc $X_i$,offset,$X_j$}& \pageref{inst:put-arc}       \\
{\mf proceed $X_i$}          	& \pageref{proceed}      	\\ 
\\
{\bf Type unification}  	&                       	\\
{\mf build\_str t}    		& \pageref{inst:unify-type}     \\
{\mf build\_ref\_and\_unify i}  & \pageref{inst:unify-type}     \\
{\mf build\_ref i}     		& \pageref{inst:unify-type}     \\
{\mf build\_self\_ref i}    	& \pageref{inst:unify-type}     \\
{\mf build\_var t}    		& \pageref{inst:unify-type}     \\
{\mf unify\_feat i}    		& \pageref{inst:unify-type}     \\
\\
{\bf Program processing}        &                               \\
{\mf get\_structure t/n,$X_i$}  & \pageref{inst:get-structure}  \\
{\mf unify\_variable $X_i$}     & \pageref{inst:unify-variable} \\
{\mf unify\_value $X_i$}        & \pageref{inst:unify-value}    \\
{\mf put\_rule $L$}		& \pageref{put-rule}		\\
\end{tabular}
\end{minipage} \hspace{1cm}
\begin{minipage}{6cm}
\begin{tabular}{lr}
{\bf Control}			&				\\
{\mf first\_key}		& \pageref{key-insts}		\\
{\mf next\_key}			& \pageref{key-insts}		\\
{\mf check\_key}		& \pageref{key-insts}		\\
{\mf tst\_active\_edges $l$}	& \pageref{fig:edge-insts}	\\
{\mf next\_active\_edge $l$}	& \pageref{fig:edge-insts}	\\
{\mf tst\_complete\_edges $l'$}	& \pageref{fig:edge-insts}	\\
{\mf next\_complete\_edge $l$}	& \pageref{fig:edge-insts}	\\
{\mf call}			& \pageref{fig:call-inst}	\\
{\mf load\_fs $r$}		& \pageref{fig:rule-insts}	\\
{\mf copy\_active\_edge $l$}	& \pageref{fig:rule-insts}	\\
{\mf copy\_complete\_edge $X$}	& \pageref{fig:rule-insts}	\\
\\
{\bf Auxiliary functions}       &                               \\
{\mf bind(addr1,addr2)}         & \pageref{fun:bind}   \\
{\mf deref(a):address}          & \pageref{fun:deref}   \\
{\mf unify(addr1,addr2):boolean}& \pageref{fun:unify}   \\
{\mf fail()}                    & \pageref{fail}    \\
\end{tabular}
\end{minipage}
\end{center}
\end{table}

\chapter{The Hebrew Grammar}
\label{app:grammar}
\begin{verbatim}
%%%%%%%%%%%%%%%%%%%%%%%%%%%%%%%%%%%%%%%%%%%%%%%%%%%%%%%%%%%%%%%%%%%
%
%  File: Hebrew grammar
%
%  Includes: 1. Schema 1
%            2. Schema 2
%            3. Schema 4
%            4. Schema 5
%            5. 
%            6. Head feature principle
%            7. Valence principle (subj, comps, spr)
%            8. Semantics principle (cont) - partial
%            9. Contextual Consistency (conx) - approximate
%           10. Quantifier storage - preliminary
% 
%
%%%%%%%%%%%%%%%%%%%%%%%%%%%%%%%%%%%%%%%%%%%%%%%%%%%%%%%%%%%%%%%%%%%

%%%**********************  Type Hierarchy
%th

bot sub [sign,list,
         set,cat,sem_obj,sem_det,conx,qfpsoa,index,per,num,gend,
         head,vform,pform,defness,marking,boolean].

sign sub [word,phrase] intro [cat:cat,cont:sem_obj,conx:conx,qstore:set_quant].
  word sub [].
  phrase sub [].

cat sub [] intro [head:head,subj:list,comps:list,spr:list,marking:marking].

sem_obj sub [psoa,nom_obj,quant].
  nom_obj sub [pron,npro] intro [index:index,restr:set_psoa].
    pron sub [].
    npro sub [].
  psoa sub [] intro [nucleus:qfpsoa].
  quant sub [] intro [det:sem_det, restind:npro].

    sem_det sub [forall,exists,the].
      forall sub [].
      exists sub [].
      the sub [].

conx sub [] intro [backgr:set_psoa].
    
qfpsoa sub [un_relation,cn,naming].
  un_relation sub [walk,sing,red,big, bin_relation] intro [agent:index].
    walk sub [].
    sing sub [].
    red sub [].
    big sub [].
  bin_relation sub [see,eat, tri_relation] intro [theme:index].
    see sub [].
    eat sub [].
  tri_relation sub [sell,give] intro [patient:index].
    sell sub [].
    give sub [].
  cn sub [book,apple] intro [instance:index].
    book sub [].
    apple sub [].
  naming sub [dan,dana] intro [bearer:index].
    dan sub [].
    dana sub [].

index sub [] intro [per:per,num:num,gend:gend].
  per sub [first,second,third].
    first sub [].
    second sub [].
    third sub [].
  num sub [sg,pl].
    sg sub [].
    pl sub [].
  gend sub [masc,fem].
    masc sub [].
    fem sub [].

head sub [subst,func].
  func sub [mark] intro [spec:sign].
    mark sub [det].
      det sub [].
  subst sub [nominal,verb,prep].
  nominal sub [noun,adj,numeral] intro [defness:defness].
      noun sub [].
      adj sub [] intro [mod:sign].
      numeral sub [].
    prep sub [] intro [pform:pform].
    verb sub [] intro [vform:vform].
% I decided not to add a 'mod' feature to all substantials, since in
% most of the cases (excluding adjectives) its value is 'none'.

defness sub [indef,def].
  indef sub [].
  def sub [].

vform sub [fin,bse].
  fin sub [].
  bse sub [].

pform sub [to,in].
  to sub [].
  in sub [].

marking sub [marked,unmarked].
  marked sub [comp,determiner,quantifier].
    comp sub [].
    determiner sub [].
    quantifier sub [].
  unmarked sub [].

boolean sub [yes,no].
  yes sub [].
  no sub [].

list sub [e_list,ne_list].
  ne_list sub [] intro [hd:bot,tl:list].
  e_list sub [].

set sub [e_set,ne_set,set_psoa,set_quant].
  e_set sub [].
  ne_set sub [ne_set_psoa,ne_set_quant] intro [elt:bot,elts:set].
  set_psoa sub [e_set, ne_set_psoa].
    ne_set_psoa sub []. % intro [elt:psoa, elts:set_psoa].
  set_quant sub [e_set, ne_set_quant].
    ne_set_quant sub []. %  intro [elt:quant, elts:set_quant].

%%%**********************  Macros
%macros

propn(Num,Gen,Name) macro
(word,cat:(cat,head:noun,subj:e_list,comps:e_list,spr:e_list),
      cont:(npro,index:(Ind,per:third,num:Num,gend:Gen),restr:elt:Sem),
      conx:backgr:(ne_set_psoa,elt:(Sem,psoa,nucleus:(Name,bearer:Ind)),
                               elts:e_set)).

np(Per,Num,Gen,Index) macro
(sign, cat:head:noun,
       cont:(nom_obj,index:(Index,per:Per,num:Num,gend:Gen))).

noun(Num,Gen,Sem,Def) macro
(word,cat:(head:(noun,defness:Def),subj:e_list,comps:e_list,spr:e_list),
      cont:(nom_obj,index:(Ind,per:third,num:Num,gend:Gen),
            restr:(elt:(psoa,nucleus:(Sem,instance:Ind)),elts:e_set)),
      qstore:e_set).

intrans macro (cat:comps:e_list).

trans macro
(cat:comps:(hd:(@ np(Per,Num,Gen,Theme)),tl:e_list),
 cont:nucleus:theme:Theme).

ditrans macro
(cat:comps:(hd:(@ np(Per,Num,Gen,Theme)),
            tl:hd:(@ np(Per1,Num1,Gen1,Patient)),
            tl:tl:e_list),
 cont:nucleus:patient:Patient).

verb(Per,Num,Gen,Sem) macro
(word,cat:(cat,head:(verb,vform:fin),
               subj:hd:(@ np(Per,Num,Gen,SubjInd)),
               marking:unmarked),
     cont:nucleus:(Sem,agent:SubjInd),
     conx:backgr:e_set,
     qstore:e_set).

nominal(Def,Ind) macro
(sign,cat:(head:(nominal,defness:Def),
           subj:e_list,comps:e_list,spr:e_list,marking:unmarked),
      cont:(nom_obj,index:Ind)).

adj(Num,Gen,Sem,Def) macro
(word,cat:(head:(adj,defness:Def,
                 mod:(@ nominal(Def,Ind))),
           subj:e_list,comps:e_list,spr:e_list),
      cont:(nom_obj,index:(Ind,num:Num,gend:Gen),
            restr:(elt:(psoa,nucleus:Sem,nucleus:agent:ModInd))),
      qstore:e_set).

*****  Empty Categories

%%%**********************  Grammmar Rules
%grammar

% Schema 1 (ch. 9, p. 347)
% Subject - Head
subject_head rule 

(phrase,cat:(cat,head:Head,
             subj:e_list,
             comps:Comps,
             spr:Spr,marking:Marking),
       cont:Cont,
       conx:backgr:BM,
       qstore:QM)
===>
cat>			% subject
(Subj,sign,cat:cat,
        cont:sem_obj,
        conx:backgr:BS,
        qstore:QS),	
cat>			% head
(sign,cat:(cat,head:Head,
             subj:(hd:Subj,tl:e_list),
             comps:(Comps,e_list),	% comps is required to be empty so that
				% subject is added after all the complements.
             spr:Spr,marking:Marking),
     cont:(Cont,sem_obj),
     conx:backgr:BH,
     qstore:QH),
goal> union(QS,QH,QM),
goal> union(BS,BH,BM).

% Schema 2 (ch. 9, p. 348)
% Head - Complement
head_complement rule

(phrase,cat:(cat,head:Head,
             subj:Subj,
             comps:Comps,
             spr:Spr,marking:Marking),
       cont:Cont,
       conx:backgr:BM,
       qstore:QM)
===>
cat>
(sign,cat:(cat,head:Head,	% head
             subj:Subj,
             comps:(hd:(Comp,sign,cat:cat,
                                      cont:sem_obj,
                                      conx:backgr:BC,
                                      qstore:QC),
                           tl:Comps),
             spr:Spr,marking:Marking),
     cont:Cont,
     conx:backgr:BH,
     qstore:QH),
cat>
Comp,				% complement
goal> union(BH,BC,BM),
goal> union(QH,QC,QM).


% Schema 4 (ch. 1, p. 51)
% Marker - Head
marker_head rule

(phrase,cat:(cat,head:Head,
             subj:Subj,
             comps:Comps,
             spr:Spr,marking:Marking),
       cont:Cont,
       conx:backgr:BM,
       qstore:(elt:Elt,elts:Elts))
===>
cat>					% marker
(word,cat:(cat,head:(mark,spec:HeadDtr),
                subj:list,comps:list,spr:list,marking:(Marking,marked)),
        cont:(Elt,quant,det:sem_det,restind:sem_obj),
        conx:backgr:BD,
        qstore:e_set),
cat>					% head
(HeadDtr,sign,cat:(cat,
                   head:Head,
                   subj:Subj,
                   comps:Comps, 
                   spr:Spr, marking:unmarked),
	cont:Cont,
	conx:backgr:BH,
	qstore:Elts),
goal> union(BD,BH,BM).	


empty
(sign,cat:(head:(det,spec:(sign,cat:(head:noun,subj:e_list,comps:e_list,
                                     spr:e_list),
                                cont:(Npro,index:(per:third,num:sg)),
                                qstore:e_set)),
           subj:e_list,comps:e_list,spr:e_list,marking:quantifier),
      cont:(quant,det:exists,restind:Npro),
      conx:backgr:e_set,
      qstore:e_set).

% Schema 5 (ch. 9, p. 403)
% Head - Adjunct
%
% modification: the marking feature is shared by the adjunct and the
% head (to require definiteness agreement)
head_adjunct rule

(phrase,cat:Cat,
       cont:Cont,
       conx:backgr:BM,
       qstore:QM)
===>
cat>			% head
(HeadDtr,sign,cat:Cat,
        cont:sem_obj,
        conx:backgr:BH,
        qstore:QH),
cat>			% adjunct
(sign,cat:head:(adj,defness:defness,mod:HeadDtr),
        cont:Cont,
        conx:backgr:BA,
        qstore:QA),
goal> union(BH,BA,BM),
goal> union(QH,QA,QM).

%%%**********************  Lexical Entries
%lexicon

dan --->
(@ propn(sg,masc,dan)).

dana --->
(@ propn(sg,fem,dana)).

sepr --->
(@ noun(sg,masc,book,indef)).

ha-sepr --->
(@ noun(sg,masc,book,def)).

sparim --->
(@ noun(pl,masc,book,indef)).

$ar --->
(@ verb(third,sg,masc,sing),(@ intrans)).

$ara --->
(@ verb(third,sg,fem,sing),(@ intrans)).

^akal --->
(@ verb(third,sg,masc,eat),(@ trans)).

natan --->
(@ verb(third,sg,masc,give),(@ ditrans)).


^adomm --->
(@ adj(sg,masc,red,indef)).

gadol --->
(@ adj(sg,masc,big,indef)).

ha-gadol --->
(@ adj(sg,masc,big,def)).

gdolim --->
(@ adj(pl,masc,big,indef)).
\end{verbatim}

\bibliographystyle{fullname}
\addcontentsline{toc}{chapter}{\protect\numberline{}{Bibliography}}

\begin{thebibliography}{}

\bibitem[\protect\citename{Aho and Ullman}1972]{parsing}
Aho, Alfred~V. and Jeffrey~D. Ullman.
\newblock 1972.
\newblock {\em The Theory of Parsing, Translation and Compiling}, volume 1:
  Parsing.
\newblock Prentice-Hall, Inc., Englewood Cliffs, N.J.

\bibitem[\protect\citename{A\"{\i}t-Kaci}1991]{wam}
A\"{\i}t-Kaci, Hassan.
\newblock 1991.
\newblock {\em Warren's Abstract Machine: A Tutorial Reconstruction}.
\newblock Logic Programming. The {MIT} Press, Cambridge, Massachusetts.

\bibitem[\protect\citename{A\"{\i}t-Kaci \bgroup et al.\egroup
  }1989]{lattice-ops}
A\"{\i}t-Kaci, Hassan, Robert Boyer, Patrick Lincoln, and Roger Nasr.
\newblock 1989.
\newblock Efficient implementation of lattice operations.
\newblock {\em {ACM} Transactions on Programming Languages and Systems},
  11(1):115--146, January.

\bibitem[\protect\citename{A\"{\i}t-Kaci and Di~Cosmo}1993]{prl7}
A\"{\i}t-Kaci, Hassan and Roberto Di~Cosmo.
\newblock 1993.
\newblock Compiling order-sorted feature term unification.
\newblock {PRL} Technical Note~7, Digital Paris Research Laboratory, December.

\bibitem[\protect\citename{A\"{\i}t-Kaci and Podelski}1993]{life-meaning}
A\"{\i}t-Kaci, Hassan and Andreas Podelski.
\newblock 1993.
\newblock Towards a meaning of {LIFE}.
\newblock {\em Journal of Logic Programming}, 16(3-4):195--234, July-August.

\bibitem[\protect\citename{Barton, Berwick, and Ristad}1987]{barberris}
Barton, G.~Edward, Robert~C. Berwick, and Eric~Sven Ristad.
\newblock 1987.
\newblock {\em Computational Complexity and Natural Language}.
\newblock Computational Models of Cognition and Perception. {MIT} Press,
  Cambridge, Mass.

\bibitem[\protect\citename{Beierle and Meyer}1994]{beierle-meyer}
Beierle, Christoph and Gregor Meyer.
\newblock 1994.
\newblock Run-time type computations in the {W}arren abstract machine.
\newblock {\em Journal of Logic Programming}, 18:123--148.

\bibitem[\protect\citename{Bentur, Angel, and Segev}1992]{ibm92}
Bentur, Esther, Aviella Angel, and Danit Segev.
\newblock 1992.
\newblock Computerized analysis of {H}ebrew words.
\newblock {\em Hebrew Linguistics}, 36:33--38, December.
\newblock (in {H}ebrew).

\bibitem[\protect\citename{Bird}1990]{bird90}
Bird, Steven.
\newblock 1990.
\newblock {\em Constraint-Based Phonology}.
\newblock {Ph.D.} thesis, University of Edinburgh.

\bibitem[\protect\citename{Bird}1992]{bird92}
Bird, Steven.
\newblock 1992.
\newblock Finite state phonology in {HPSG}.
\newblock In {\em Proceedings of {COLING-92}}, pages 74--80.

\bibitem[\protect\citename{Calcagno, Kathol, and Pollard}1993]{hpsg-bib}
Calcagno, Mike, Andreas Kathol, and Carl Pollard.
\newblock 1993.
\newblock A bibliography of books, theses and articles in or on {HPSG}.
\newblock Unpublished manuscript, August.

\bibitem[\protect\citename{Carpenter}1991]{carp91}
Carpenter, Bob.
\newblock 1991.
\newblock Typed feature structures: A generalization of first-order terms.
\newblock In Vijai Saraswat and Ueda Kazunori, editors, {\em Logic Programming
  -- Proceedings of the 1991 International Symposium}, pages 187--201,
  Cambridge, MA. {MIT} Press.

\bibitem[\protect\citename{Carpenter}1992a]{ale}
Carpenter, Bob.
\newblock 1992a.
\newblock {ALE} -- the attribute logic engine: User's guide.
\newblock Technical report, Laboratory for Computational Linguistics,
  Philosophy Department, Carnegie Mellon University, Pittsburgh, PA 15213,
  December.

\bibitem[\protect\citename{Carpenter}1992b]{carp92}
Carpenter, Bob.
\newblock 1992b.
\newblock {\em The Logic of Typed Feature Structures}.
\newblock Cambridge Tracts in Theoretical Computer Science. Cambridge
  University Press.

\bibitem[\protect\citename{Chayen and Dror}1976]{chendror}
Chayen, M.~J. and Z.~Dror.
\newblock 1976.
\newblock {\em Introduction to {H}ebrew Transformational Grammar}.
\newblock University Publishing Projects Ltd., Jerusalem.
\newblock (in Hebrew).

\bibitem[\protect\citename{Choueka and Ne'eman}1995]{choueka95}
Choueka, Yaacov and Yoni Ne'eman.
\newblock 1995.
\newblock {"Nakdan-T"}, a text vocalizer for modern {H}ebrew.
\newblock In {\em Proceedings of the Fourth {B}ar-{I}lan Symposium on
  Foundations of Artificial Intelligence}, June.

\bibitem[\protect\citename{Cohen}1984]{cohen84}
Cohen, Daniel.
\newblock 1984.
\newblock {\em Mechanical Syntactic Analysis of a {H}ebrew Sentence}.
\newblock {Ph.D.} thesis, Hebrew University of Jerusalem.
\newblock (In Hebrew).

\bibitem[\protect\citename{D\"{o}rre and Dorna}1993]{cuf}
D\"{o}rre, Jochen and Michael Dorna.
\newblock 1993.
\newblock {CUF} -- a formalism for linguistic knowledge representation.
\newblock In Jochen D\"{o}rre, editor, {\em Computatioal Aspects of
  Constrained-Based Linguistic Description {I}, {DYANA-2} delivarable
  {R1.2.A}}, {ILLC}/Department of Philosophy, University of Amsterdam, Nieuwe
  Doelenstraat 15, NL-1012 CP Amsterdam, August.

\bibitem[\protect\citename{Emerson}1983]{emerson}
Emerson, E.~Allen.
\newblock 1983.
\newblock Alternative semantics for temporal logics.
\newblock {\em Theoretical Computer Science}, 26:121--130.

\bibitem[\protect\citename{Erbach}1994]{profit}
Erbach, Gregor.
\newblock 1994.
\newblock {ProFIT}: Prolog with features, inheritance and templates.
\newblock {CLAUS} Report~42, Computerlinguistik, Universit\"at des Saarlandes,
  D-66041, Saarbr\"ucken, Germany, July.

\bibitem[\protect\citename{Franz}1990]{franz}
Franz, Alex.
\newblock 1990.
\newblock A parser for {HPSG}.
\newblock Report {CMU-LCL-90-3}, Laboratory for Computational Linguistics,
  Department of Philosophy, Carnegie Mellon University, Pittsburgh, PA 15213,
  July.

\bibitem[\protect\citename{Gabrilovich}In preparation]{gabr:thesis}
Gabrilovich, Evgeniy.
\newblock In preparation.
\newblock Natural language generation by abstract machine.
\newblock Master's thesis, Technion, Israel Institute of Technology, Haifa,
  Israel.

\bibitem[\protect\citename{Gazdar \bgroup et al.\egroup }1985]{gpsg}
Gazdar, G.~E., E.~Klein, K.~Pullum, and Ivan~A. Sag.
\newblock 1985.
\newblock {\em Generalized Phrase Structure Grammar}.
\newblock Harvard University Press, Cambridge, Mass.

\bibitem[\protect\citename{Gazdar and Mellish}1989]{gazmel}
Gazdar, Gerald and Chris Mellish.
\newblock 1989.
\newblock {\em Natural Language Processing in {PROLOG}}.
\newblock Addison-Wesley.

\bibitem[\protect\citename{Gerdemann}1993]{troll}
Gerdemann, Dale.
\newblock 1993.
\newblock Troll: Type resolution system -- fundamental principles and user's
  guide.
\newblock Unpublished manuscript, September.

\bibitem[\protect\citename{Gerdemann and Hinrichs}1988]{unicorn}
Gerdemann, Dale and Erhard~W. Hinrichs.
\newblock 1988.
\newblock Unicorn: A unification parser for attribute-value grammars.
\newblock {\em Studies in the Linguistic Sciences}, 18(2):41--86.

\bibitem[\protect\citename{Glinert}1989]{glinert}
Glinert, Lewis.
\newblock 1989.
\newblock {\em The Grammar of Modern {H}ebrew}.
\newblock Cambridge University Press, Cambridge.

\bibitem[\protect\citename{G\"{o}tz}1994]{thilo:master}
G\"{o}tz, Thilo~W.
\newblock 1994.
\newblock A normal form for types feature structures.
\newblock Master's thesis, Eberhard-Karls Universit\"at, T\"ubingen, March.

\bibitem[\protect\citename{Haas}1989]{haas}
Haas, Andrew.
\newblock 1989.
\newblock A parsing algorithm for unification grammar.
\newblock {\em Computational Linguistics}, 15(4):219--232, December.

\bibitem[\protect\citename{Hermenegildo}1986]{herm86}
Hermenegildo, Manuel~V.
\newblock 1986.
\newblock {\em An Abstract Machine Based Execution Model for Computer
  Architecture Design and Efficient Implementation of Logic Programs in
  Parallel}.
\newblock {Ph.D.} thesis, Department of Computer Science, The University of
  Texas at Austin, Austin, Texas 78712-1188, August.

\bibitem[\protect\citename{Johnson}1988]{johnson88}
Johnson, Mark.
\newblock 1988.
\newblock {\em Attribute-Value Logic and the Theory of Grammar}, volume~16 of
  {\em {CSLI} Lecture Notes}.
\newblock {CSLI}, Stanford, California.

\bibitem[\protect\citename{{{JPSG} Working Group}}In Preparation]{japanese}
{{JPSG} Working Group}.
\newblock In Preparation.
\newblock {JPSG}: A constraint based approach to japanese grammar.
\newblock Technical memo, Institute for New Generation Computer Technology.

\bibitem[\protect\citename{Kaplan and Bresnan}1982]{lfg}
Kaplan, R. and J.~Bresnan.
\newblock 1982.
\newblock Lexical functional grammar: A formal system for grammatical
  representation.
\newblock In J.~Bresnan, editor, {\em The Mental Representation of Grammatical
  Relations}. MIT Press, Cambridge, Mass., pages 173--281.

\bibitem[\protect\citename{Kaplan}1973]{kaplan73}
Kaplan, Ronald~M.
\newblock 1973.
\newblock A general syntactic processor.
\newblock In Randall Rustin, editor, {\em Natural Language Processing},
  number~8 in Courant Computer Science Symposium. Algorithmics Press, P.O. Box
  97, New York, N.Y. 10012, pages 194--241.

\bibitem[\protect\citename{Kasper and Rounds}1986]{kasperounds}
Kasper, Robert~T. and William~B. Rounds.
\newblock 1986.
\newblock A logical semantics for feature structures.
\newblock In {\em Proceedings of the 24th Annual Meeting of the Association for
  Computational Linguistics}, pages 257--265.

\bibitem[\protect\citename{Kay}1973]{kay73}
Kay, Martin.
\newblock 1973.
\newblock The {MIND} system.
\newblock In Randall Rustin, editor, {\em Natural Language Processing},
  number~8 in Courant Computer Science Symposium. Algorithmics Press, P.O. Box
  97, New York, N.Y. 10012, pages 155--188.

\bibitem[\protect\citename{Kay}1983]{fug}
Kay, Martin.
\newblock 1983.
\newblock Unification grammar.
\newblock Technical report, Xerox Palo Alto Research Center, Palo Alto, CA.

\bibitem[\protect\citename{King}1989]{king89}
King, Paul.
\newblock 1989.
\newblock {\em A Logical Formalism for {HPSG}}.
\newblock Doctoral dissertation, University of Manchester, Manchester, UK.

\bibitem[\protect\citename{King}1992]{king92}
King, Paul.
\newblock 1992.
\newblock Unification grammars and descriptive formalisms: Lecture notes for a
  graduate level course.
\newblock Unpublished manuscript.

\bibitem[\protect\citename{Kodri\u{c}, Popowich, and Vogel}1992]{hpsg-pl}
Kodri\u{c}, Sandi, Fred Popowich, and Carl Vogel.
\newblock 1992.
\newblock The {HPSG-PL} system.
\newblock {CSS-IS TR} 92-05, School of Computing Science, Centre for Systems
  Science, Simon Fraser University, Burnaby, B. C. Canada V5A 1S6, Jun.

\bibitem[\protect\citename{Kolliakou}1996]{greek}
Kolliakou, Dimitra.
\newblock 1996.
\newblock Definiteness and the make-up of nominal categories.
\newblock In Claire Grover and Enric Vallduv\'\i, editors, {\em Studies in
  {HPSG}}, volume~12 of {\em Edinburgh Working Papers in Cognitive Science}.
  Centre for Cognitive Science, The University of {E}dinburgh, chapter~4, pages
  121--164.

\bibitem[\protect\citename{Kursawe}1987]{kursawe}
Kursawe, Peter.
\newblock 1987.
\newblock How to invent a {P}rolog machine.
\newblock {\em New Generation Computing}, 5:97--114.

\bibitem[\protect\citename{Landin}1964]{landin}
Landin, P.~J.
\newblock 1964.
\newblock The mechanical evaluation of expressions.
\newblock {\em The Computer Journal}, 6(4):308--320.

\bibitem[\protect\citename{Lloyd}1987]{lloyd}
Lloyd, John~Wylie.
\newblock 1987.
\newblock {\em Foundations of Logic Programming}.
\newblock Springer series in Symbolic Computation -- Artificial Intelligence.
  Springer, Berlin, second edition.

\bibitem[\protect\citename{Manaster-Ramer}1987]{manaster}
Manaster-Ramer, Alexis.
\newblock 1987.
\newblock {\em Mathematics of Language}.
\newblock John Benjamins Publishing Company, Amsterdam/Philadelphia.

\bibitem[\protect\citename{Moshier}1988]{moshier}
Moshier, Drew.
\newblock 1988.
\newblock {\em Extensions to Unification Grammars for the Description of
  Programming Languages}.
\newblock {Ph.D.} thesis, University of Michigan, Ann Arbor.

\bibitem[\protect\citename{Moshier and Rounds}1987]{moshier-rounds}
Moshier, Drew~M. and William~C. Rounds.
\newblock 1987.
\newblock A logic for partially specified data structures.
\newblock In {\em 14th Annual {ACM} Symposium on Principles of Programming
  Languages}, pages 156--167, January.

\bibitem[\protect\citename{M\"uller}1996]{hpsgbib}
M\"uller, Stefan.
\newblock 1996.
\newblock {HPSG} literature.
\newblock Available as an electronic document at
  \verb+http://cl-www.dfki.uni-sb.de/HPSG/+.

\bibitem[\protect\citename{Nerbonne}1992]{nerbonne92}
Nerbonne, John.
\newblock 1992.
\newblock Feature-based lexicons: An example and a comparison to {DATR}.
\newblock Research Report RR-92-04, Deutsches Forschungszentrum f\"{u}r
  K\"{u}nstliche Intelligenz {GmbH}, February.

\bibitem[\protect\citename{Nerbonne, Netter, and Pollard}1994]{german}
Nerbonne, John, Klaus Netter, and Carl Pollard, editors.
\newblock 1994.
\newblock {\em German in {H}ead-{D}riven {P}hrase {S}tructure {G}rammar}.
\newblock Number~46 in {CSLI} lecture notes. {CSLI}, Stanford, Ca.

\bibitem[\protect\citename{Netter}1994]{netter94}
Netter, Klaus.
\newblock 1994.
\newblock Towards a theory of functional heads.
\newblock In John Nerbonne, Klaus Netter, and Carl Pollard, editors, {\em
  {G}erman in {H}ead-{D}riven {P}hrase {S}tructure {G}rammar}, volume~46 of
  {\em {CSLI} Lecture Notes}. {CSLI}, Stanford, CA, chapter~9, pages 297--340.

\bibitem[\protect\citename{Nilsson}1993]{nilson93}
Nilsson, Ulf.
\newblock 1993.
\newblock Towards a methodology for the design of abstract machines for logic
  programming languages.
\newblock {\em Journal of Logic Programming}, 16:163--189.

\bibitem[\protect\citename{Nirenburg and Ben-Asher}1984]{nirenburg}
Nirenburg, Sergei and Yosef Ben-Asher.
\newblock 1984.
\newblock {HUHU} -- the {H}ebrew {U}niversity {H}ebrew understander.
\newblock {\em Computer Languages}, 9({3/4}).

\bibitem[\protect\citename{Ornan}1964]{ornan64}
Ornan, Uzzi.
\newblock 1964.
\newblock {\em Noun Phrases in Modern {H}ebrew Literature}.
\newblock {Ph.D.} thesis, Hebrew University, Jerusalem.
\newblock (in Hebrew).

\bibitem[\protect\citename{Ornan}1979]{ornan79}
Ornan, Uzzi.
\newblock 1979.
\newblock {\em The Simple Sentence}.
\newblock Academon, Jerusalem, Israel.
\newblock (in Hebrew).

\bibitem[\protect\citename{Ornan}1986]{ornan86}
Ornan, Uzzi.
\newblock 1986.
\newblock Phonemic script: A central vehicle for processing natural language --
  the case of {H}ebrew.
\newblock Technical Report 88.181, {IBM} Research Center, Haifa, Israel.

\bibitem[\protect\citename{Ornan}1994]{ornan94}
Ornan, Uzzi.
\newblock 1994.
\newblock Basic concepts in "romanization" of scripts.
\newblock Technical Report {LCL 94-5}, Laboratory for Computational
  Linguistics, Technion, Haifa, Israel, March.

\bibitem[\protect\citename{Ornan and Katz}1995]{ornan95}
Ornan, Uzzi and Michael Katz.
\newblock 1995.
\newblock A new program for {H}ebrew index based on the {P}honemic {S}cript.
\newblock Technical Report {LCL 94-7}, Laboratory for Computational
  Linguistics, Technion, Haifa, Israel, July.

\bibitem[\protect\citename{Ousterhout}1994]{tcltk}
Ousterhout, John~K.
\newblock 1994.
\newblock {\em {Tcl} and the {Tk} Toolkit}.
\newblock Addison-Wesley Professional Computing Series. Addison-Wesley.

\bibitem[\protect\citename{Penn}1993]{penn}
Penn, Gerald.
\newblock 1993.
\newblock A comprehensive {HPSG} grammar in {ALE}.
\newblock Technical report, Laboratory for Computational Linguistics, Carnegie
  Mellon University, Pittsburgh, PA.

\bibitem[\protect\citename{Pereira and Warren}1983]{parsing-as-deduction}
Pereira, Fernando C.~N. and David H.~D. Warren.
\newblock 1983.
\newblock Parsing as deduction.
\newblock In {\em Proceedings of the 21st Annual Meeting of the Association for
  Computational Linguistics}, pages 137--144, June.

\bibitem[\protect\citename{Pollard and Sag}1987]{hpsg1}
Pollard, Carl and Ivan~A. Sag.
\newblock 1987.
\newblock {\em Information Based Syntax and Semantics}.
\newblock Number~13 in {CSLI} Lecture Notes. CSLI.

\bibitem[\protect\citename{Pollard and Sag}1994]{hpsg2}
Pollard, Carl and Ivan~A. Sag.
\newblock 1994.
\newblock {\em Head-Driven Phrase Structure Grammar}.
\newblock University of Chicago Press and CSLI Publications.

\bibitem[\protect\citename{Pollard and Moshier}1990]{polmosh90}
Pollard, Carl~J. and M.~Drew Moshier.
\newblock 1990.
\newblock Unifying partial descriptions of sets.
\newblock In Philip~P. Hanson, editor, {\em Information, Language and
  Cognition}, volume~1 of {\em Vancouver Studies in Cognitive Science}.
  University of British Columbia Press, Vancouver 1990, chapter~10, pages
  285--322.

\bibitem[\protect\citename{Popowich and Vogel}1991]{popvog91}
Popowich, Fred and Carl Vogel.
\newblock 1991.
\newblock A logic-based implementation of head-driven phrase structure grammar.
\newblock In Charles~Grant Brown and Gregers Koch, editors, {\em Natural
  Language Understanding and Logic Programming, {III}}, pages 227--245.
  Elsevier Science Publishers (North-Holland).

\bibitem[\protect\citename{Prudian and Pollard}1985]{prupol85}
Prudian, Derek and Carl Pollard.
\newblock 1985.
\newblock Parsing head-driven phrase structure grammar.
\newblock In {\em Proceedings of the 23rd Annual Meeting of the Association for
  Computational Linguistics}, Chicago, IL. University of Chicago.

\bibitem[\protect\citename{Russell}1993]{russell}
Russell, Dale~W.
\newblock 1993.
\newblock {\em Language Acquisition in a Unification-Based Grammar Processing
  System Using a Real-World Knowledge Base}.
\newblock {Ph.D.} thesis, Department of Computer Science, University of
  Illinois at Urbana-Champaign, Urbana, Illinois, July.

\bibitem[\protect\citename{Russinoff}1992]{russ92}
Russinoff, David~M.
\newblock 1992.
\newblock A verified {P}rolog compiler for the {Warren Abstract Machine}.
\newblock {\em Journal of Logic Programming}, 13:367--412.

\bibitem[\protect\citename{Sag and Fodor}1994]{sag-fodor94}
Sag, Ivan~A. and Janet~Dean Fodor.
\newblock 1994.
\newblock Extraction without traces.
\newblock In Raul Aranovich, William Byrne, Susanne Preuss, and Martha
  Senturia, editors, {\em Proceedings of the Thirteenth West Coast Conference
  on Formal Linguistics}, Stanford. CSLI Publications/SLA.

\bibitem[\protect\citename{Samuelsson}1995]{samuelsson95}
Samuelsson, Christer.
\newblock 1995.
\newblock An efficient algorithm for surface generation.
\newblock In {\em Proceedings of the International Joint Conference on
  Artificial Intelligence}.

\bibitem[\protect\citename{Schumann}1991]{schumann}
Schumann, Johann M.~Ph.
\newblock 1991.
\newblock {\em Efficient Theorem Provers based on an Abstract Machine}.
\newblock {Ph.D.} thesis, Technische Universit\"{a}t M\"{u}nchen, Institut
  F\"{u}r Informatik.

\bibitem[\protect\citename{Shieber}1985]{restriction}
Shieber, Stuart~M.
\newblock 1985.
\newblock Using restriction to extend parsing algorithms for
  complex-feature-based formalisms.
\newblock In {\em Proceedings of the 23rd Annual Meeting of the ACL}, pages
  145--152, July.

\bibitem[\protect\citename{Shieber}1986]{shieber86}
Shieber, Stuart~M.
\newblock 1986.
\newblock {\em An Introduction to Unification Based Approaches to Grammar}.
\newblock CSLI Lecture Notes. CSLI.

\bibitem[\protect\citename{Shieber}1992]{shieber92}
Shieber, Stuart~M.
\newblock 1992.
\newblock {\em Constraint-Based Grammar Formalisms}.
\newblock {MIT Press}, Cambridge, Mass.

\bibitem[\protect\citename{Shieber, Schabes, and
  Pereira}1994]{deductive-parsing}
Shieber, Stuart~M., Yves Schabes, and Fernando C.~N. Pereira.
\newblock 1994.
\newblock Principles and implementation of deductive parsing.
\newblock Technical Report {TR-11-94}, Center for Research in Computing
  Technology, Division of Applied Sciences, Harvard University, April.

\bibitem[\protect\citename{Sikkel}1993]{sikkel}
Sikkel, Klaas.
\newblock 1993.
\newblock {\em Parsing Schemata}.
\newblock Klaas Sikkel, Enschede.

\bibitem[\protect\citename{Smolka}1988]{smolka}
Smolka, Gert.
\newblock 1988.
\newblock A feature logic with subsorts.
\newblock {LILOG} Report~33, {IBM} Deutschland, May.

\bibitem[\protect\citename{Smolka and Treinen}1994]{smolka-treinen}
Smolka, Gert and Ralf Treinen.
\newblock 1994.
\newblock Records for logic programming.
\newblock {\em Journal of Logic Programming}, 18:229--258.

\bibitem[\protect\citename{Swift and Warren}1993]{oldt}
Swift, Terrance and David Warren.
\newblock 1993.
\newblock Compiling {OLDT} evaluation: Background and overview.
\newblock Technical Report 92/04, {SUNY} Stony Brook, March.

\bibitem[\protect\citename{Warren}1983]{waren83}
Warren, David H.~D.
\newblock 1983.
\newblock An abstract {P}rolog instruction set.
\newblock Technical Note 309, {SRI} International, Menlo Park, CA., August.

\bibitem[\protect\citename{Warshall}1962]{warshall}
Warshall, Stephen.
\newblock 1962.
\newblock A theorem on boolean matrices.
\newblock {\em Journal of the {ACM}}, 9(1):11--12, January.

\bibitem[\protect\citename{Wintner}1991]{shuly:master}
Wintner, Shuly.
\newblock 1991.
\newblock Syntactic analysis of {H}ebrew sentences.
\newblock Master's thesis, Technion, Israel Institute of Technology, Haifa,
  Israel, July.
\newblock (In Hebrew, abstract in English).

\bibitem[\protect\citename{Wintner}1992]{shuly:patr}
Wintner, Shuly.
\newblock 1992.
\newblock Syntactic analysis of {H}ebrew sentences using {PATR}.
\newblock In Uzzi Ornan, Gideon Ariely, and Edit Doron, editors, {\em Hebrew
  Computational Linguistics}. Ministry of Science and Technology, chapter~4,
  pages 105--115.
\newblock (In Hebrew).

\bibitem[\protect\citename{Wintner and Francez}1995a]{shuly:nlulp-95}
Wintner, Shuly and Nissim Francez.
\newblock 1995a.
\newblock An abstract machine for typed feature structures.
\newblock In {\em Proceedings of the 5th Workshop on Natural Language
  Understanding and Logic Programming}, pages 205--220, Lisbon, May.

\bibitem[\protect\citename{Wintner and Francez}1995b]{shuly:iwpt95}
Wintner, Shuly and Nissim Francez.
\newblock 1995b.
\newblock Parsing with typed feature structures.
\newblock In {\em Proceedings of the Fourth International Workshop on Parsing
  Technologies}, pages 273--287, Prague, September.

\bibitem[\protect\citename{Wintner and Francez}to appear]{shuly:jolli}
Wintner, Shuly and Nissim Francez.
\newblock to appear.
\newblock Off-line parsability and the well-foundedness of subsumption.
\newblock {\em Journal of Logic, Language and Information}.

\bibitem[\protect\citename{Wintner, Gabrilovich, and Francez}1997]{amalia-man}
Wintner, Shuly, Evgeniy Gabrilovich, and Nissim Francez, 1997.
\newblock {\em {AMALIA} -- {A}bstract {MA}chine for {LI}inguistic
  {A}pplications -- User's Guide}.
\newblock Laboratory for Computational Linguistics, Computer Science
  Deparmtent, Technion, Israel Institute of Technology, 32000 Haifa, Israel,
  January.

\bibitem[\protect\citename{Wintner and Ornan}1991]{shuly:syntactic-analysis}
Wintner, Shuly and Uzzi Ornan.
\newblock 1991.
\newblock Syntactic analysis of {H}ebrew sentences.
\newblock In {\em Proceedings of the 8th Israeli Symposium on Artificial
  Intelligence and Computer Vision}, pages 201--230. Information Processing
  Association of Israel, December.

\bibitem[\protect\citename{Wintner and Ornan}1996]{shuly:jnle}
Wintner, Shuly and Uzzi Ornan.
\newblock 1996.
\newblock Syntactic analysis of hebrew sentences.
\newblock {\em Natural Language Engineering}, 1(3):261--288.

\bibitem[\protect\citename{Yizhar}1993]{dana}
Yizhar, Dana.
\newblock 1993.
\newblock Computational grammar for {H}ebrew noun phrases.
\newblock Master's thesis, Computer Science Department, Hebrew University,
  Jerusalem, Israel, June.
\newblock (In Hebrew).

\bibitem[\protect\citename{Zajac}1992]{tfs}
Zajac, R\'{e}mi.
\newblock 1992.
\newblock Inheritance and constraint-based grammar formalisms.
\newblock {\em Computational Linguistics}, 18(2):159--182.

\end{thebibliography}

\end{document}